\documentclass[twocolumn]{IEEEtran} 
\usepackage[T1]{fontenc}
\usepackage{ae,aecompl}
\usepackage{balance}
\usepackage{amsthm}
\usepackage{amssymb}
\usepackage{times}
\usepackage{caption}
\usepackage[labelformat=simple]{subcaption}
\usepackage{graphicx}
\usepackage{graphics}
\usepackage{blkarray}
\usepackage{booktabs}
\usepackage{xcolor} 
\usepackage{bbold}
\usepackage{caption}
\usepackage{subcaption}
\usepackage{multirow}
\usepackage{comment}
\usepackage{epstopdf}
\usepackage{tikz}
\usepackage{todonotes}
\graphicspath{{./figures/}}
\usepackage{csquotes}
\usepackage{siunitx}
\usepackage{cite}
\usepackage{epigraph}
\usepackage{empheq}
\usepackage{comment}
\usepackage{cancel}
\usepackage{algorithm} 
\usepackage{algorithmic}  
\usepackage{mathtools}
\usepackage{balance}
\usepackage{dirtytalk}

\usepackage{listings}
\usepackage{xcolor}
\usepackage[algo2e,linesnumbered,ruled,vlined]{algorithm2e}

\usepackage{hyperref}
\DeclareMathOperator*{\argmin}{\arg\!\min}

\newcommand{\bt}{\mathcal{T}}

\SetKwRepeat{Do}{do}{while}

\makeatother

\DeclareCaptionLabelSeparator{periodspace}{.\quad}
\newtheorem{lemma}{Lemma}
\newtheorem{remark}{Remark}

\newtheorem{example}{Example}

\newtheorem{corollary}{Corollary}

\newtheorem{definition}{Definition}
\newtheorem{experiment}{Experiment}

\makeatletter
\newcommand*{\rom}[1]{\expandafter\@slowromancap\romannumeral #1@}
\makeatother

\def\eqalignno#1{\let\\=\cr\displ@y \tabskip\@centering
  \halign to\displaywidth{\hfil$\@lign\displaystyle{##}$\tabskip\z@skip
    &$\@lign\displaystyle{{}##}$\hfil\tabskip\@centering
    &\llap{$\@lign##$}\tabskip\z@skip\crcr
    #1\crcr}}
\def\leqalignno#1{\let\\=\cr\displ@y \tabskip\@centering
  \halign to\displaywidth{\hfil$\@lign\displaystyle{##}$\tabskip\z@skip
    &$\@lign\displaystyle{{}##}$\hfil\tabskip\@centering
    &\kern-\displaywidth\rlap{$\@lign##$}\tabskip\displaywidth\crcr
    #1\crcr}}

\makeatletter
\providecommand\@dotsep{5}
\renewcommand{\listoftodos}[1][\@todonotes@todolistname]{%
  \@starttoc{tdo}{#1}}
\makeatother

\def\BibTeX{{\rm B\kern-.05em{\sc i\kern-.025em b}\kern-.08em
    T\kern-.1667em\lower.7ex\hbox{E}\kern-.125emX}}

\begin{document}

\title{Handling Concurrency in Behavior Trees} 

\author{Michele Colledanchise and Lorenzo Natale
         \thanks{The authors are with the Humanoids Sensing and Perception Laboratory, Center for Robotics and Intelligent Systems (CRIS), Istituto Italiano di Tecnologia. Genoa, Italy.
e-mail: {\tt{ michele.colledanchise@iit.it, lorenzo.natale@iit.it}}}}

\markboth{IEEE Transactions On Robotics, Vol. XX, No. Y, Month
2021}
{Murray and Balemi: Using the Document Class IEEEtran.cls} 

\maketitle

\begin{abstract}
This paper addresses the concurrency issues affecting Behavior Trees (BTs), a popular tool to model the behaviors of autonomous agents in the video game and the robotics industry. 

BT designers can easily build complex behaviors composing simpler ones, which represents a key advantage of BTs. The parallel composition of BTs expresses a way to combine concurrent behaviors that has high potential, since composing pre-existing BTs in parallel results easier than composing in parallel classical control architectures, as finite state machines or teleo-reactive programs. However, BT designers rarely use such composition due to the underlying concurrency problems similar to the ones faced in concurrent programming. As a result, the parallel composition, despite its potential, finds application only in the composition of simple behaviors or where the designer can guarantee the absence of conflicts by design.

In this paper, we define two new BT nodes to tackle the concurrency problems in BTs and we show how to exploit them to create predictable behaviors. In addition, we introduce measures to assess execution performance and show how different design choices affect them. We validate our approach in both simulations and the real world. Simulated experiments provide statistically-significant data, whereas real-world experiments show the applicability of our method on real robots. We provide an open-source implementation of the novel BT formulation and published all the source code to reproduce the numerical examples and experiments. 

 \end{abstract}
\section{Introduction}
\IEEEPARstart{W}e study concurrent robot behaviors encoded as Behavior Trees (BTs)~\cite{BTBook}. Robotics applications of BTs span from manipulation~\cite{rovida2017extended, zhang2019ikbt,csiszar2017behavior}
to non-expert programming~\cite{coronado2018development,paxton2017costar,shepherd2018engineering}. Other works include task planning \cite{neufeld2018hybrid}, human-robot interaction~\cite{kim2018architecture, axelsson2019modelling,ghadirzadeh2020human}, learning~\cite{sprague2018adding, banerjee2018autonomous,hannaford2019hidden,scheidelearning},  UAV~\cite{safronov2019asynchronous,sprague2018improving, ogren, bruggemann2017analysing, crofts2017behaviour,molina2020building}, multi-robot systems~\cite{biggar2020framework,tadewos2019fly,kuckling2018behavior,ozkahraman2020combining}, and system analysis~\cite{biggar2020principled,de2020reconfigurable,ogren2020convergence}.
The Boston Dynamics's Spot uses BTs to model the robot's mission~\cite{spot}, the Navigation Stack  and the task planner of ROS2 uses BTs to encode the high level robot's behavior~\cite{macenski2020marathon,PlanSys2}.

The particular syntax and semantic of BTs, which we will describe in the paper, allows creating easily complex behaviors composing simpler ones.
A BT designer can compose behaviors in different ways, each with its own semantic. The \emph{Parallel} composition allows a designer to describe the concurrent execution of several sub-behaviors. In BTs, this composition scales better as the complexity of the behavior increases, compared to other control architectures where the system's complexity results from the product of its sub-systems' complexities~\cite{BTBook}. However, the Parallel composition still entails concurrency issues  (e.g., race
conditions, starvation, deadlocks, etc.), like any other control architecture. As a result, such composition gets applied only to orthogonal tasks. 

In the BT literature, the Parallel composition finds applications only to the composition of orthogonal tasks, where the designer guarantees the absence of conflicts. In this paper, we show how we can extend the use of BTs to address the concurrency issues above. In particular, we show how to obtain synchronized concurrent BTs execution, exclusive access to resources, and predictable execution times. We define performance measures and analyze how they are affected by different design choices. We also provide reproducible experimental validation by publishing the implementation of our BT library, code, and data related to our experiments.

\begin{figure*}[t]

\centering
\begin{subfigure}[t]{0.6\columnwidth}
\centering
  \includegraphics[width=\columnwidth]{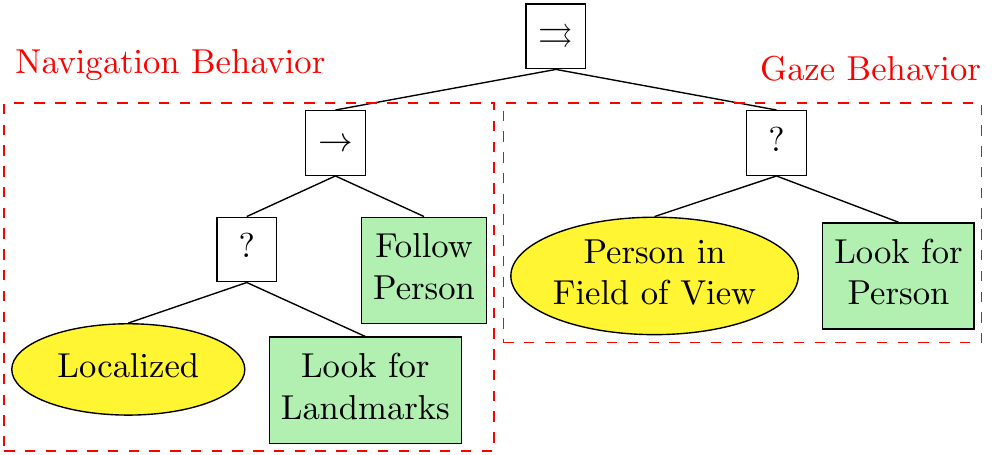}
    \caption{Flawed BT execution. The conflicting actions \emph{Look for Person} and \emph{Look for Landmarks} may be executed concurrently.}
    \label{fig:intro:incorrect}
      \end{subfigure}
      \hfill
  \begin{subfigure}[t]{0.7\columnwidth}
  \centering
  \includegraphics[width=\columnwidth]{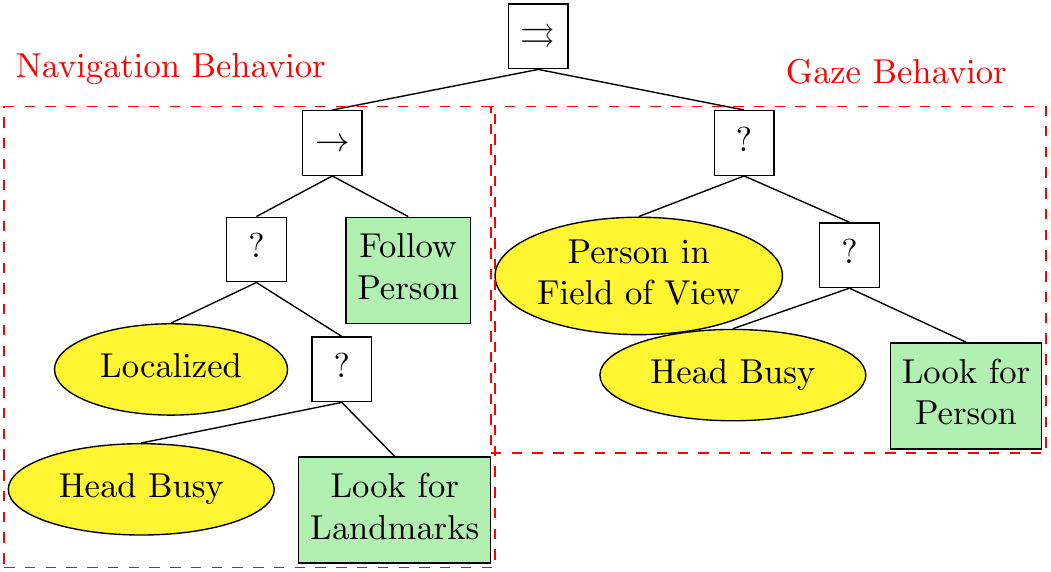}
      \caption{Correct BT execution using the classical BT nodes.}
          \label{fig:intro:correct}
  \end{subfigure}
  \hfill
  \begin{subfigure}[t]{0.6\columnwidth}
\centering
  \includegraphics[width=\columnwidth]{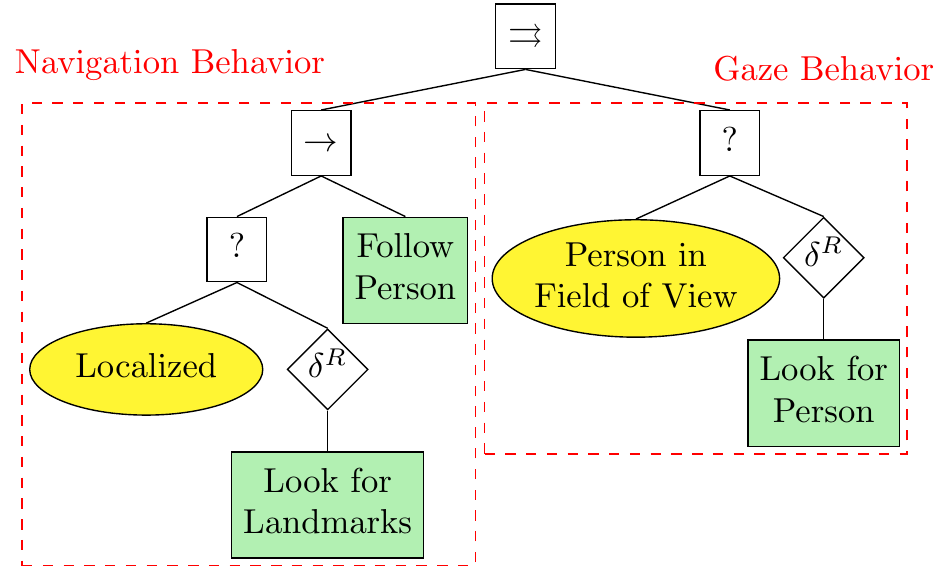}
    \caption{Correct BT execution using the proposed BT nodes.}
    \label{fig:intro:proposed}
      \end{subfigure}
  \caption{Example of flawed and correct concurrent BT execution. The gaze and navigation behaviors are executed in parallel.}
\end{figure*}

Software developers from the video game industry conceived BTs to model the behaviors of Non-Player Characters (NPCs)~\cite{isla2005handling,millington2009artificial}.
Controlled Hybrid Systems (CHSs)~\cite{lunze2009handbook}, which combine continuous and discrete dynamics, were a natural formulation of NPCs' execution and control.  However, it turned out that CHSs were not fit to program an NPC, as CHSs implement the discrete dynamics in the form of Finite State Machines (FSMs). The developers realized that FSMs scale poorly, hamper modification and reuse, and have proved inadequate to represent complex deliberation~\cite{sloan2011feasibility,champandard10reasons}. In this context, the issues lie in the transitions of FSMs, which implement a \emph{one-way control transfer}, similar to a GOTO statement of computer programming. In 1968, Edsger Dijkstra observed that \say{\emph{the GOTO  statement as it stands is just too primitive; it is too much an invitation to make a mess of one's program [...] 
the GOTO statement should be abolished from all higher-level programming languages}}~\cite{dijkstra1968go}. In the computer programming community, Dijksta's observation started a controvert debate against the expressivity power of GOTO statements~\cite{rubin1987goto, benander1990empirical, ashenhurst1987goto}. Finally, the community followed Dijksta's advice, and modern software no longer contains GOTO statements. However, we still can find GOTO statements everywhere in the form of \emph{transitions} in FSMs and therefore CHSs.

The robotics community identified similarities in the desired behaviors for NPCs and robots. In particular, both NPCs and robots act in an unpredictable and dynamic environment; both need to encode different behaviors in a hierarchy, both need a compact representation for their behaviors. Quickly, BTs became a modular, flexible, and reusable alternative over FSM to model robot behaviors~\cite{iovino2020survey}. Moreover, the robotic community showed how BTs generalize successful robot control architectures such
as the Subsumption Architecture and the Teleo-Reactive
Paradigm~\cite{BTBook}.
Using BTs, the designer can compose simple robot behaviors using the so-called \emph{control flow nodes}: Sequence, Fallback, Decorator, and Parallel. As mentioned, the Parallel composition of independent behaviors can arise several concurrency problems in any modeling language, and BTs are no exception. However, the parallel composition of BTs remains less sensitive to dimensionality problems than classical FSMs~\cite{BTBook}.

Another similarity between computer programming and robot behavior design lies in the concurrent execution of multiple tasks. A computer programmer adopts synchronization techniques to achieve \emph{execution synchronization}, where a process has to wait until another process provides the necessary data, or \emph{data synchronization}, where multiple processes have to use or access a critical resource and a correct synchronization strategy ensures that only one process at a time can access them~\cite{taubenfeld2006synchronization}.
The solutions adopted in concurrent programming made a tremendous impact on software development. Another desired quality of a process, particularly in real-time systems, is the \emph{predictability}, that is, the ability to ensure the execution of a process regardless of outside factors that will jeopardize it. In other words, the application will behave as intended in terms of functionality, performance, and response time.

Nowadays, robot software follows a modular approach, in which computation and control use concurrent execution of interconnected modules. This philosophy is promoted by middlewares like ROS and has become a de\-facto standard in robotics. The presence of concurrent behaviors requires facing the same issues affecting concurrent programming, which deals with the execution of several concurrent processes that need to be synchronized to achieve a task or simply to avoid being in conflict with one another. 
In this context, proper synchronization and resource management become beneficial to achieve faster and reproducible behaviors, especially at the developing stage, where actions may run at a different speed in the real world and in a simulation environment. Increasing predictability reduces the difference between simulated and real-world robot executions and increases the likelihood of task completion within a given time constraint. We believe that proper parallel task executions will bring benefits in terms of efficiency and multitasking to BTs in a similar way as in computer programming.

The requirement of concurrent or predictable behaviors may also come from non-technical specifications. For example, the Human-Robot Interaction (HRI) community stressed the importance of synchronized robots' behaviors in several contexts. The literature shows evidence of more \say{believable} robots behaviors when they exhibit contingent  movements~\cite{fischer2013impact} (e.g., gaze and arm movement when giving directions), coordinated robots and human movements~\cite{lee2011vision} (e.g., a rehabilitation robot moves at the patient's speed), and coordinated gestures and dialogues~\cite{kopp2006towards} (e.g., the robot's gesture synchronized during dialogues).

In this paper, we extend our previous works~\cite{colledanchise2018improving, colledanchise2019analysis}, we define Concurrent BTs (CBTs), where nodes expose information regarding progress and resource used, we also define and implement two new control flow nodes that allow progress and data synchronization techniques and show how to improve behavior predictability. In addition, we introduce measures to assess execution performance and show how design choices affect them. 
To clarify what we mean by these concepts, we consider the following task: \emph{A robot has to follow a person}. 
The robot's behavior can be encoded as the concurrent execution of two sub-behaviors: \emph{navigation} and \emph{gazing}. However, the navigation behavior is such that, whenever the robot gets lost, it moves the head, looking for visual landmarks to localize itself.  
Figure~\ref{fig:intro:incorrect} encodes the BT of this behavior.
At this stage, the semantic of BT is not required to understand the problem.
Note how, whenever the robot gets lost, two actions require the use of the head: the actions \emph{Look for Landmarks} and  \emph{Look for Person}. To avoid conflicts, we have to modify to be as in Figure~\ref{fig:intro:correct}. However, such BT goes against the separation of roles as the BT designer needs to know beforehand the actions' resources. In this paper, we propose two control flow nodes that allow the synchronization of such BT in a less invasive fashion, as in Figure~\ref{fig:intro:proposed}.

The concurrent execution of legacy BTs represents another example of such a synchronization mechanism.
Clearly, the design of a single action that performs both tasks represents a \say{better} synchronized solution. However, creating the single action for composite behaviors jeopardizes the advantages of BTs in terms of modular and reusable behavior.

To summarize, in this paper, we provide an extension of our previous work \cite{colledanchise2018improving} and \cite{colledanchise2019analysis}, where the new results are:
\begin{itemize}
\item  We moved the synchronization logic from the parallel node to the decorator nodes. This enables a higher expressiveness of the synchronization.
\item We provide reproducible experimental validation on simulated data.
\item We provide experimental validation on three real robots. 
\item We compared our approach with two alternative task synchronization techniques. 
\item We provide the code to extend an existing software library of BTs, and its related GUI, to encode the proposed synchronization nodes.
\item We provide a theoretical discussion of our approach and identify the assumptions under which the property on BTs are not jeopardized by the synchronization.
\end{itemize}

The outline of this paper is as follows:   We present the related work and compare it with our approach in Section~\ref{sec:related}. We overview the background in BTs in Section~\ref{sec:background}. We present the first contribution of this paper on BT synchronization in Section~\ref{sec:synchronizaton}. Then we present the second contribution on performance measures in Section~\ref{sec:measures}.  In Section~\ref{sec:experimental} we provide experimental validation with numerical experiments to gather statistically significant data and compare our approach with existing ones. We made these experiments reproducible. We also validated our approach on real robots in three different setups to show the applicability of the approach to real problems. We describe the software library we developed, and we refer to the online repository in Section~\ref{sec:library}. We study the new control nodes from a theoretical standpoint and study how design choices affect the performances in Section~\ref{sec:analysis}. We conclude the paper in Section~\ref{sec:conclusions}.

\section{Related Work}
\label{sec:related}

This section shows how BT designers in the community exploit the parallel composition, and we highlight the differences with the proposed approach.  We do not compare our approach with generic scheduling algorithms~\cite{brunner2019autonomous}, as our interest lies in the concurrent behaviors encoded as BTs. 

The parallel composition has found relatively little use, compared to the other compositions, due to the intrinsic concurrency issues similar to the ones of computer programming, such as race conditions and deadlocks. Current applications that make use of the parallel composition work under the assumption that sub-BTs lie on orthogonal state spaces \cite{champandard2007enabling, rovida2017extended} or that sub-BTs executed in parallel have a predefined priority assigned~\cite{weber2010reactive} where, in conflicting situations, the sub-BTs with the lower priority stops. Other applications impose a mutual exclusion of actions in sub-BTs whenever they have potential conflicts (e.g., sending commands to the same actuator)~\cite{BTBook} or 
they assume that sub-BTs that are executed in parallel are not in conflict by design.

The parallel composition found large use in the BT-based task planner  \emph{A Behavior Language} (ABL)~\cite{mateas2002behavior} and in its further developments. ABL was originally designed for the game \emph{Fa\c{c}ade}, and it has received attention for
its ability to handle planning and acting at different deliberation layers, in particular, in Real-Time Strategy games~\cite{weber2010reactive}.
ABL executes sub-BTs in parallel and resolves conflicts between multiple concurrent actions by defining a fixed priority order. This solution threatens the reusability and modularity of BTs and introduces a latent hierarchy in the BT.

The parallel composition found use also in multi-robot applications, both with centralized~\cite{agis2020event} and distributed fashions~\cite{colledanchise2016advantages,yang2020hierarchical}, resulting in improved fault tolerance and other performances. 
The parallel node involves multiple robots, each assigned to a specific task using a task-assignment algorithm. A task-assignment algorithm ensures the absence of conflicts.

None of the existing work in the BT literature adequately addressed the synchronization issues that arise when using a parallel BT node. They assume or impose strict constraints on the actions executed and often introduce undesired latent hierarchies difficult to debug.

A recent work~\cite{rovida2018motion} proposed BTs for executing actions
in parallel, even when they lie on the same state space (e.g.,
they use the same robot arm). The authors implement the coordination mechanism with a BT that activates and deactivates motion primitives based on their pre-conditions. Such a framework avoids that more actions access a critical resource concurrently. In our work, we are interested in synchronizing the progress of actions that a BT can execute concurrently.

We address the issues above by defining BT nodes that expose information regarding progress and resource uses. We define an absolute and relative synchronized parallel BT node execution and a resource handling mechanism.  We provide a set of statistically meaningful experiments and real-robot executions. We also provide an extension to the software library to obtain such synchronizations and real-robot examples. This makes our paper fundamentally different than the ones presented above and the BT literature.

\begin{figure}[h]
\centering
\begin{subfigure}[t]{0.4\columnwidth}
\centering
\includegraphics[width=0.8\columnwidth]{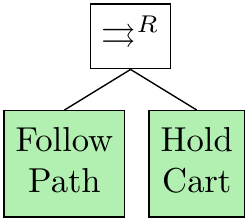}
\caption{BTs synchronization using the previous formulation.}
\label{fig:rw:bt:old}
\end{subfigure}
\begin{subfigure}[t]{0.45\columnwidth}
\centering
\includegraphics[width=0.6\columnwidth]{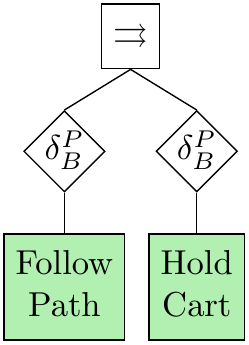}
\caption{BTs synchronization using the proposed formulation.}
\label{fig:rw:bt:new}
\end{subfigure}
\caption{BT synchronization using the previous~\cite{colledanchise2018improving} (left) and the proposed formulation (right). }
\label{fig:rw:bt:oldvsnew}
\end{figure}

\begin{figure}[h!]
\centering
\includegraphics[width=\columnwidth]{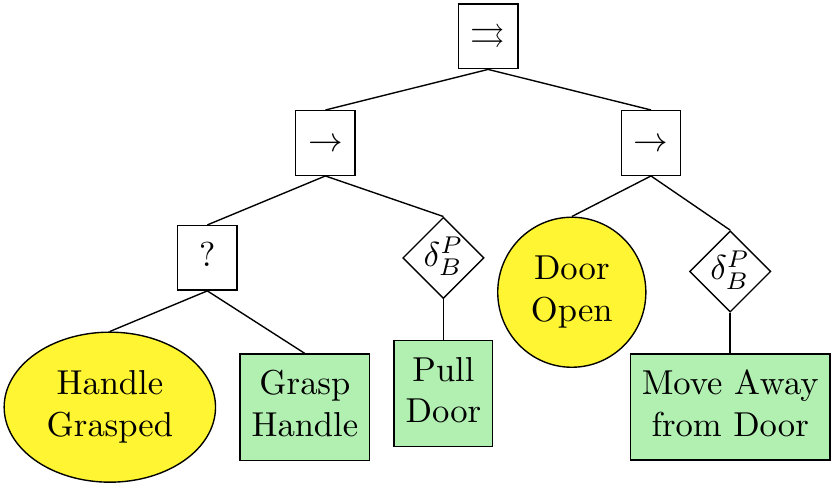}
\caption{A more complex version of the BT for Example~\ref{ex:absolute} allowed by the new formulation only.}
\label{fig:ex:absolute:bt:complex}
\end{figure}

In our previous work~\cite{colledanchise2018improving,colledanchise2019analysis}, we extend the semantic of the parallel node to allow synchronization. Figure~\ref{fig:rw:bt:old} shows an example of a synchronized BT using our such approach. In this paper, we go beyond our previous work by moving the synchronization logic inside a decorator node, as Figure~\ref{fig:rw:bt:new}. That allows the synchronization to deeper branches of the BT, as in Figure~\ref{fig:ex:absolute:bt:complex}, and multiple cross synchronization. In Section~\ref{sec:experimental}, we will also show a synchronization experiment possible only with the new semantic.

\newpage

\section{Background}
\label{sec:background}

This section briefly presents the classical and the state-space formulation of BTs. A detailed description of BTs is available in the literature~\cite{BTBook}.

\subsection{Classical Formulation of Behavior Trees}
\label{sec:background.BT}

A BT is a graphical modeling language that represents actions orchestration. It is a directed rooted tree where the internal nodes represent behavior compositions and leaf nodes represent actuation or sensing operations. We follow the canonical nomenclature for root, parent, and child nodes.

The children of a BT node are placed below it, as in Figure~\ref{fig:ex:absolute:bt:complex}, and they are executed in the order from left to right. The execution of a BT begins from the root node. It sends \emph{ticks}, which are activation signals, with a given frequency to its child. A node in the tree is executed if and only if it receives ticks. When the node no longer receives ticks, its execution stops.  The child returns to the parent a status, which can be either \emph{Success}, \emph{Running}, or \emph{Failure} according to the node's logic. Below we present the most common BT nodes and their logic.

In the classical representation, there are four operator nodes (Fallback, Sequence, Parallel, and Decorator) and two execution nodes (Action and Condition). There exist additional operators, but we will not use them in this paper.

\paragraph*{Sequence}

When a Sequence node receives ticks, it routes them to its children in order from the left. It returns Failure or Running whenever a child returns Failure or Running, respectively. It returns Success whenever all the children return Success. When child $i$ returns Running or Failure, the Sequence node does not send ticks to the next child (if any) but keeps ticking all the children up to child $i$.
The Sequence node is graphically represented by a square with the label \say{$\rightarrow$}, as in Figure~\ref{fig:ex:absolute:bt:complex}, and its pseudocode is described in Algorithm~\ref{bts:alg:sequence}.



\begin{algorithm2e}[h!]
\SetKwProg{Fn}{Function}{}{}

\Fn{Tick()}
{
  \For{$i \gets 1$ \KwSty{to} $N$}
  {
    \ArgSty{childStatus} $\gets$ \ArgSty{child($i$)}.\FuncSty{Tick()}\\
    \uIf{\ArgSty{childStatus} $=$ \ArgSty{Running}}
    {
      \Return{Running}
    }
    \ElseIf{\ArgSty{childStatus} $=$ \ArgSty{Failure}}
    {
      \Return{Failure}
    }
  }
  \Return{Success}
  }
  \caption{Pseudocode of a Sequence operator with $N$ children}
  \label{bts:alg:sequence}
\end{algorithm2e}

\paragraph*{Fallback}

When a Fallback node receives ticks, it routes them to its children in order from the left. It returns a status of Success or Running whenever a child returns Success or Running respectively. It returns Failure whenever all the children return Failure. When child $i$ returns Running or Success, the Fallback node does not send ticks to the next child (if any) but keeps ticking all the children up to the child $i$.
The Fallback node is represented by a square with the label \say{$?$}, as in Figure~\ref{fig:ex:absolute:bt:complex}, and its pseudocode is described in Algorithm~\ref{bts:alg:fallback}.

\begin{algorithm2e}[h!]
\SetKwProg{Fn}{Function}{}{}

\Fn{Tick()}
{
  \For{$i \gets 1$ \KwSty{to} $N$}
  {
    \ArgSty{childStatus} $\gets$ \ArgSty{child($i$)}.\FuncSty{Tick()}\\
    \uIf{\ArgSty{childStatus} $=$ \ArgSty{Running}}
    {
      \Return{Running}
    }
    \ElseIf{\ArgSty{childStatus} $=$ \ArgSty{Success}}
    {
      \Return{Success}
    }
  }
  \Return{Failure}
  }
  \caption{Pseudocode of a Fallback operator with $N$ children}
    \label{bts:alg:fallback}
\end{algorithm2e}

\paragraph*{Parallel}
When the Parallel node receives ticks, it routes them to all its children. It returns Success if $M \geq N$ children return Success, it returns Failure if more than $N - M$ children return Failure, and it returns Running otherwise.
The Parallel node is graphically represented by a square with the label \say{$\rightrightarrows$}, as in Figure~\ref{fig:ex:absolute:bt:complex}, and its pseudocode is described in Algorithm~\ref{bts:alg:parallel}.
%
\begin{algorithm2e}[h!]
\SetKwProg{Fn}{Function}{}{}

\Fn{Tick()}
{
  \ForAll{$i \gets 1$ \KwSty{to} $N$}
  {
    \ArgSty{childStatus}[i] $\gets$ \ArgSty{child($i$)}.\FuncSty{Tick()}\\
    }
    \uIf{$\Sigma_{i: \ArgSty{childStatus}[i]=Success} = M$}
    {
      \Return{Success}
    }
    \ElseIf{$\Sigma_{i: \ArgSty{childStatus}[i] =Failure} > N - M $}
    {
      \Return{Failure}
    
  }\Else{
  \Return{Running}
  }
  }
    \caption{Pseudocode of a Parallel operator with $N$ children}
  \label{bts:alg:parallel}
\end{algorithm2e}

\paragraph*{Decorator}
A Decorator node represents a particular control flow node with only one child. When a Decorator node receives ticks, it routes them to its child according to custom-made policy. It returns to its parent a return status according to a custom-made policy. The Decorator node is graphically represented as a rhombus, as in Figure~\ref{fig:ex:absolute:bt:complex}. BT designers use decorator nodes to introduce additional semantic of the child node's execution or to change the return status sent to the parent node. 

\paragraph*{Action}
An action performs some operations as long as it receives ticks. It returns Success whenever the operations are completed and Failure if the operations cannot be completed. It returns Running otherwise. When a running Action no longer receives ticks, its execution stops.
An Action node is graphically represented by a rectangle, as in Figure~\ref{fig:ex:absolute:bt:complex},  and its pseudocode is described in Algorithm~\ref{bts:alg:action}.

\begin{algorithm2e}[h]
\SetKwProg{Fn}{Function}{}{}

\Fn{Tick()}
{
 \ArgSty{DoAPieceOfComputation()} \\
    \uIf{action-succeeded}
    {
      \Return{Success}
    }
    \ElseIf{action-failed}
    {
      \Return{Failure}
    }
    \Else
    {
    \Return{Running}
    }
   }
  \caption{Pseudocode of a BT Action}
  \label{bts:alg:action}
\end{algorithm2e}
\paragraph*{Condition}
Whenever a Condition node receives ticks, it checks if a proposition is satisfied or not. It returns Success or Failure accordingly. A Condition is graphically represented by an ellipse, as in Figure~\ref{fig:ex:absolute:bt:complex}, and its pseudocode is described in Algorithm~\ref{bts:alg:condition}.

\begin{algorithm2e}[h!]
\SetKwProg{Fn}{Function}{}{}

\Fn{Tick()}
{
    \uIf{condition-true}
    {
      \Return{Success}
    }
    \Else
    {
      \Return{Failure}
    }
   }
  \caption{Pseudocode of a BT Condition}
  \label{bts:alg:condition}
\end{algorithm2e}
\vspace*{-1em}
\subsection{Control Flow Nodes With Memory}

To avoid
the unwanted re-execution of some nodes, and save computational resources, the BT community developed control flow nodes with memory~\cite{millington2009artificial}.
Control flow nodes with memory keep stored which child has returned Success or Failure, avoiding their re-execution. Nodes with memory are graphically represented with the addition
of the symbol \say{$*$} as superscript (e.g., a Sequence node with memory is graphically represented
by a box with the label \say{$\rightarrow^*$}). The memory is cleared when the parent node returns either
Success or Failure so that, at the next activation, all children are re-considered. Every execution of a control flow node with memory can be obtained
employing the related non-memory control flow node using auxiliary conditions and shared memory~\cite{BTBook}. Provided a shared memory, these nodes become syntactic sugar.

\subsection{Asynchronous Action Execution}
Algorithm~\ref{bts:alg:action} performs a step of computation at each tick. 
It implements an action execution that is synchronous to the ticks' traversal.
However, in a typical robotics system, action
nodes control the robot by sending commands to a robot's interface to execute a particular skill, such as an arm movement or a navigation skill; these skills are often executed by independent components running on a distributed system. Therefore, the action execution gets delegated to different executables that communicate via a middleware.

As discussed in the literature \cite{BTBook,colledanchise2021implementation}, the designer needs to ensure that the skills running in the robot independently from the BT get properly interrupted when the corresponding action no longer receives ticks. 

\begin{figure}[h!]
\centering
\includegraphics[width=\columnwidth]{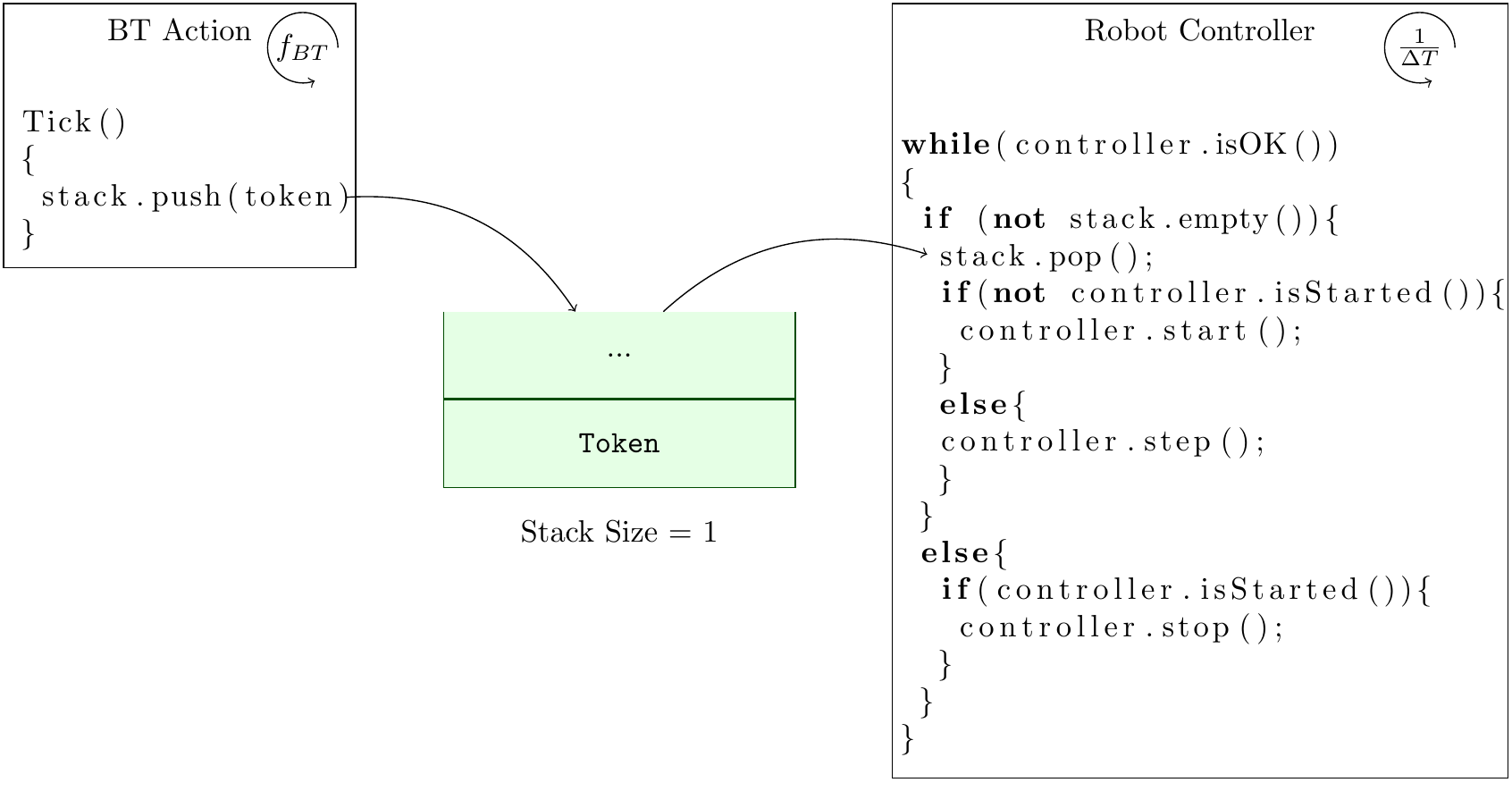}
\caption{Example of asynchronous external action execution. $f_{BT}$ is the tick frequency and $\Delta T$ is the quantum period.}
\label{fig:stack}
\end{figure}

To support the preemption and synchronization, BT designers split the actions execution in smaller steps, each executed within a \emph{quantum}, that is, a time window during which the action gets executed by the robot asynchronously with respect to the BT. During this time, the action cannot be interrupted. The action starts when the first tick is received, and it proceeds for another quantum only when the next tick is received. At the end of each quantum, a running action yields control back to the BT. 
 This logic resembles process scheduling, where a scheduler provides computing time, to a process, and then it takes back control to choose the next process to run. 
Figure~\ref{fig:stack} shows an example of how a BT action interacts with an external executable that controls the robot. The figure depicts two threads, one that ticks the action node (within the BT executable) and one that controls the robot (within an external executable). When the action node receives a tick from its parent, it pushes a token onto a stack, shared with the external executable that controls the robot. Such executable controls the robot if and only if there is a token in the stack. This behavior also is outlined in the algorithm in Figure~\ref{fig:stack}.
The executable checks the stack periodically, if there is a token in the stack, it consumes it, and it executes a control step. If the stack is empty, the executable halts the controller execution. If the BT tick frequency is faster than the controller quantum (e.g., twice the thread's frequency), this mechanism ensures that the controller operates without interruptions, but it also guarantees that the controller is halted when ticks are no longer dispatched to the action node without delay (this is achieved using the size of the stack equal to one).


It is clear that, in BTs, the tick frequency plays an important role in action preemption. To allow \say{fast} preemption, the quantum of actions should be short and the tick frequency should be high. Intuitively, a blocking action, which does not allow to be interrupted throughout its entire execution, will continue to take control of the robot also if it no longer receives ticks.  BTs orchestrate behaviors at a relatively high level of abstraction.
In general, to avoid preemption delay, the time spanned between a tick and the next one must be shorter than the smallest action quantum in the BT. 
In our experience, a quantum of $\Delta T\approx100ms$ (i.e., an update frequency of $10Hz$) and a tick frequency of $20Hz$ represents a good trade-off between action responsiveness and required ticks traversal frequency.
 
The interaction between the BT and the executable is designed to ensure that the robot controller continues to operate if the BT ticks the action node without interruptions. Otherwise, i.e. if no ticks are received within the assigned quantum, the controller is halted.


%

	\section{Concurrent BTs}
\label{sec:synchronizaton}
This section introduces the first contribution of the paper. We present the Concurrent BTs (CBTs), an extension to classical BTs with the addition of the execution progress and the resource allocated in the formulation. We extend our previous work on parallel synchronization of BTs~\cite{colledanchise2018improving, colledanchise2019analysis} employing decorator nodes that allow progress and resource synchronization. Here we define these nodes by their pseudocode. We provide the source code for some of the examples provided.\footnote{\url{https://github.com/miccol/tro2021-code}}

In Section~\ref{sec:analysis} we provide the formal definition and the state-space formulation of the nodes. We also prove that, under some assumptions, the proposed nodes do not jeopardize the BT properties.

\paragraph*{Concurrent BTs}
A CBT is a  BT whose nodes expose information about the execution progress $p(x_k)$ and the resource required $Q(x_k)$ and priority $\rho(x_k)$ at system's state $x_k \in \mathbb{R}^n$. In addition, the nodes contain the user-defined function $g(x_k)$ that represents the priority increase whenever the execution of a node gets denied by a resource not available; we will present the details in this section.
In Section~\ref{sec:analysis} we provide the formal definition of CBTs and the formulation of the Sequence and Fallback composition.

\begin{figure}[h!]
\centering
\begin{subfigure}[t]{0.32\columnwidth}
\centering
\includegraphics[width=0.5\columnwidth]{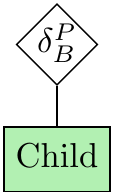}
\caption{ Absolute synchronization. $B$ indicates the set of barriers.}
\label{fig:pdec}
\end{subfigure}
\begin{subfigure}[t]{0.32\columnwidth}
\centering
\includegraphics[width=0.5\columnwidth]{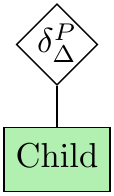}
\caption{ Relative synchronization. $\Delta$ indicates the threshold value.}
\end{subfigure}
\begin{subfigure}[t]{0.32\columnwidth}
\centering
\includegraphics[width=0.5\columnwidth]{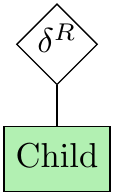}
\caption{ Resource Synchronization Decorator node.}
\label{fig:rdec}
\end{subfigure}
\caption{Graphical representation of a Synchronization Decorator nodes.}
\end{figure}

\paragraph*{ProgressSynchronization Decorator}
When a ProgressSynchronization Decorator receives a tick, it ticks its child if the child's progress at the current state $p(x_k)$ is lower than the current barrier $b(x_k)$. The decorator returns to the parent Success if the child returns Success, it returns to the parent Failure if the child returns Failure. It returns Running otherwise. 
The ProgressSynchronization Decorator node is graphically represented by a rhombus with the label \say{$\delta^P_b$}, as in Figure~\ref{fig:pdec}, and its pseudocode is described in Algorithm~\ref{alg:progress}.

We will calculate the barrier $b(x_k)$ either in an absolute or relative fashion, as we will show in  Sections~\ref{PM:AS} and~\ref{PM:RS}.

\begin{algorithm2e}[h!]
\SetKwProg{Fn}{Function}{}{}
\Fn{Tick()}
{

\If{\ArgSty{child}.p($x_k$) $\leq$ \ArgSty{b($x_k$)} }
{
      \ArgSty{childStatus} $\gets$  \ArgSty{child.Tick()}\\
  \Return{childStatus}
}
    \Return{Running}
  }
    \caption{Pseudocode of a ProgressSynchronization Decorator.}
 \label{alg:progress}
\end{algorithm2e}

\vspace{-1em}

\newpage

\paragraph*{ResourceSynchronization Decorator}
When a ResourceSynchronization Decorator receives a tick, it ticks its child if the resources required by the child $i$, $Q_i(x_k)$, are either available or assigned to that child already.   When the decorator ticks a child, it also assigns all the resources in $Q_i(x_k)$ to that child. Whenever the child no longer requires a resource in $Q_i(x_k)$, such resource get released.
The decorator returns to the parent Success if the child returns Success, It returns to the parent Failure if the child returns Failure. It returns Running if either the child return running or if the child is waiting for a resource. $R$ is the set of all the resources of the system. The decorator keeps also a priority value for the subtree accessing a resource, to avoid starvation, as we prove it in Section~\ref{sec:analysis}. Whenever the decorator receives a tick and does not send it to the child (as the resources are not ailable), the priority value evolves according to the $g(x_k)$. In Section~\ref{sec:experimental} we will show two examples that highlight how the choice of the function $g$ avoids starvation. The BT keeps track of the node currently using a resource $q$, via the function $\alpha(q)$. All the resource decorator nodes share the value of such function.

The ResourceSynchronization Decorator node is graphically represented by a rhombus with the label \say{$\delta^R$}, as in Figure~\ref{fig:rdec}. Algorithm~\ref{alg:resource} describes its pseudocode, in particular, for each resource $q$ required by the decorator's child (Line~2), if the resource results assigned to another child (Line~3), then the priority of the child to get the resource $q$ increases according to the function $g$.  
The algorithm then assigns the resources to the child with the highest priority (Lines 7-9) and releases the child's resources if either it no longer requires it (Lines 10-11).

\begin{remark}
We are not interested in a scheduler that fairly assigns the resources as it is done, for example, in the Operating Systems. The designer may implement a fair scheduling policy and encode it in the function $g$.
However, if an action has always higher priority than another one to get a resource, this should be modeled via a Sequence or Fallback composition.
\end{remark}
\vspace*{-1em}


\begin{algorithm2e}[h!]
\SetKwProg{Fn}{Function}{}{}
\Fn{Tick()}
{

\For{\ArgSty{q} \KwSty{in} \ArgSty{child.Q($x_k$)}}{
  \If{(\ArgSty{$\alpha(q)$} \KwSty{not} = \ArgSty{child}) \KwSty{and}  (\ArgSty{$\alpha(q)$} \KwSty{not} \ArgSty{$\emptyset$}) }{
        \ArgSty{child.$\rho(x_k)$} $\gets$ \ArgSty{child.$\rho(x_{k-1})$} + \ArgSty{$g(x_k)$}\\
    \Return{Running}
  }
}
\For{\ArgSty{q} \KwSty{in} \ArgSty{R}}{
\If{\ArgSty{q} \KwSty{in} \ArgSty{child.Q($x_k$)}}{
	\If{$\alpha(q)$ \KwSty{not} = \ArgSty{child} $child.\rho(x_k) > \alpha(q).\rho(x_k)$}{\ArgSty{$\alpha(q)$} $\gets$  \ArgSty{child}}
      
} \ElseIf{$\alpha(q)$ = \ArgSty{child}}{
      \ArgSty{$\alpha(q)$} $\gets$  \ArgSty{$\emptyset$} 
}
}
      \ArgSty{childStatus} $\gets$  \ArgSty{child.Tick()}\\
  \Return{childStatus}
  }
    \caption{Pseudocode of a ResourceSynchronization Decorator.}
 \label{alg:resource}
\end{algorithm2e}

%

\clearpage	

\subsection{Absolute Progress Synchronization}
\label{PM:AS}

A BT achieves an absolute progress synchronization by setting, a-priori, a finite ordered set $\mathcal{B}$ of values for the progress. These values are used as \emph{barriers} at the task level~\cite{taubenfeld2006synchronization}. Whenever a child of an AbsoluteProgressSync Decorator has the progress equal to or greater than a progress barrier in $\mathcal{B}$
it no longer receives ticks until all the other nodes whose parent is an instance of such decorator have the progress equal to or greater than the barrier's value.
We now present a use case example for the absolute progress synchronization, taking inspiration from the literature~\cite{chitta2010planning},~\cite{hern2018boston}.

\begin{figure}[b]
\begin{subfigure}[b]{0.49\columnwidth}
\includegraphics[width=\columnwidth]{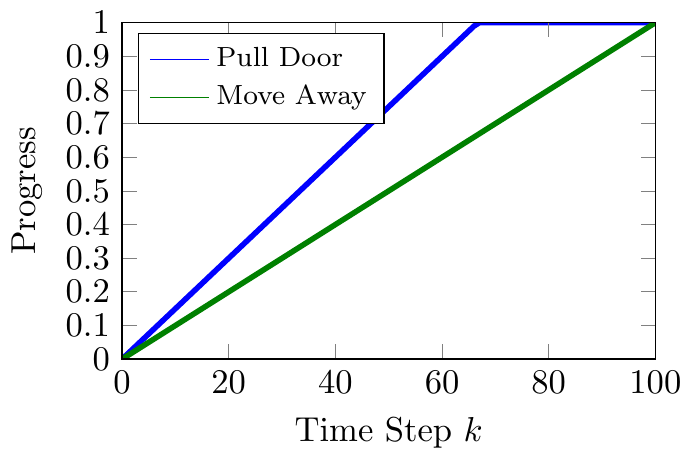}
\caption{Without synchronization.}
\label{fig:ex:absolute:progress:unsync}
\end{subfigure}
\hfill
\begin{subfigure}[b]{0.49\columnwidth}
\includegraphics[width=\columnwidth]{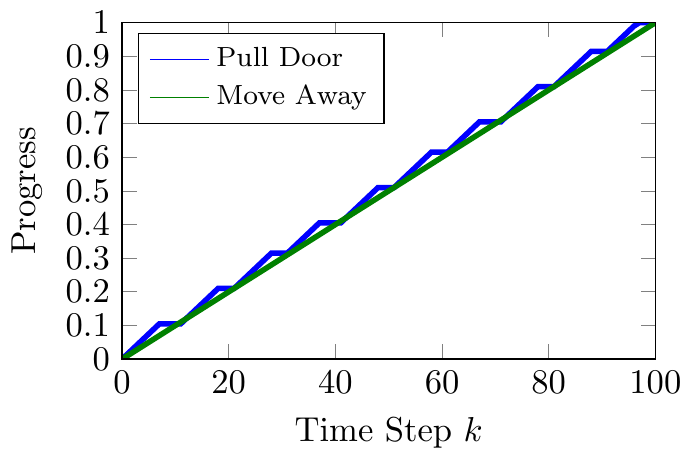}
\caption{With synchronization.}
\label{fig:ex:absolute:progress:sync}
\end{subfigure}
\caption{Progress profiles of the actions of Example~\ref{ex:absolute}.}
\label{fig:ex:absolute:progress}
\end{figure}

\begin{example}[Absolute]
\label{ex:absolute}

\begin{figure}[h!]
\centering
\begin{subfigure}[t]{0.45\columnwidth}
\centering
\includegraphics[width=0.8\columnwidth]{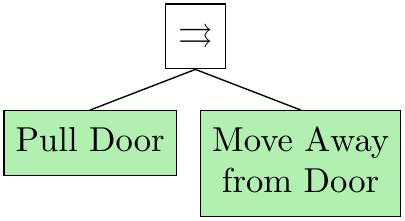}
\caption{Without synchronization.}
\label{fig:ex:absolute:bt:unsync}
\end{subfigure}
\hfill
\begin{subfigure}[t]{0.45\columnwidth}
\centering
\includegraphics[width=0.8\columnwidth]{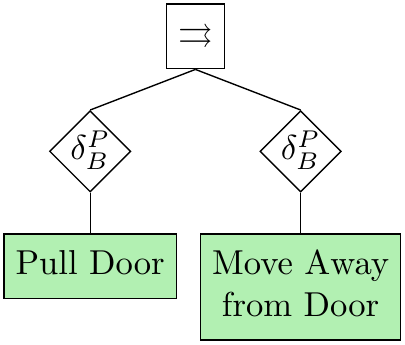}
\caption{With synchronization.}
\label{fig:ex:absolute:bt:sync}
\end{subfigure}

\caption{BT encoding the desired behavior of Example~\ref{ex:absolute}}
\label{fig:ex:absolute:bt}
\end{figure}

A robot has to pull a door open. To accomplish this task, the robot must execute two behaviors concurrently: an arm movement behavior to pull the door open and a base movement behavior to make the robot move away from the door while this opens, as the BT in Figure~\ref{fig:ex:absolute:bt:unsync}. 

The progress profile of the two sub-BTs, \say{Pull Door} $\bt_1$ and \say{Move Away from Door} $\bt_2$, holds the equations below:

\begin{equation}
    p_i(x_k)= 
\begin{cases}
    0 &\text{ if }k = 0\\
    p_i(x_{k-1}) + a_i,              & \text{otherwise}
\end{cases}
\label{ex:absolute:eq:progress}
\end{equation}
with $a_1 = 0.015$ (Pull Door), $a_2 = 0.01$ (Move Away from Door). 
The equations describe a linear progress profile for both behaviors, with the action \say{Pull Door} faster than the action \say{Move Away from Door}.
However, to ensure that the task is correctly executed, the robot must execute the two behaviors above in a synchronized fashion. The BT in Figure~\ref{fig:ex:absolute:bt:sync} encode such synchronized behavior, with the following barriers
\begin{equation}
B = \{0.1, 0.2, 0.3, 0.4, 0.5, 0.6, 0.7, 0.8, 0.9\}
\end{equation}

Figure~\ref{fig:ex:absolute:progress} shows the progress profiles of the actions with and without synchronization. We see how, in the synchronized case, the arm movement keeps stopping to wait for the base movement at the points of executions defined in the barrier.

\end{example}

\newpage
\subsection{Relative Progress Synchronization}
\label{PM:RS}
In this case, synchronization does not follow a common progress indicator, but it is relative to another node's execution. A BT achieves a relative progress synchronization by setting a-priori a threshold value $\Delta \in [0, 1]$. Whenever a child of a RelativeProgressSync Decorator exceeds the minimum progress, among all the other nodes whose parent is an instance such decorator,  by $\Delta$, it no longer receives ticks.

We now present a use case example for the relative progress synchronization, taking inspiration from the literature~\cite{fischer2013impact}. We provide an implementation this example in Section~\ref{sec:experimental}.

\begin{example}[Relative]

\begin{figure}[h!]
\begin{subfigure}[t]{0.45\columnwidth}
\includegraphics[width=0.8\columnwidth]{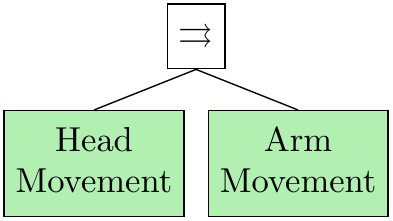}
\caption{Without synchronization.}
\label{fig:ex:relative:bt:unsync}
\end{subfigure}
\hfill
\begin{subfigure}[t]{0.45\columnwidth}
\includegraphics[width=0.8\columnwidth]{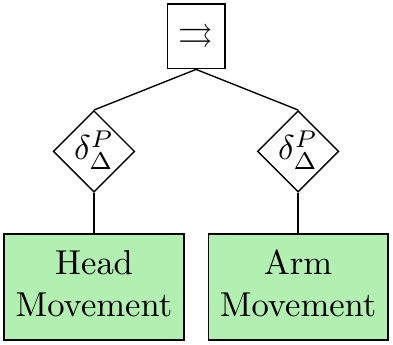}
\caption{With synchronization.}
\label{fig:ex:relative:bt:sync}
\end{subfigure}

\caption{BT encoding the desired behavior of Example~\ref{ex:relative}}
\label{fig:ex:relative:bt}
\end{figure}

\label{ex:relative}
A service robot has to give directions to visitors to a museum. To make the robot's motions look natural, whenever the robot gives a direction, it points with its arm and the head to that direction as in the BT in Figure~\ref{fig:ex:relative:bt:unsync}. 

The progress profile of the two sub-BTs, \say{Head Movement} $\bt_1$ and \say{Arm Movement} $\bt_2$, holds the equations below:

\begin{equation}
    p_i(x_k)= 
\begin{cases}
    0 &\text{ if }k = 0\\
    p_i(x_{k-1}) + a_i,              & \text{otherwise}
\end{cases}
\label{ex:relative:eq:progress}
\end{equation}
with $a_1 = 0.01$ (Arm), $a_2 = 0.05$ (Head). Figure~\ref{fig:ex:relative:progress:sync} shows the progress profile of the sub-BTs.

The equations describe a linear progress profile for both behaviors, with the action \say{Move Head} faster than the action \say{Move Arm}.
However, the arm and head may require different times to perform the motion, according to the direction to point at. Hence, to look natural, the head movement must follow the arm movement to avoid the unnatural behavior where the robot looks first to a direction and then points at it, or the other way round. The BT in Figure~\ref{fig:ex:relative:bt:sync} encode such synchronized behavior, with $\Delta = 0.1$

Figure~\ref{fig:ex:relative:progress} shows the progress profiles of the actions. We see how the head movement stops when its progress surpasses the arm movement's progress by $0.1$, around time step $k=10$.

\begin{figure}[b]
\begin{subfigure}[b]{0.49\columnwidth}
\includegraphics[width=\columnwidth]{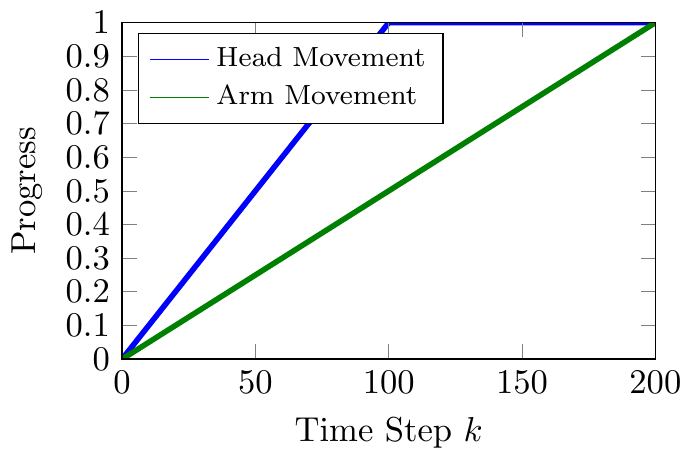}
\caption{Without synchronization.}
\label{fig:ex:relative:progress:unsync}
\end{subfigure}
\hfill
\begin{subfigure}[b]{0.49\columnwidth}
\includegraphics[width=\columnwidth]{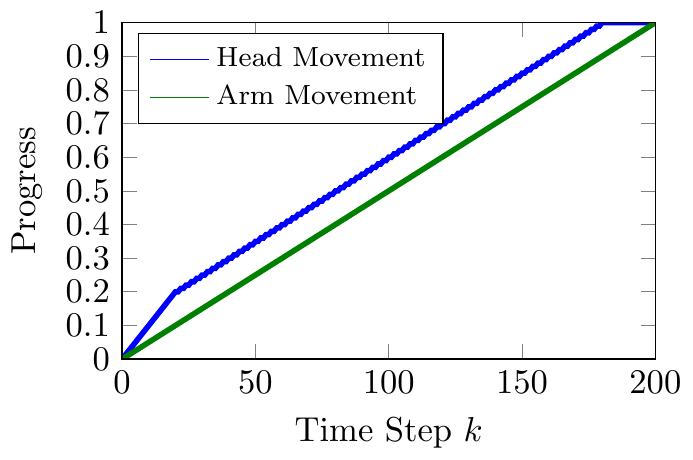}
\caption{With synchronization.}
\label{fig:ex:relative:progress:sync}
\end{subfigure}
\caption{Progress profiles of the actions of Example~\ref{ex:relative}.}
\label{fig:ex:relative:progress}
\end{figure}

\end{example}

\paragraph*{Perpetual Actions}
We can use the relative synchronized parallel node also to impose coordination between \emph{perpetual} actions, (i.e., an action that, even in the ideal case, does not have a fixed duration, hence a progress profile), as in the following example taken from the literature~\cite{rovida2018motion}. We will present an implementation of the example above in Section~\ref{sec:experimental}.


\begin{example}[Perpetual Actions]
\label{ex:perpetual}

\begin{figure}[h!]
\centering
\begin{subfigure}[t]{0.43\columnwidth}
\centering
\includegraphics[width=0.7\columnwidth]{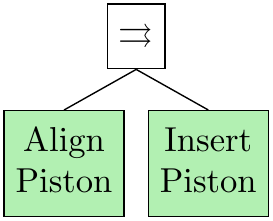}
\caption{Without synchronization.}
\label{fig:ex:perpetual:bt:unsync}
\end{subfigure}
\begin{subfigure}[t]{0.43\columnwidth}
\centering
\includegraphics[width=0.7\columnwidth]{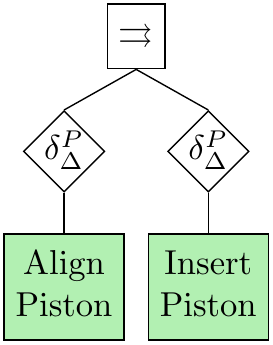}
\caption{With synchronization.}
\label{fig:ex:perpetual:bt:sync}
\end{subfigure}

\caption{BT encoding the desired behavior of Example~\ref{ex:perpetual}}
\label{fig:ex:perpetual:bt}
\end{figure}
An industrial manipulator has to insert a piston into a cylinder of a motor block.
It is an instance of a typical peg-in-the-hole problem with the
additional challenge of the freely swinging piston rod. To correctly insert the piston, the latter must be kept aligned during the insertion into the cylinder.

We can describe this behavior as a parallel BT composition of two sub-BTs: one for inserting the piston and one for keeping the piston aligned with the cylinder as in Figure~\ref{fig:ex:perpetual:bt:unsync}. The Insert Piston action stops when the end-effector senses that the piston hits the cylinder's base, hence its progress cannot be computed beforehand. Since the inserting behavior and the alignment behavior are executed concurrently, the robot may insert the piston too fast, resulting in a collision between the piston and the cylinder's edge.

Figure~\ref{fig:ex:perpetual:bt:sync} shows a BT of a synchronized execution, where the progress of the piston insertion (Insert Piston action) has only two values, $0$ and $1$. It equals $1$ whenever the piston is being inserted and $0$ otherwise. Similarly, for the alignment behavior (Align Piston action). The insertion behavior stops while the robot is aligning the piston. 
\end{example}

\begin{remark}
In real-world scenarios, we  can compute the progress either in an open-loop (i.e., at each tick received it increments the progress) or in a closed-loop (i.e., using the sensors to compute the actual progress) fashion.
\end{remark}
\subsection{Resource Synchronization}
This section shows how CBTs can execute multiple actions in parallel without resource conflicts. 
This synchronization becomes useful when executing in parallel BTs that have some actions in common, as shown in Example~\ref{ex:resource}, adapted from the BT literature. This often happens whenever we want to execute concurrently existing BTs.
\begin{example}
\label{ex:resource}

The BT in Figure~\ref{fig:ex:resource:bt:unsync} shows a BT for a missile evasion tactic, taken from the literature\cite{yao2015adaptive}. The BT has three sub-BTs that run in parallel: \say{Turn on Countermeasure Maneuvers}, \say{Countermeasure Maneuvers once},  \say{Dispense Chaff and Flares Every 10 Seconds}, \say{Turn Clockwise if an Enemy on a Range}.
 
The actions \say{Countermeasure Maneuver} and \say{Turn Clockwise} run in parallel and both use the aircraft's actuation.
There are cases in which both actions receive ticks, resulting in possible conflicts. We can use the resource decorator node to avoid such conflicts, as in the BT in Figure~\ref{fig:ex:resource:bt:sync}.

 The authors of \cite{yao2015adaptive} did not address the concurrency issue above. However, taking advantage of the composability of BTs, we did the modification easily.

\begin{figure}[h!]
\centering
\begin{subfigure}[t]{0.49\columnwidth}
\centering
\includegraphics[width=\columnwidth]{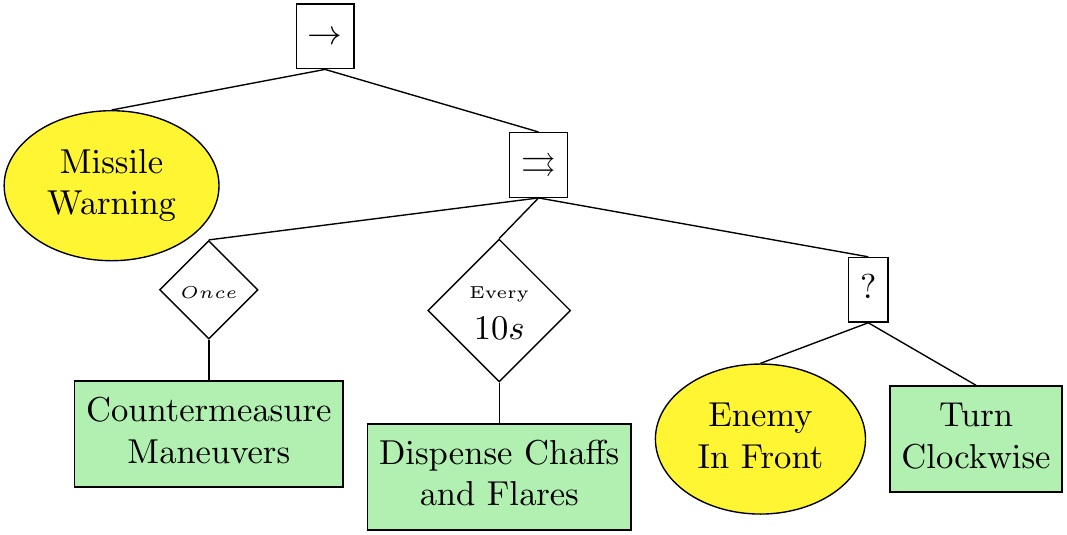}
\caption{Without synchronization.}
\label{fig:ex:resource:bt:unsync}
\end{subfigure}
\begin{subfigure}[t]{0.49\columnwidth}
\centering
\includegraphics[width=\columnwidth]{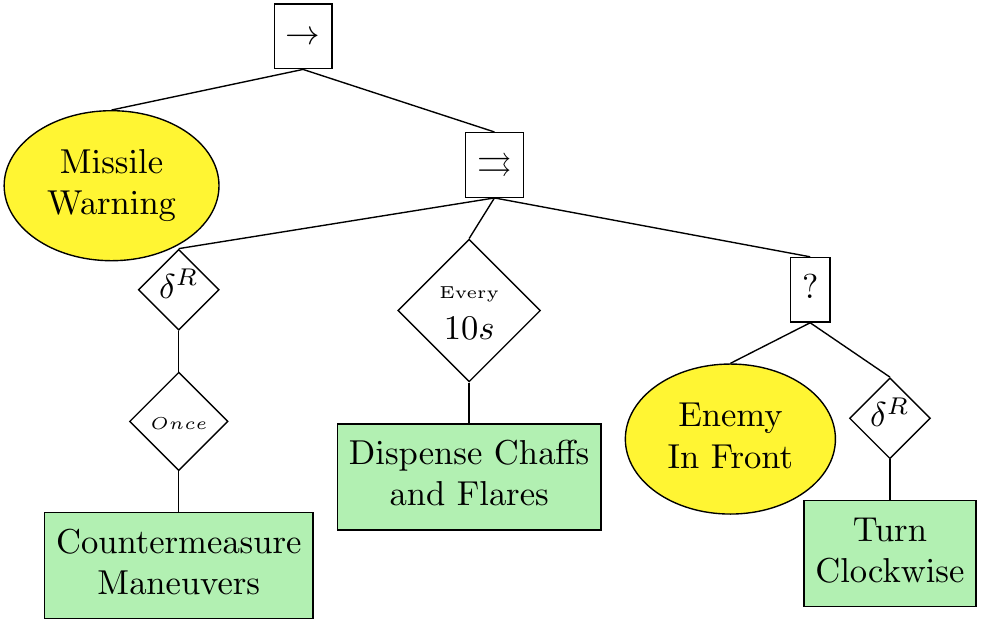}
\caption{With synchronization.}
\label{fig:ex:resource:bt:sync}
\end{subfigure}

\caption{BT encoding the desired behavior of Example~\ref{ex:resource}, adapted from~\cite{yao2015adaptive}}
\label{fig:ex:resource:bt}
\end{figure}

\end{example}

%

\subsection{Improving Predictability}
\label{sec:predictability}
We can use progress synchronization to impose a given progress profile constraint. The idea is to define an artificial action with the desired progress profile (over time) defined a priori and putting it as a child of an absolute synchronized parallel node with the actions whose progress is to be constrained. 
However, since we can only stop actions (i.e., BTs have no means to speed up actions), we can only define such artificial action as progress upper bound. This type of progress profile creation may become very useful at the developing stage since the actions may run at a different speed in the real world and in a simulation environment. Improving predictability reduces the difference between simulated and real-world robot execution.


\begin{example}
\label{ex:predictability}
\begin{figure}[h!]
\centering
\includegraphics[width=0.3\columnwidth]{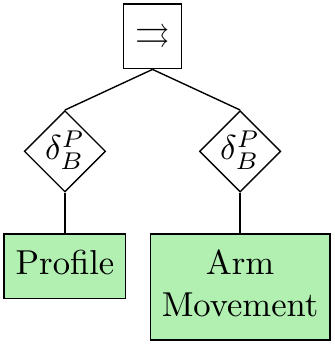}
\caption{BT encoding the desired behavior of Example~\ref{ex:predictability}}
\label{fig:ex:predictability:bt}
\end{figure}

A robot has to move its arm following a sigmoid profile (i.e., the
movements are first slow, then fast, then slow again). However, the manipulation action is designed to follow a linear profile (i.e. same movement's speed throughout
the execution). To impose the desired profile we create the action \say{Profile} that models the sigmoid and we impose a progress synchronization with the manipulation action. The BT in Figure~\ref{fig:ex:predictability:bt} shows the BT to encode this task. 
\end{example}

\begin{figure}[h!]
\begin{subfigure}[b]{0.49\columnwidth}
\includegraphics[width=\columnwidth]{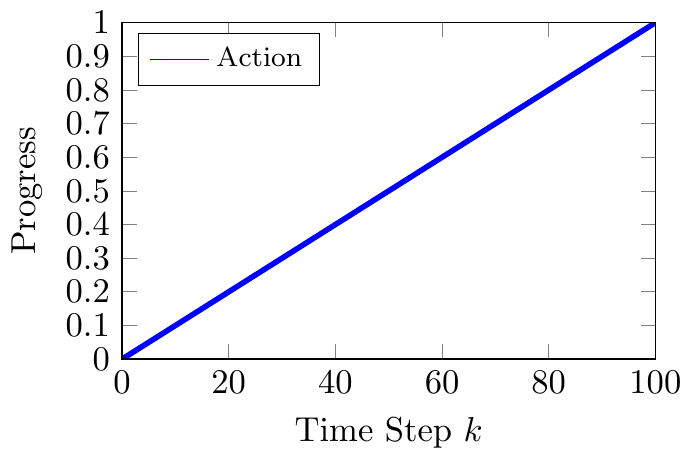}
\label{fig:ex:predictability:progress:unsync}
\vspace*{-1.5em}
\caption{Without Synchronization.}
\end{subfigure}
\begin{subfigure}[b]{0.49\columnwidth}
\includegraphics[width=\columnwidth]{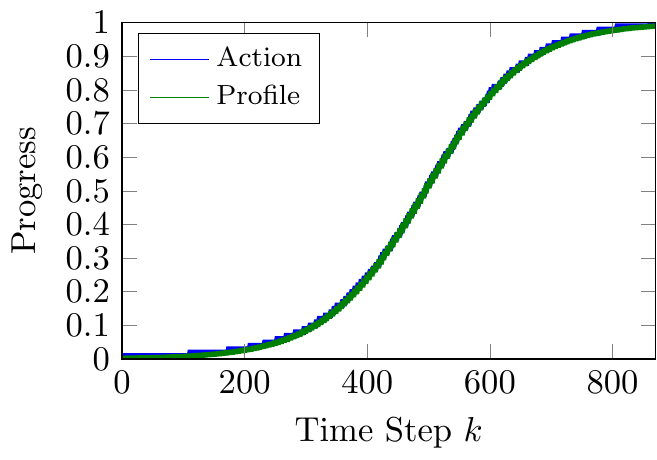}
\label{fig:ex:predictability:progress:sync}
\vspace*{-1.5em}
\caption{With Synchronization.}
\end{subfigure}
\caption{Progress profiles of Example~\ref{ex:predictability} with and without synchronization.}
\label{fig:ex:predictability:progress}
\end{figure}

Figure~\ref{fig:ex:predictability:progress} shows the progress profiles of the action with and without the progress profile imposition. Note how the action's progress profile changed without editing the action. However, this was possible as the action, originally, has a faster progress profile than the desired one, as we have no non-intrusive means to speed up actions.

\vspace*{-1.5em}

%


%
%
%
%

%

\newpage
\section{Synchronization Measures}
\label{sec:measures}
This section presents the second contribution of the paper. We define measures for the concurrent execution of BTs used to establish execution performance. We show measures for both progress synchronization and predictability.  In Section~\ref{sec:experimental} we show how the design choices for relative and absolute parallel nodes affect the performance.

\subsection{Progress Synchronization Distance}

\begin{definition}
\label{PM:def:performance}

Let $N$ be the number of nodes that have as a parent the same instance of a progress decorator node, the progress distance at state $x_k$ is defined as:

\begin{equation}
\pi(x_k) \triangleq \sum_{i = 1} ^N{\sum_{j = 1}^N{\frac{|p_i(x_k) - p_j(x_k)|}{2}}}
\end{equation}
where $p_i\in [0, 1]$ is the progress of the $i$-th child, as in Section~\ref{sec:synchronizaton} above. 
\end{definition}

We sum the progress difference for each pair of nodes that have as parent the same instance of a progress decorator node. We divide by $2$ to avoid double count the differences. We use the absolute difference instead of a, e.g., squared difference to assign equal weight to the spread of the progresses.

Intuitively, a small progress distance results in high performance for both relative and absolute progress synchronization.

\subsection{Predictability Distance}
\label{pm.subsec:timeline}

A useful method to measure predictability is to set the desired progress value and compute the average variation from the expected and the true time instant in which the action has a progress that is closest to the desired one. We can use this measure to assess the deviation from the ideal execution.

\begin{definition}
\label{def:pred}
Given a progress value $\bar p \in [0,1]$, a time step $
\tilde t_k \triangleq \argmin_{t_k}(p(x(t_k)) - \bar p)$, and $\hat t_k$ be the time instance when $p(x(t_k))$ is expected to be equal to $\bar p$. The time predictability distance relative to progress $\bar p$ is defined as:
\begin{equation}
P(\bar p) \triangleq |\tilde t_k - \hat t_k|
\end{equation}
\end{definition}
\begin{remark}
A node may not obtain the exact desired progress value as the progress may be defined at discrete points of execution. Hence in Definition~\ref{def:pred} we compute the difference between the desired progress value ad the closest one obtained.
\end{remark}

\newpage
\section{Experimental Validation}
\label{sec:experimental}

We conducted numerical experiments
that allow us to collect statistically significant data to study how the design choices affect the performance measures defined in Section~\ref{sec:measures} and to compare our approach against other solutions. We made the source code available online for reproducibility.\footnote{\url{https://github.com/miccol/tro2021-code}} We also conducted
experiments on real robots to show the applicability of our
approach in the real world. We made available online a
video of these experiments.\footnote{\url{https://youtu.be/zCBuTYogb_U}}
\subsection{Numerical Experiments}

We are ready to show how the number of barriers in $\mathcal{B}$ (for absolute synchronization) and the threshold value $\Delta$ (for relative synchronization) affect the performance, computed using the measures defined in Section \ref{sec:measures}. For illustrative purposes, we define custom-made actions with different progress profiles. To collect statistically significant data, we ran the BT of each experiment 10000 times; we use boxplots to compactly show the minimum, the maximum, the median, and the interquartile range of the measures proposed. Each experiment starts with the progress of actions equal to $0$ and ends when all the actions progress reach $1$.

\paragraph*{How the number of progress barriers affects the performance of absolute synchronization}
We now present an experiment that highlights how the number of progress barriers in $\mathcal{B}$  affects the performance of absolute synchronization.

\begin{figure}[h!]
\centering
\begin{subfigure}{0.45\columnwidth}
\centering
\includegraphics[width=0.6\columnwidth]{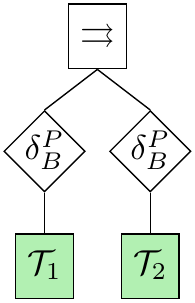}
\caption{Experiments~\ref{PM.ex.dummy}}
\label{fig:ne:progress:bt:absolute}
\end{subfigure}
\begin{subfigure}{0.45\columnwidth}
\centering
\includegraphics[width=0.6\columnwidth]{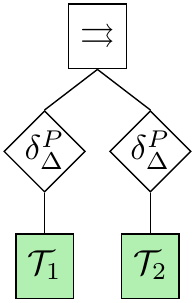}
\caption{Experiments~\ref{PM.ex.dummy.delta}}
\label{fig:ne:progress:bt:relative}

\end{subfigure}
\caption{BT used for Experiments~\ref{PM.ex.dummy} and ~\ref{PM.ex.dummy.delta}.}
\label{fig:ne:progress:bt}
\end{figure}

\begin{experiment}
\label{PM.ex.dummy}
Consider the BT in Figure~\ref{fig:ne:progress:bt:absolute} where the progress decorator implements an absolute synchronization with equidistant barriers (i.e., a barrier at each $\frac{1}{|B|}$ progress) and the sub-BTs are such that the progress profile of each $\bt_i$ holds Equation~\eqref{eq:ne:barriers:progress} below:
\begin{equation}
    p_i(x_k)= 
\begin{cases}
    0 &\text{ if }k = 0\\
    p_i(x_{k-1}) + a_i + \omega_i(x_k),              & \text{otherwise}
\end{cases}
\label{eq:ne:barriers:progress}
\end{equation}

with $a_1 = 0.03$, $a_2= 0.02 $, and $\omega_i(x_k) \in [-\bar \omega, \bar \omega]$ a random number, sampled from an uniform distribution, in the interval $[-\bar \omega, \bar \omega]$.

The model above describes an action whose progress evolves linearly with a fixed value ($a_i$) and with some disturbance ($\omega_i(x_k)$), modeling possible uncertainties in the execution that affect the progress. 

\begin{figure}[h!]
\centering
\includegraphics[width=0.7\columnwidth]{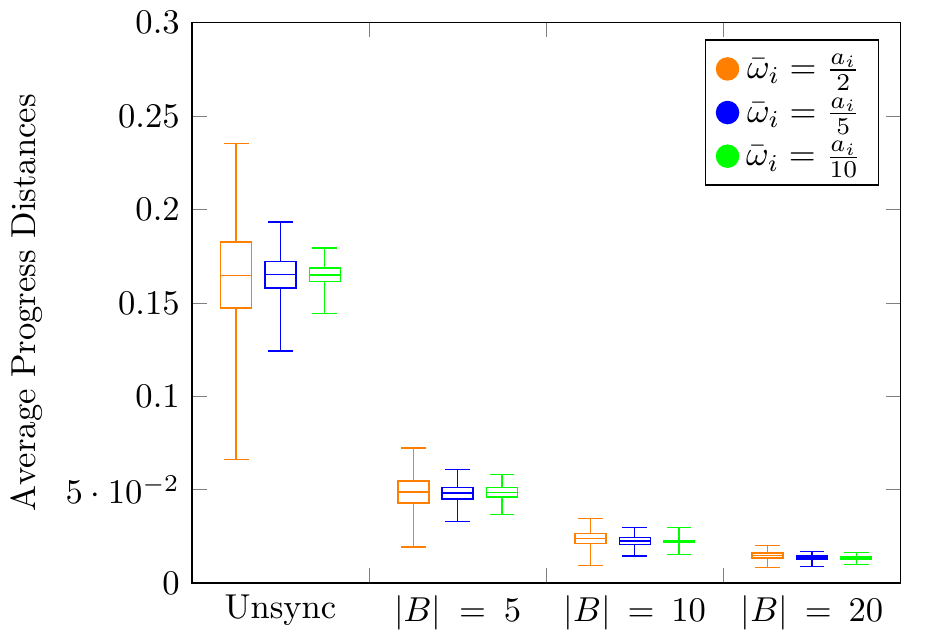}
\caption{
Boxplot of the progress distances of Experiment~\ref{PM.ex.dummy} with different numbers of barriers $|\mathcal{B}|$ and different values of $\bar \omega$. $|\mathcal{B}|= 0$ corresponds to the unsynchronized execution.}
\label{fig:analysis:absolute}
\end{figure}

Figure~\ref{fig:analysis:absolute} shows the results of running  10000 times the BT in Experiment~\ref{PM.ex.dummy} in different settings and computing the average progress distance throughout the execution. We observe better performance with a large number of barriers and smaller  $\bar \omega$. This shows that a higher number of barriers prevents the progress of the actions to differ from each other (see Algorithm~\ref{alg:progress}). 
Note also that the synchronization yields a reduced variance even with large values of $\omega$.

\end{experiment}

\begin{remark}
In Experiment~\ref{PM.ex.dummy}, we consider equidistant progress barriers. We expect similar results with non-equidistant barriers, except for the corner case in which all the barriers are agglomerated in a specific progress value.
\end{remark}

\paragraph*{How the threshold value affects the performance of relative synchronization}
We now present an experiment that highlights how the value of $\Delta$ affects the performance of relative synchronization.

\begin{experiment}
\label{PM.ex.dummy.delta}
Consider the BT in Figure~\ref{fig:ne:progress:bt:relative} where the progress decorator implements a relative synchronization and the sub-BTs are such that the progress profile of each $\bt_i$ holds Equation~\eqref{eq:ne:barriers:progress:bis} below.

\begin{equation}
    p_i(x_k)= 
\begin{cases}
    0 &\text{ if }k = 0\\
    p_i(x_{k-1}) + a_i + \omega_i(x_k),              & \text{otherwise}
\end{cases}
\label{eq:ne:barriers:progress:bis}
\end{equation}

with $a_1 = 0.03$, $a_2= 0.02 $, and $\omega_i(x_k) \in [-\bar \omega, \bar \omega]$ a random number, sampled from an uniform distribution, in the interval $[-\bar \omega, \bar \omega]$.

The model above describes an action whose progress evolves linearly with a fixed value ($a_i$) and with some disturbance ($\omega_i(x_k)$), modeling possible uncertainties in the execution that affect the progress. 

\vspace*{1em}
\begin{figure}[h!]
\centering
\includegraphics[width=0.7\columnwidth]{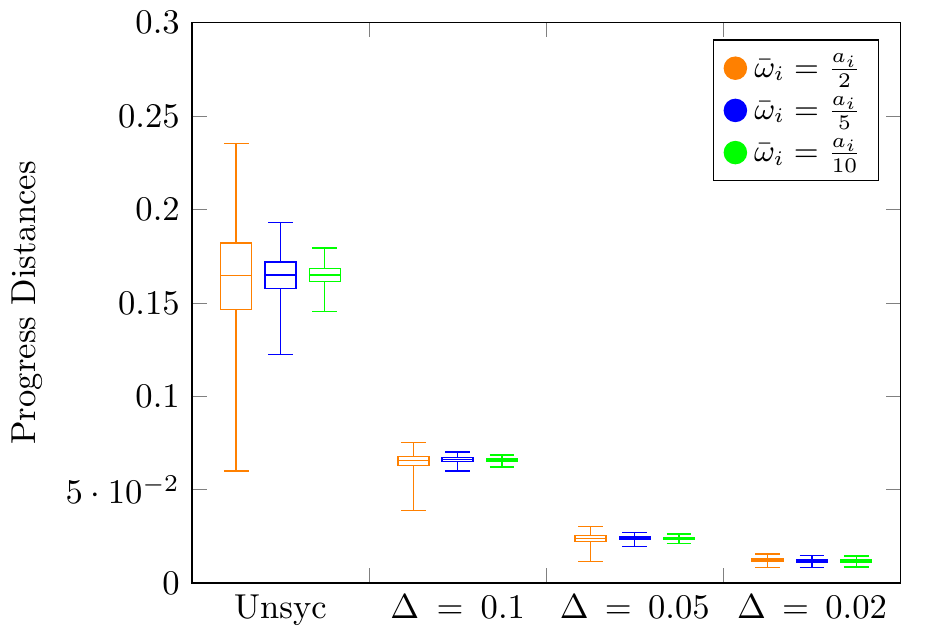}
\caption{Boxplot of the progress distances of Experiment~\ref{PM.ex.dummy.delta} with different values for $\Delta$ and $\bar \omega$. $\Delta = 1$ corresponds to the unsynchronized execution.}
\label{fig:analysis:relative}
\end{figure}

Figure~\ref{fig:analysis:relative} shows the results of running  10000 times the BT in Experiment~\ref{PM.ex.dummy.delta} in different settings and computing the average progress distance throughout the execution.
 We observe that the performance increases with a smaller $\Delta$ and decreases with a larger $\bar \omega$. This shows that a smaller $\Delta$ prevents the progress of the actions to differ from each other (see Algorithm~\ref{alg:progress}s). Note also that the synchronization yields a reduced variance even with large values of $\omega$.

\end{experiment}

\begin{remark}
The synchronization may deteriorate other desired qualities. For example, since actions are waiting for one another, the overall execution may be slower than the slowest action. Moreover, a small value for $\Delta$  or a larger number of barriers can result in highly intermittent behaviors.
\end{remark}

\begin{remark}

The decorators can be placed in different parts of the BT and not as direct children of a parallel node, as shown in Section~\ref{sec:related}.
\end{remark}

\begin{remark}
A single action that performs both tasks represents a better synchronized solution. However, for reusability purposes or for the separation of concerns, the designer may want to implement the behavior using two separated actions.
\end{remark}

\begin{figure}[h!]
\centering
\includegraphics[width=1\columnwidth]{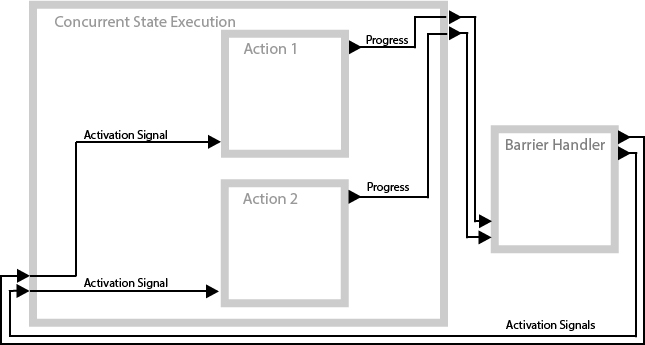}
\caption{Concurrent RAFCon state machine for Experiments~\ref{PM.ex.comparison.abs} and \ref{PM.ex.comparison.rel}. \emph{Action 1} and \emph{Action 2} increase the progress if and only the activation signal (their input) equals $1$.
The \emph{Concurrent State Execution}, which is a RAFCon \texttt{concurrency-state} and executes the two sub-states (\emph{Action 1} and \emph{Action 2}) concurrently. The \emph{Barrier Handler} computes the activation signals for the actions (i.e., it is set to $0$ if the action's progress surpasses the current barrier (for absolute synchronization) or the other action's progress by $\Delta$ (for relative synchronization); it is set to $1$ otherwise)}
\label{fig:comparison:refcon}
\end{figure}

\paragraph*{Progress synchronization comparison}
We now present an experiment that compares the synchronization performance. We compare our approach with three different alternative ones: One using elements from the C++11's standard library\footnote{\url{https://en.cppreference.com/w/cpp/thread/barrier}}, as it is the programming language used in the BT library; and one using the DLR's RMC Advanced Flow Control (RAFCon)~\cite{brunner2016rafcon}, as it is a tool to develop concurrent robotic tasks using hierarchical state machine with an intuitive graphical user interface, addressing similar issues of BTs.\footnote{Both implementations are available at \url{github.com/miccol/TRO2021-code}}
Figure~\ref{fig:comparison:refcon} shows the concurrent state machine developed in Rafcon for the Experiments~\ref{PM.ex.comparison.abs} and \ref{PM.ex.comparison.rel}.

\newpage

\begin{experiment}
\label{PM.ex.comparison.abs}

Consider the BT in Figure~\ref{fig:ne:progress:bt:absolute} where the progress decorator implements an absolute synchronization with equidistant barriers (i.e., a barrier at each $\frac{1}{|B|}$ progress) and the sub-BTs are such that the progress profile of each $\bt_i$ holds Equation~\eqref{eq:ne:barriers:progress} with $\bar{\omega}= 0.015$.

The model above describes the same BT used in Experiment~\ref{PM.ex.dummy} with the given value for $\bar{\omega}$.

\begin{figure}[h!]
\centering
\includegraphics[width=0.7\columnwidth]{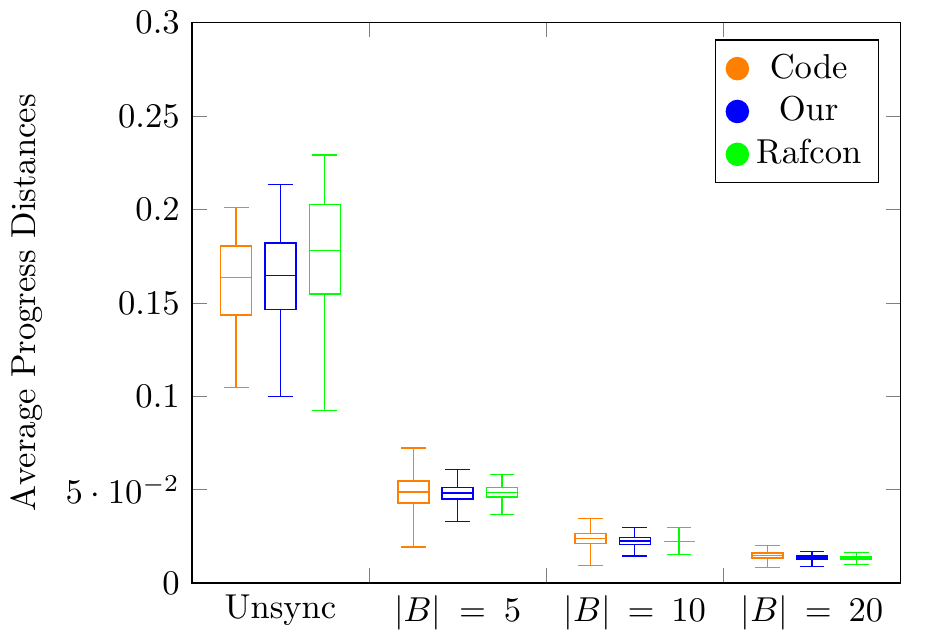}
\caption{Boxplot of the progress distances of Experiment~\ref{PM.ex.comparison.abs} with different numbers of barriers $|\mathcal{B}|$ for each method. $|\mathcal{B}|= 0$ corresponds to the unsynchronized execution. We compare the performance obtained using C++ primitives (Code), the proposed approach (Our), and the one Rafcon (Rafcon).}
\label{fig:analysis:absolute:comparison}
\end{figure}

Figure~\ref{fig:analysis:absolute:comparison} shows the results of running  10000 times the BT in Experiment~\ref{PM.ex.dummy} with the different approaches. 
Note that the unsynchronized setting (e.g., $|\mathcal{B}|= 0$) yields similar values for the different approaches. Hence the boilerplate code of the approach does not affect the performance.

\end{experiment}

\begin{experiment}
\label{PM.ex.comparison.rel}

Consider the BT in Figure~\ref{fig:ne:progress:bt:relative} where the progress decorator implements a relative synchronization and the sub-BTs are such that the progress profile of each $\bt_i$ holds Equation~\eqref{eq:ne:barriers:progress} with $\bar{\omega}= 0.015$.

The model above describes the same BT used in Experiment~\ref{PM.ex.dummy.delta} with the given value for $\bar{\omega}$.

\begin{figure}[h!]
\centering
\includegraphics[width=0.7\columnwidth]{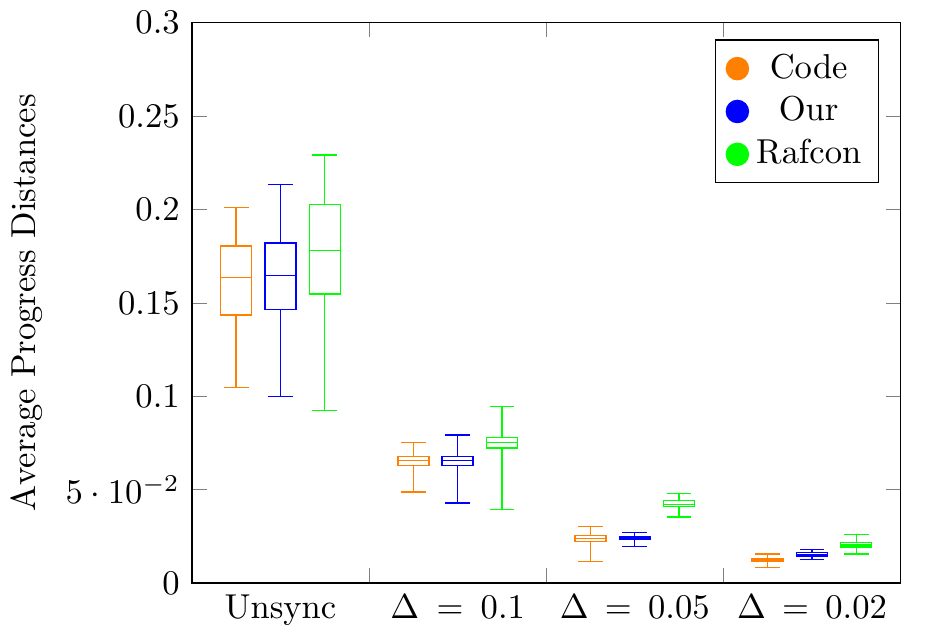}
\caption{Boxplot of the progress distances of Experiment~\ref{PM.ex.comparison.rel} with different values of $\Delta$ for each method. $\Delta = 1$ corresponds to the unsynchronized execution.}
\label{fig:comparison:relative}
\end{figure}

Figure~\ref{fig:comparison:relative} shows the results of running  10000 times the BT in Experiment~\ref{PM.ex.comparison.rel} with the different approaches. We make the same observation of the previous experiment.
Note that, as in the previous experiment, the unsynchronized setting (e.g., $\Delta= 1$) yields similar values.

\end{experiment}
\begin{remark}
Our solution keeps the same order of magnitude as the computer programming ones (the most efficient from a computation point of view) and outperforms the one of Rafcon, while also keeping the advantages of BT over state machines described in the literature~\cite{BTBook}.
\end{remark}

\paragraph*{How the number of children to synchronize affects the performance}
We now present two experiments that show how the approach scales with the number of children.
\begin{experiment}
\label{ex:number:absolute}

Consider a set of BTs that describe the  absolute progress synchronization of a different numbers of actions (Figure~\ref{fig:ne:progress:bt:absolute} shows an example of such BT with two actions). In each BT, the  actions' progress holds Equation~\eqref{eq:ne:barriers:progress}, with $\alpha = 0.03$ and $\bar \omega = 0.015$.

\begin{figure}[h!]
\centering
\includegraphics[width=0.7\columnwidth]{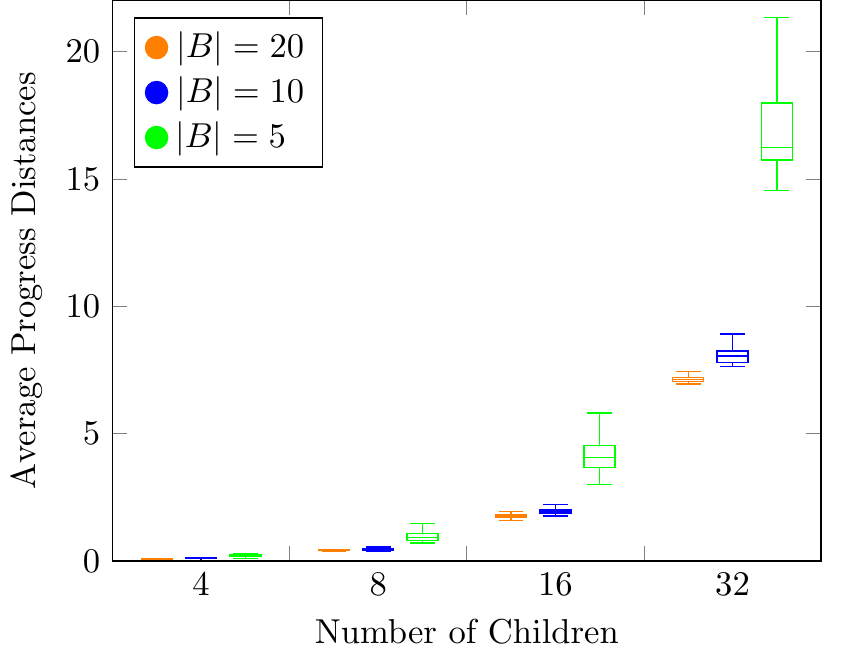}
\caption{Boxplot of the predictability distances of Experiment~\ref{PM.ex.dummy.timeline} with different values for $\Delta$ and $\bar \omega$. $\Delta= 1$ corresponds to the unsynchronized execution.}
\label{fig:analysis:number}
\end{figure}
Figure~\ref{fig:analysis:number} shows the results of running  10000 times the BT in Experiment~\ref{ex:number:absolute} with different numbers of children. We note how the performance decays linearly with the number of children (the number of children increases exponentially in the figure).

\end{experiment}

\begin{experiment}
\label{ex:number:relative}

Consider a set of BTs that describe the  absolute progress synchronization of different numbers of actions (Figure~\ref{fig:ne:progress:bt:absolute} shows an examples of such BT with two actions). In each BT, the  actions' progress holds Equation~\eqref{eq:ne:barriers:progress}, with $\alpha = 0.03$ and $\bar \omega = 0.015$.

\begin{figure}[h!]
\centering
\includegraphics[width=0.7\columnwidth]{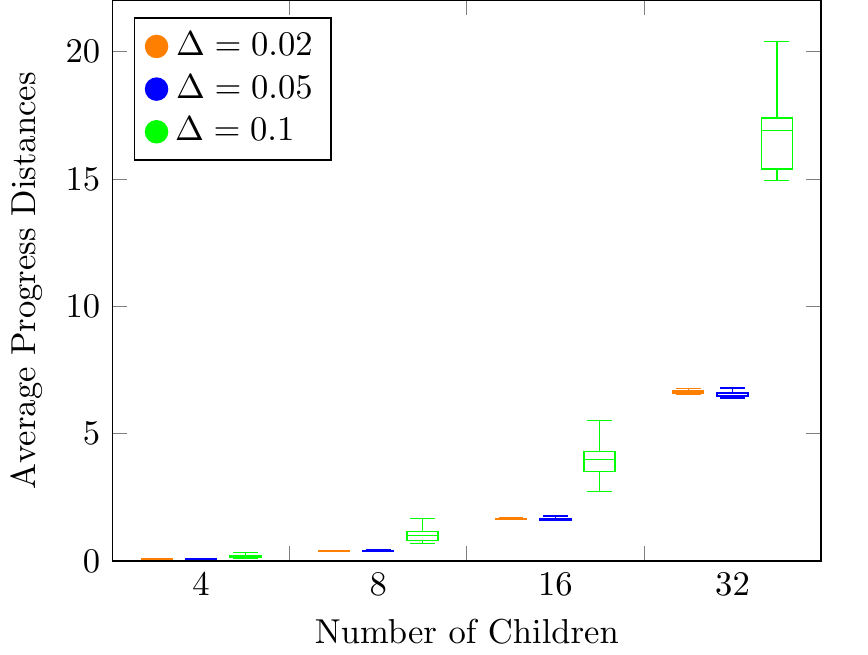}
\caption{Boxplot for predictability distance of Experiment~\ref{PM.ex.dummy.timeline} with different values for $\Delta$ and $\bar \omega$. $\Delta= 1$ corresponds to the unsynchronized execution.}
\label{fig:analysis:number:relative}
\end{figure}

Figure~\ref{fig:analysis:number:relative} shows the results of running  10000 times the BT in Experiment~\ref{ex:number:absolute} with different numbers of children. Similar to the previous experiment. The performance decays linearly with the number of children.

\end{experiment}
As expected, in both absolute and relative synchronization settings, the number of children deteriorates the progress synchronization performance. This is due to the fact that the children's progresses surpass one another, increasing the progress distance.

\newpage
\paragraph*{How the number of barriers affects the predictability}
We now present an experiment that highlights how the number of barriers for an absolute synchronized parallel node affects the predictability of an execution.
\begin{experiment}
\label{PM.ex.dummy.timeline}

Consider the BT of Figure~\ref{fig:ex:predictability:bt} where the progress decorator implements a relative synchronization and the action \say{Arm Movement}, whose progress is to be imposed, has a progress defined such that it holds Equation~\eqref{SA:ex:jitter:action} below:
\begin{equation}
    p_2(x_k)=  
\begin{cases}
    0 &\text{ if }k = 0\\
    p_2(x_{k-1}) + 2 + \omega_i(x_k),              & \text{otherwise}
\end{cases} 
\label{SA:ex:jitter:action}
\end{equation}
whereas the progress of the  action \say{Profile} holds Equation~\eqref{SA:ex:jitter:model} below:

\begin{equation}
    p_1(x_k)= 
\begin{cases}
    0 &\text{ if }k = 0\\
    p_1(x_{k-1}) + 0.1,              & \text{otherwise}
\end{cases}
\label{SA:ex:jitter:model}
\end{equation}

\begin{figure}[h!]
\centering
\includegraphics[width=0.7\columnwidth]{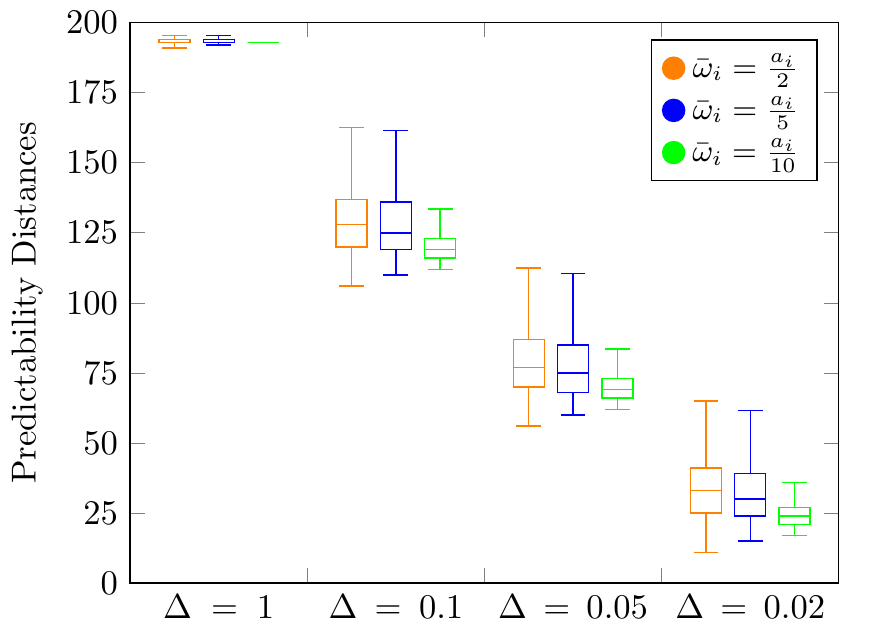}
\caption{Boxplot for predictability distance of Experiment~\ref{PM.ex.dummy.timeline} with different values for $\Delta$ and $\bar \omega$. $\Delta= 1$ corresponds to the unsynchronized execution.}
\label{fig:analysis:predictability}
\end{figure}

Figure~\ref{fig:analysis:predictability} reports the results of Experiment~\ref{PM.ex.dummy.timeline}. We observe worse performance with larger $\bar \omega$ and $\Delta$. 
\end{experiment}

\begin{remark}
In the experiments above, we showed how a designer could synchronize the progress of several subtrees in a non-invasive fashion. The designer can tune the number of
barriers for the absolute synchronization and the threshold value for the relative synchronization.
However, as mentioned above, synchronizations between actions may deteriorate other performances. Figure~\ref{ex:remark:time} shows the average times to complete the executions for Experiment~\ref{PM.ex.comparison.rel}. Similar results were found for absolute progress synchronization.

\begin{figure}[h!]
\centering
\includegraphics[width=0.7\columnwidth]{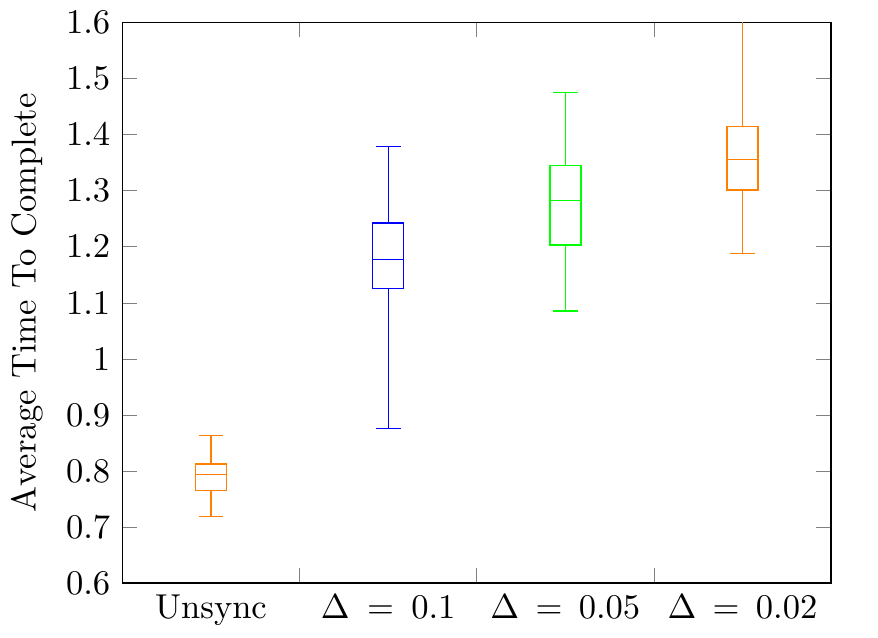}
\caption{Boxplot for time to complete.}
\label{ex:remark:time}
\end{figure}

\end{remark}

\newpage

\paragraph*{How the priority increment function $g$ affects the execution in resource synchronization}

We now present two examples of resource synchronization and how the shape of the increment function leads to different behaviors. As mentioned above, this synchronization is done among subtrees that have equal priority in accessing the resources. The function $g$ provides the designer a way to shape their resource allocation strategy.

\begin{figure}[h!]
\centering
\begin{subfigure}[b]{0.49\columnwidth}
\includegraphics[width=\columnwidth] {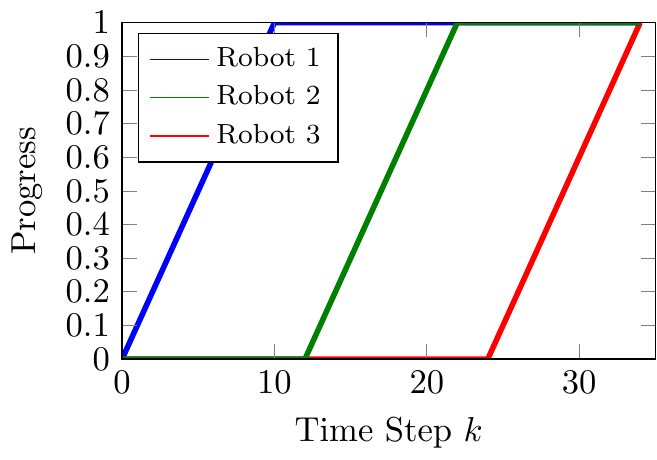}
\caption{Experiment~\ref{ex:greedy}.}
\label{fig:ex:resource:greedy}
\end{subfigure}
\begin{subfigure}[b]{0.49\columnwidth}
\includegraphics[width=\columnwidth] {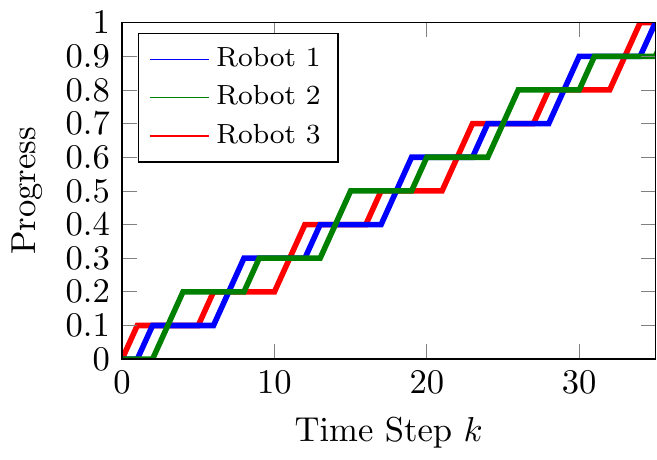}
\caption{Experiment~\ref{ex:fair}.}
\label{fig:ex:resource:fair}
\end{subfigure}
\label{fig:ex:resource}
\caption{Progress profiles of the BTs of Experiments~\ref{ex:greedy} and~\ref{ex:fair}.}

\end{figure}

Experiment~\ref{ex:greedy} and~\ref{ex:fair} show an example of usage of the resource synchronization decorator with different settings for the function $g$.
\begin{experiment}[Greedy Dining Robots]
\label{ex:greedy}

This experiment is the Dining Philosopher Problem \cite{tanenbaum2015modern} with a twist.
Consider three robots that sit in a round table with three cables: Cable $A$, Cable $B$, and Cable $C$. Each cable sits between two robots such that the Robot $1$ can grab Cable $A$ and $B$, Robot $2$ can grab Cable $B$ and $C$, and Robot $3$ can grab Cable $C$ and $A$.
Each robot needs two cables to charge its battery. 
This example represents those cases in which several software components (controlling different robots or different parts on the same robot) need to access a shared resource.

\begin{figure}[h!]
\centering
\includegraphics[width=0.5\columnwidth]{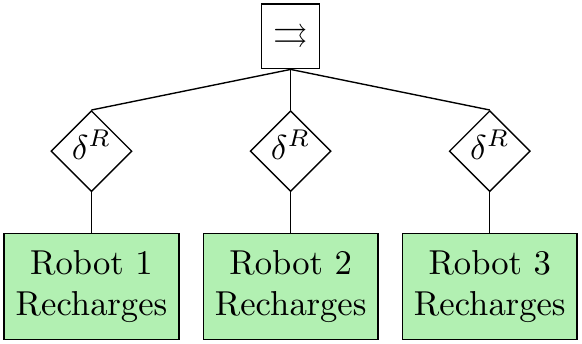}
\caption{BT encoding the desired behavior of Experiment~\ref{ex:greedy}}
\label{fig:ex:greedy:bt}
\end{figure}

The BT in Figure~\ref{fig:ex:greedy:bt} encodes a behavior of these robots that ensures resource synchronization. At each tick, the action \say{Robot $i$ Recharges} increases the battery level by 10\% of its full capacity.
The progress profile follows the battery level as follows:

\begin{equation}
    p_i(x_k)= 
\begin{cases}
    0 &\text{ if }k = 0\\
    p_i(x_{k-1}) + 0.1,              & \text{otherwise}
\end{cases}
\end{equation}
%
$\bt_1$, $\bt_2$, and $\bt_3$ are such that  

\begin{equation}
Q_1(x_k) = 
\begin{cases}
\{A, B\}  &\text{ if } p(x_k) < 1 \\
\emptyset  &\text{ otherwise }
\end{cases}
\end{equation}

\begin{equation}
Q_2(x_k) = 
\begin{cases}
\{B, C\}  &\text{ if } p(x_k) < 1 \\
\emptyset  &\text{ otherwise }
\end{cases}
\end{equation}

\begin{equation}
Q_3(x_k) = 
\begin{cases}
\{C, A\}  &\text{ if } p(x_k) < 1 \\
\emptyset  &\text{ otherwise }
\end{cases}
\end{equation}

The $g$ function are defined as follows:
\begin{eqnarray}
g_i(x_k) &= 0
\end{eqnarray}
That is, the priority does not change when the action does not receive ticks.

Figure~\ref{fig:ex:resource:greedy} shows the progress profile of the two BT. We see how, once a robot acquires the two wires, the wires are assigned to that robot until it no longer requires it (i.e., the battery is fully charged).

\end{experiment}
 \newpage
\begin{experiment}[Fair Dining Robots]
\label{ex:fair}

Consider the three robot of the Experiment~\ref{ex:greedy} above, with the difference in the  definitions of the
$g$ functions:
\begin{eqnarray}
g_i(x_k) &= 1
\end{eqnarray}
That is, the priority increases when the action does not receive ticks.

Figure~\ref{fig:ex:resource:fair} shows the progress profile of the two BT. We see how the wires are allocated in a \say{fair} fashion.

\end{experiment}

The two experiments above show that the choice of the function $g$ becomes crucial to avoid starvation. In Section~\ref{sec:analysis} we will prove under which circumstances the BT execution avoids starvation. Note that by tuning $g_i(x_k)$, we can achieve different profiles of execution, equivalent to the assignment of a quantum of time received by each robot when they get access to the shared resource.

\subsection{Real World Validation}
This section presents the experimental validation implemented on real robots. The literature inspired our experiments.

\paragraph*{Progress Synchronization}
 In Experiment~\ref{ex:icub}, below we present an implementation of Example~\ref{ex:relative} above, motivated by the impact of contingent behaviors on  the quality of
verbal human-robot interaction~\cite{fischer2013impact}.
\begin{experiment}[iCub Robot]
\label{ex:icub}
An iCub robot~\cite{metta2008icub} has to look and point to a given direction.
Figure~\ref{fig:ex:icub:plot} shows the progress plots for the two actions in the synchronized and unsynchronized case.
 Figure~\ref{fig:ex:icub:exec} shows the progress profiles of the two actions using the iCub Action Rendering Engine\footnote{\url{https://robotology.github.io/robotology-documentation/doc/html/group__actionsRenderingEngine.html}}, where we send concurrently the command to look and point at the same coordinate;  and the ones using the BT in Figure~\ref{fig:ex:bt:icub} with a relative synchronization and a threshold value of $\Delta= 0.1$, where the actions look and point performs small steps towards the desired coordinate.

\begin{figure}[h!]
\centering
\includegraphics[width=0.7\columnwidth]{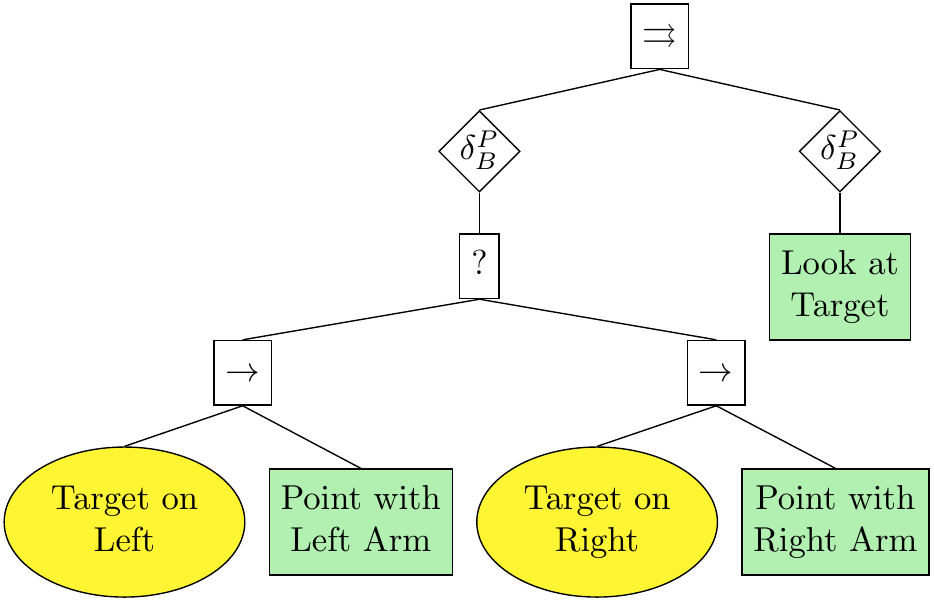}
\caption{BT encoding the behavior of Experiment~\ref{ex:icub}}
\label{fig:ex:bt:icub}
\end{figure}

\begin{figure}[h]
\begin{subfigure}[t]{0.49\columnwidth}
\includegraphics[width=\columnwidth,trim={15cm 5cm 10cm 9cm},clip]{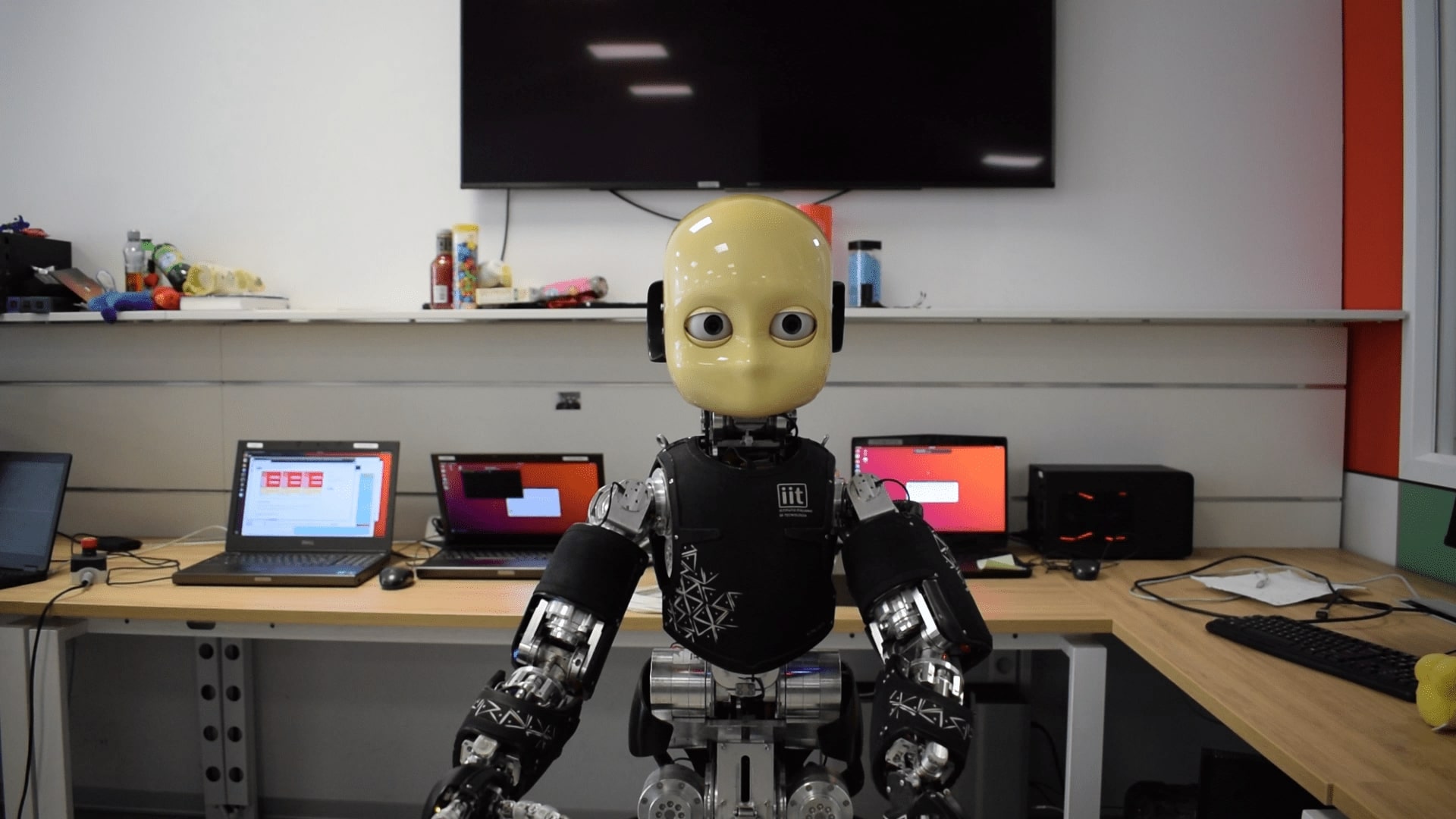}
\caption{[Unsync] Initial State.}
\end{subfigure}
\begin{subfigure}[t]{0.49\columnwidth}
\includegraphics[width=\columnwidth,trim={15cm 5cm 10cm 9cm},clip]{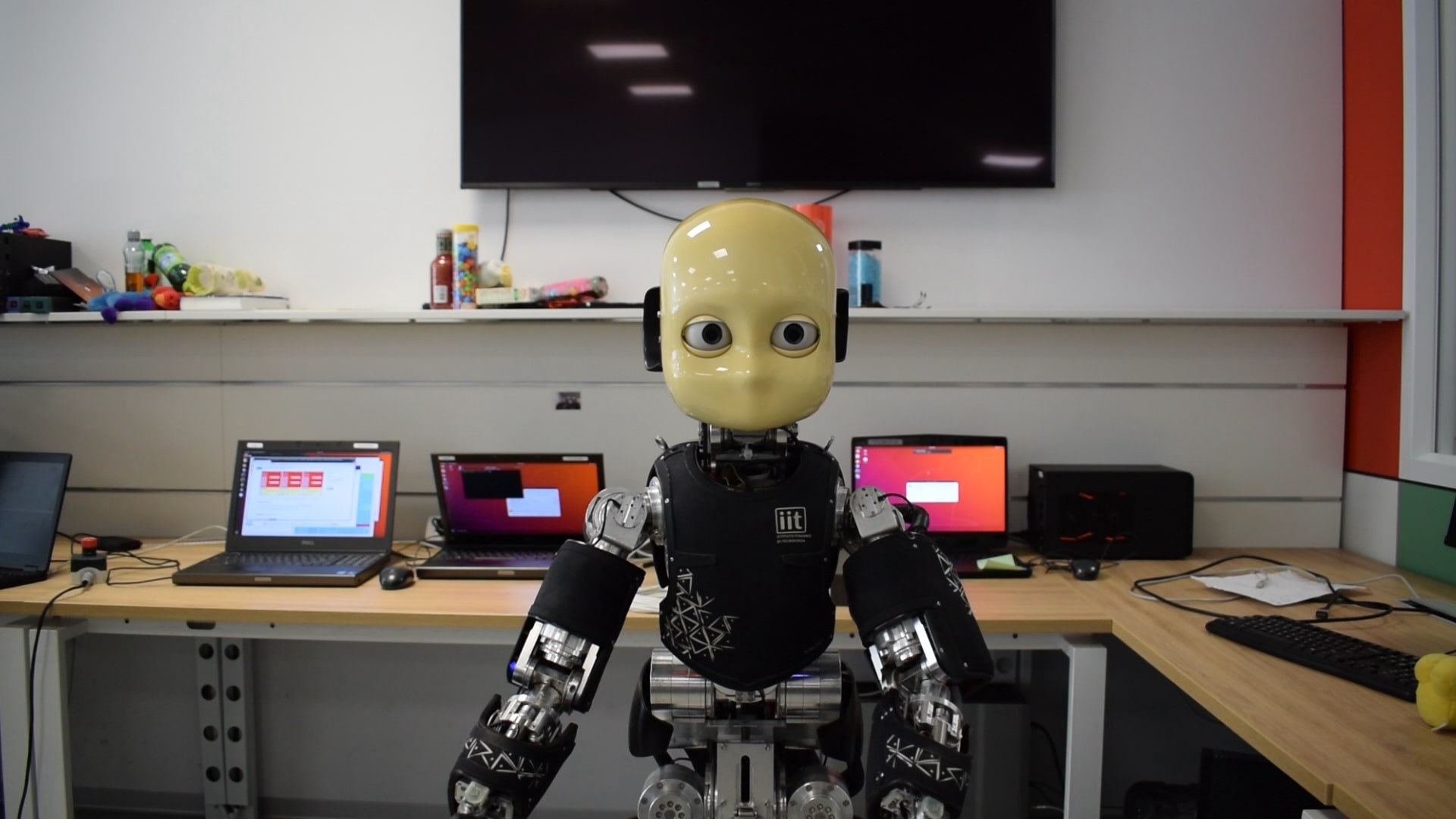}
\caption{[Sync] Initial State.}
\end{subfigure}

\vspace*{0.5em}

\begin{subfigure}[t]{0.49\columnwidth}
\includegraphics[width=\columnwidth,trim={15cm 5cm 10cm 9cm},clip]{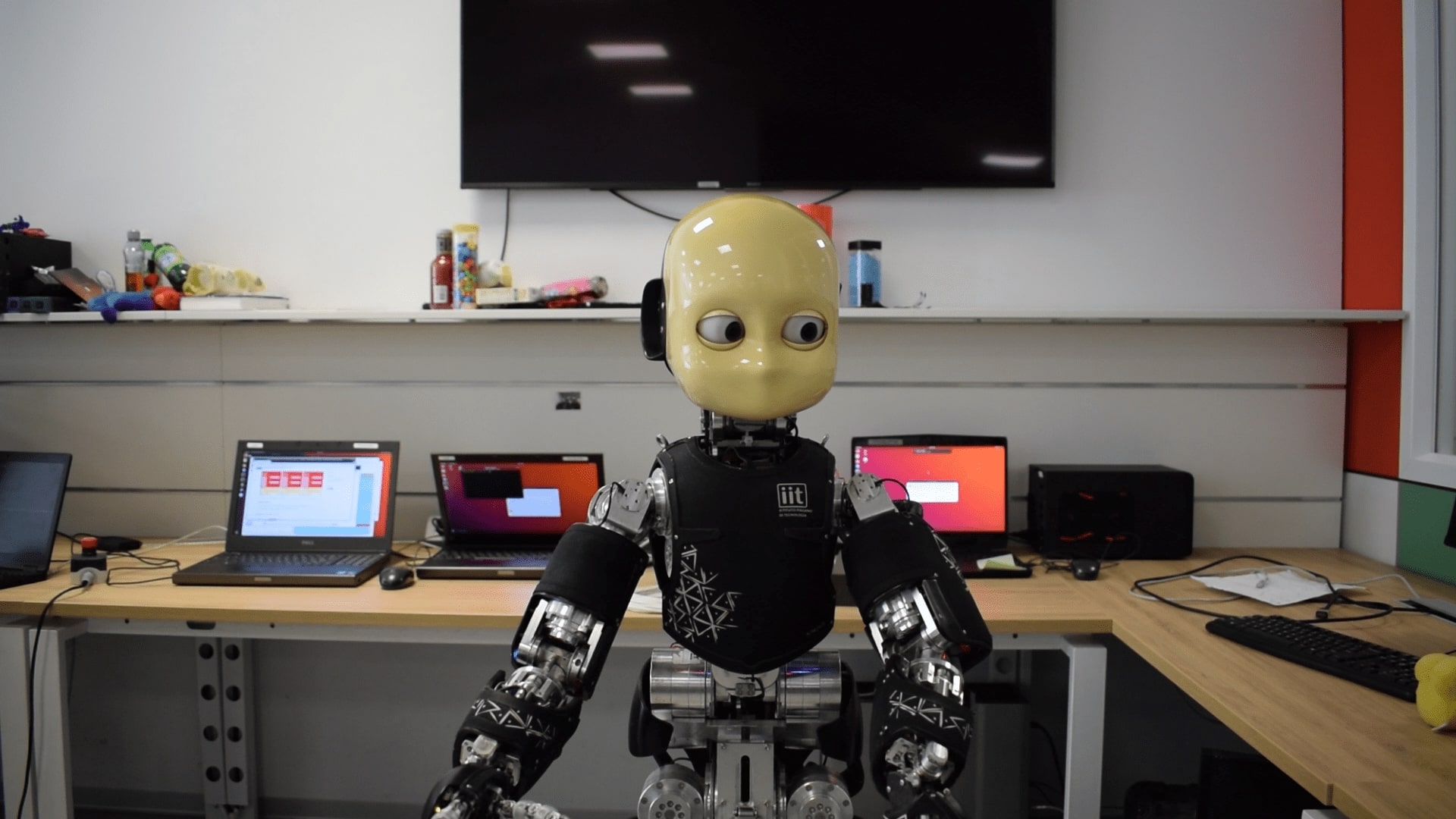}
\caption{[Unsync] The robot starts moving the head.}
\end{subfigure}
\begin{subfigure}[t]{0.49\columnwidth}
\includegraphics[width=\columnwidth,trim={15cm 5cm 10cm 9cm},clip]{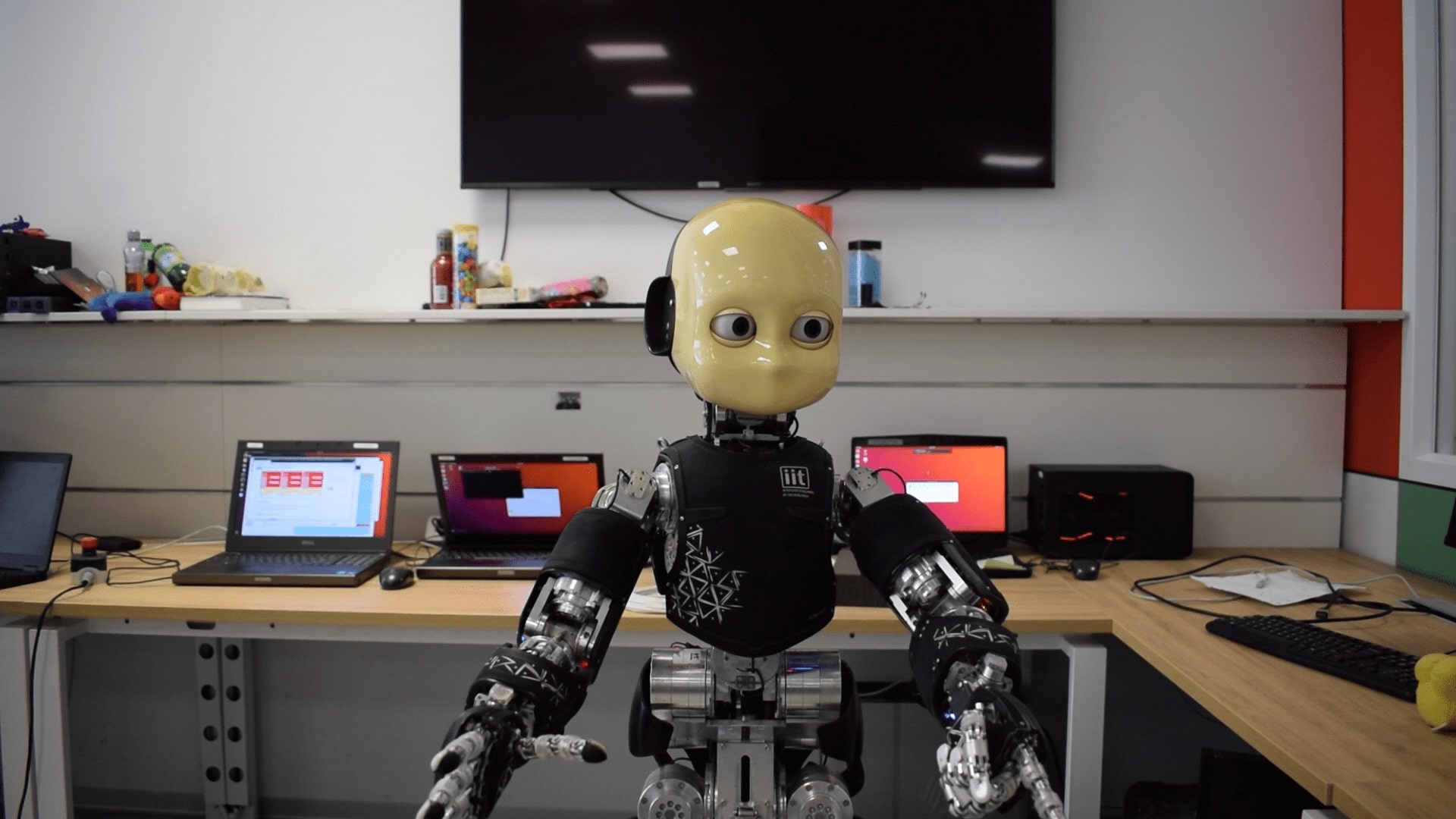}
\caption{[Sync] The robot moves head and arm slowly.}
\end{subfigure}

\vspace*{0.5em}

\begin{subfigure}[t]{0.49\columnwidth}
\includegraphics[width=\columnwidth,trim={15cm 5cm 10cm 9cm},clip]{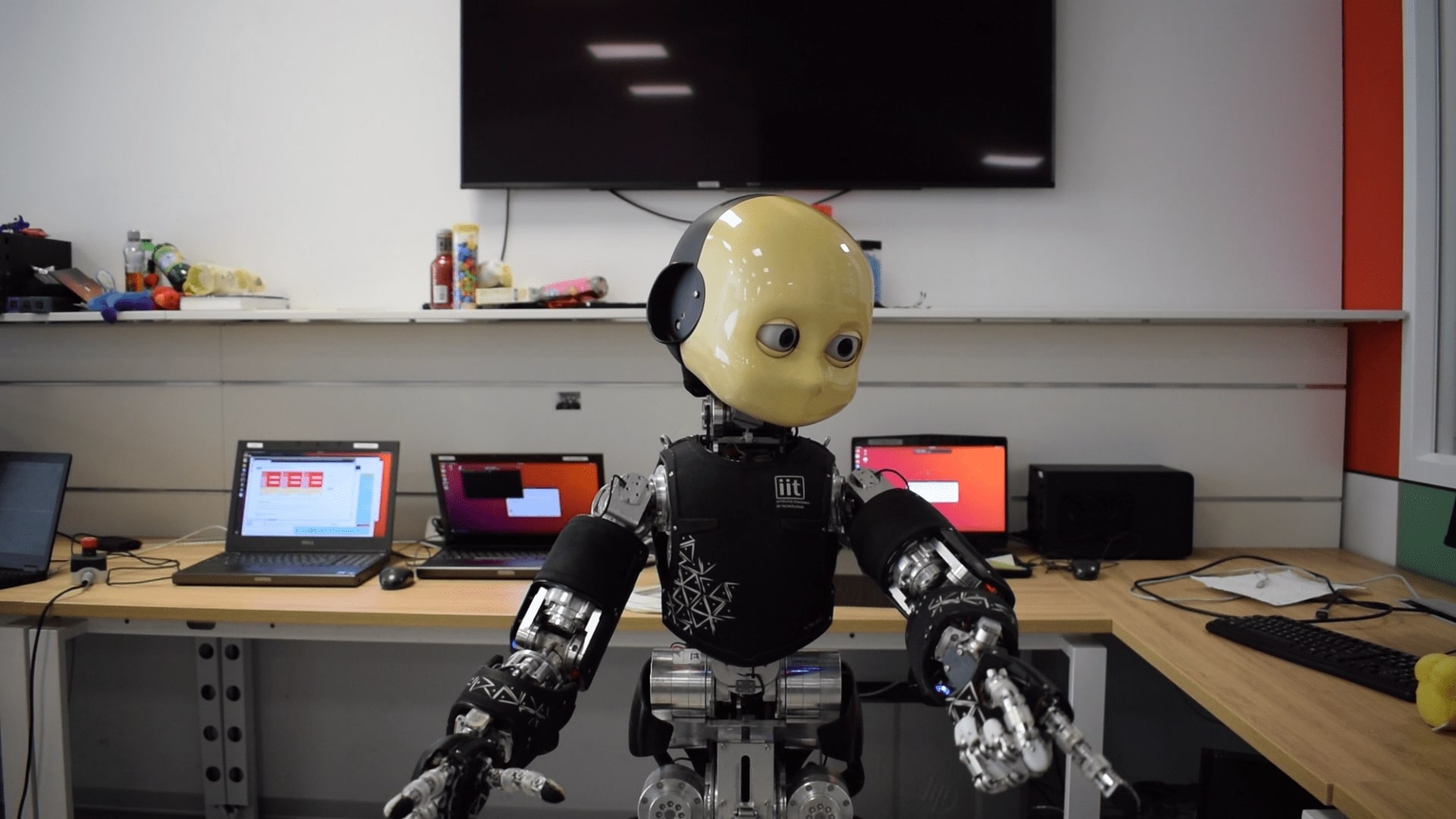}
\caption{[Unsync] The robot finishes to move the head while the arm is still moving.}
\end{subfigure}
\begin{subfigure}[t]{0.49\columnwidth}
\includegraphics[width=\columnwidth,trim={15cm 5cm 10cm 9cm},clip]{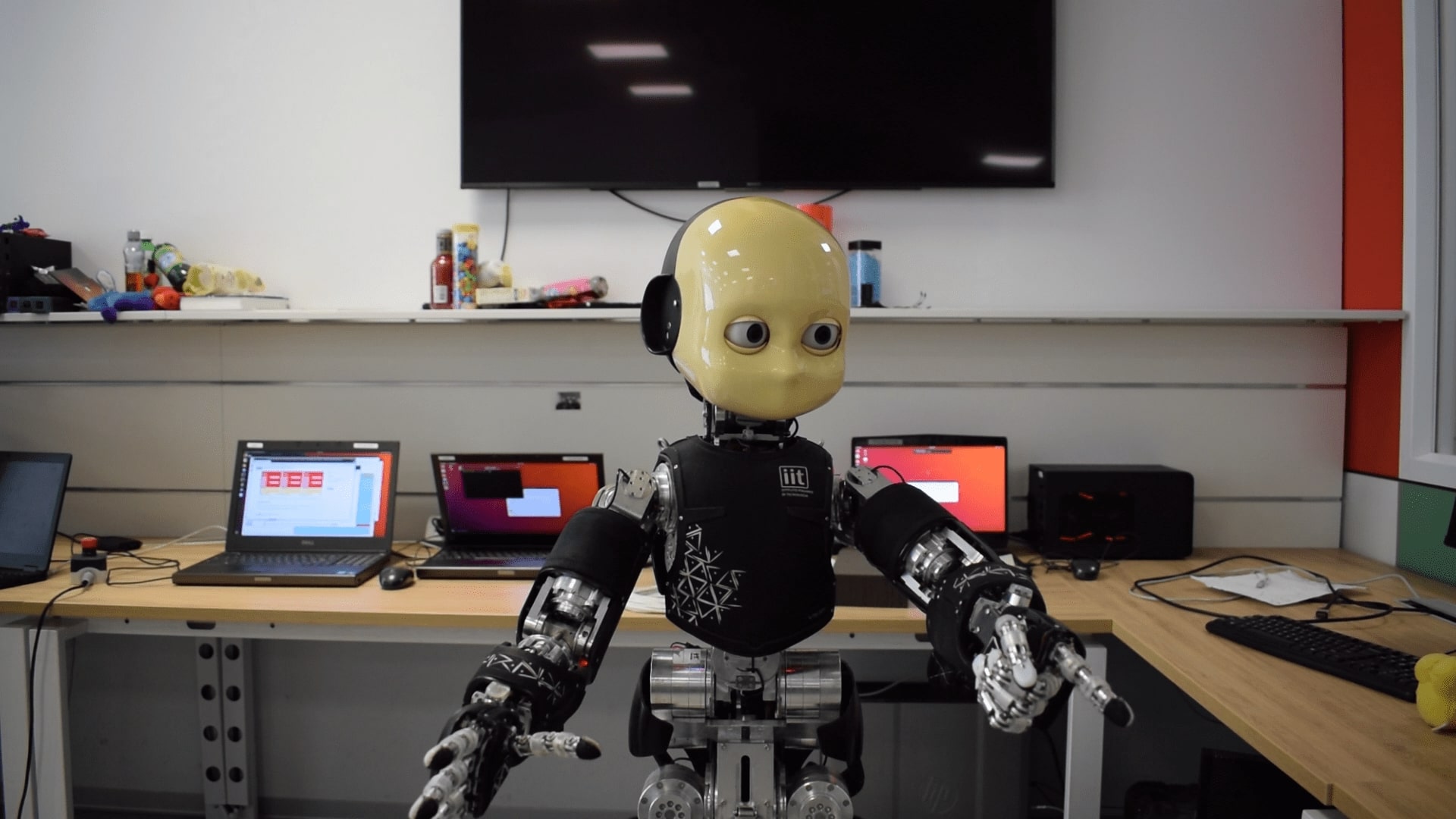}
\caption{[Sync]  The robot keeps moving head and arm slowly.}
\end{subfigure}

\vspace*{0.5em}

\begin{subfigure}[t]{0.49\columnwidth}
\includegraphics[width=\columnwidth,trim={15cm 5cm 10cm 9cm},clip]{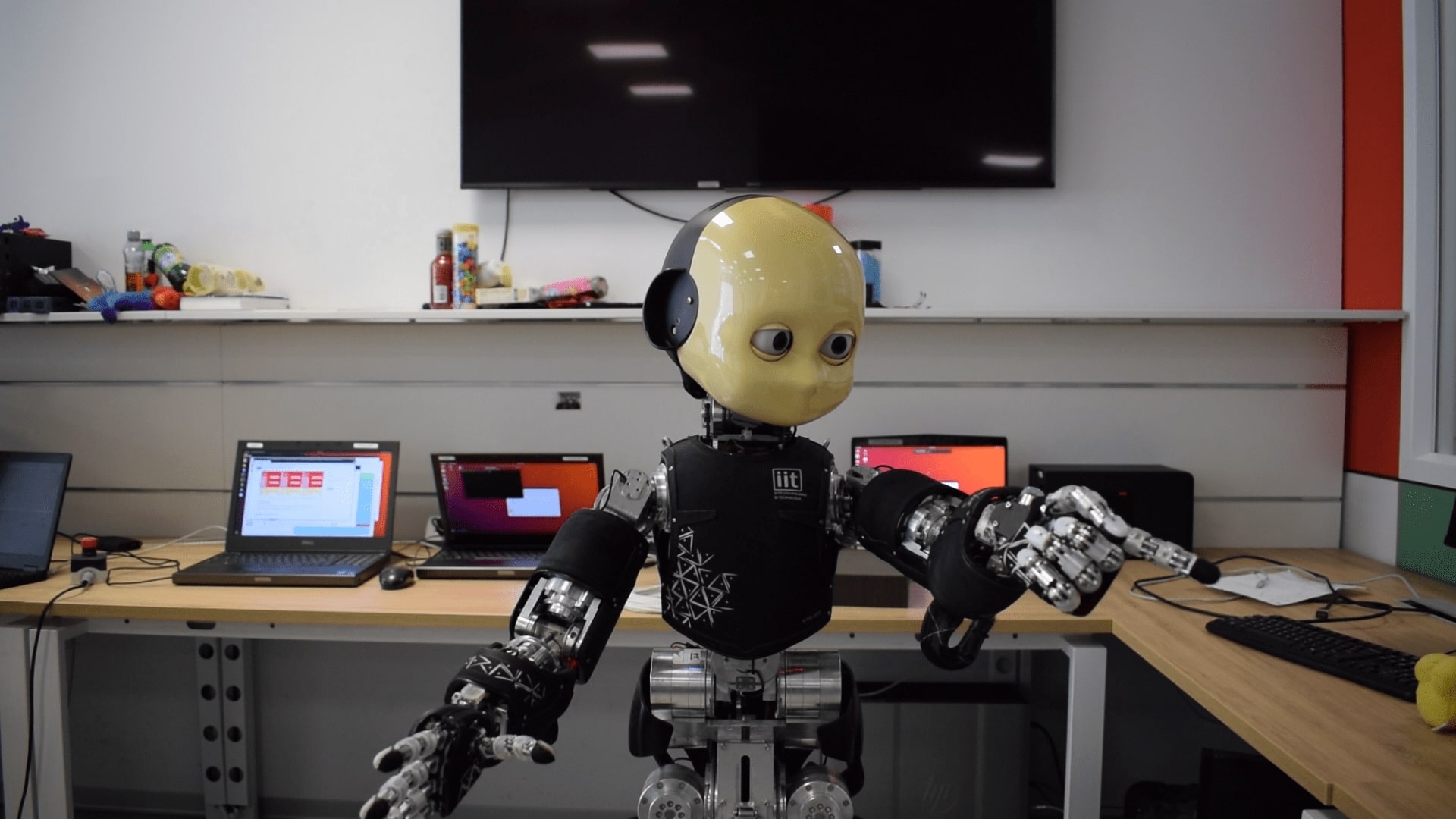}
\caption{[Unsync] The robot finishes to move the arm.}
\end{subfigure}
\begin{subfigure}[t]{0.49\columnwidth}
\includegraphics[width=\columnwidth,trim={15cm 5cm 10cm 9cm},clip]{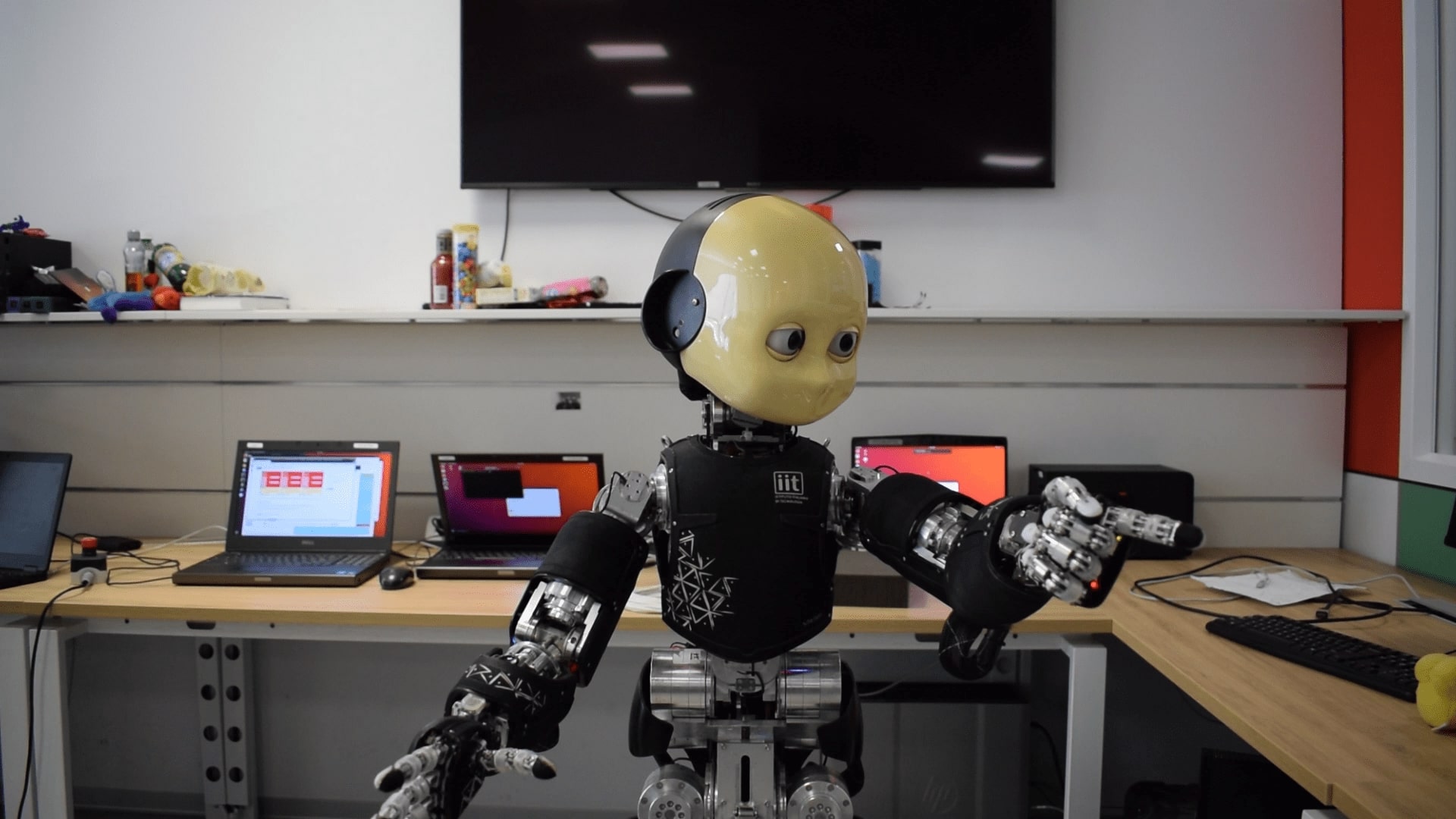}
\caption{[Sync] The robot finishes to move the head and arm.}
\end{subfigure}

\caption{Execution steps of Experiment~\ref{ex:icub} with (right) and without (left) synchronization.}
\label{fig:ex:icub:exec}
\end{figure} 

\vspace*{-0.5em}
\begin{figure}[h!]
\centering
\begin{subfigure}[t]{0.49\columnwidth}
\includegraphics[width=\columnwidth]{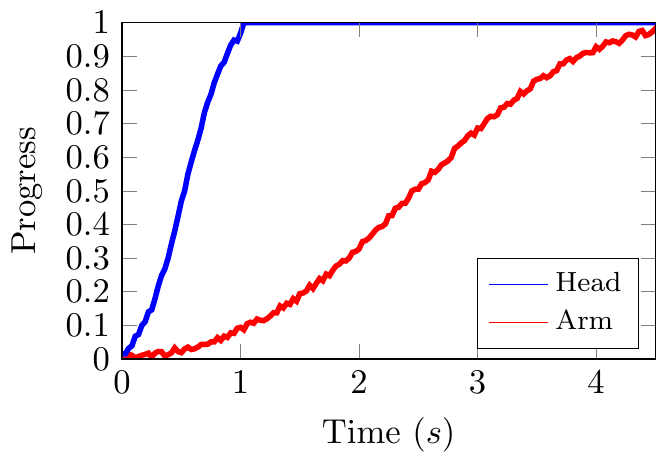}
\end{subfigure}
\begin{subfigure}[t]{0.49\columnwidth}
\includegraphics[width=\columnwidth]{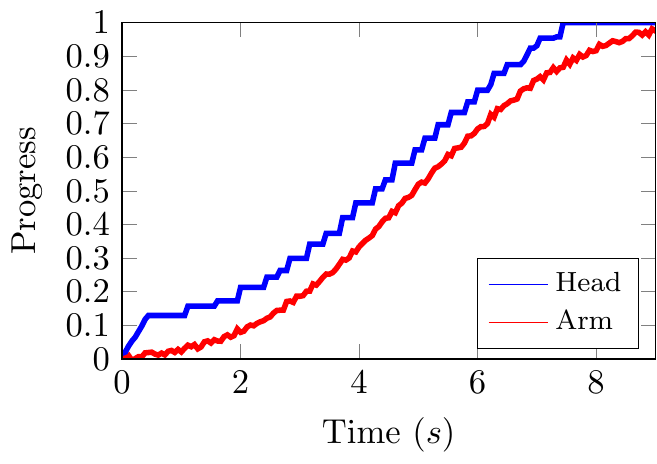}
\end{subfigure}

\caption{Progress values for Experiment~\ref{ex:icub} with (right) and without (left) synchronization}
\label{fig:ex:icub:plot}

\end{figure}

\end{experiment}

The unsynchronized execution looks unnatural as the head moves way faster than the arm. The synchronization allowed a reduction of the average progress distance from $0.4176$ to $0.0964$. From the plot in Figure~\ref{fig:ex:icub:plot} we note how, with the synchronized execution, the head stops as soon as its progress surpasses the one of the arm by $0.1$. Then it moves slower.
Moreover, the synchronized execution completes the task in about double the time. This is due to the fact that smaller movements in iCub are performed slowly, and the synchronized execution breaks down the action in small steps.

\clearpage

\paragraph*{Resource Synchronization}

We now present the use case of a resource synchronization mechanism.
We took a BT used for a use case for an Integrated Technical Partner
European Horizon H2020 project RobMosys\footnote{ \url{https://scope-robmosys.github.io/}} and then we used the resource synchronization mechanism to parallelize some tasks executed sequentially using the classical formulation of BTs.
 As noted in the BT literature~\cite{colledanchise2016advantages, brunner2019autonomous} turning a sequential behavior execution into a concurrent one becomes much simpler in BTs compared to FSMs. 

\begin{experiment}[R1 Robot]
\label{ex:r1}

An R1 robot~\cite{parmiggiani2017design} has to pick up an object from the user's hand and then navigate towards a predefined destination. 
To grasp the object from the user's hand, the robot has to put the arm in a pre-grasp position, extend its hand\footnote{The arm has a prismatic joint in the wrist.}, grasp the object, and finally, retract the hand. 

The BT in Figure~\ref{fig:ex:r1:unsync:bt} encodes the behavior of the robot designed using the classical BT nodes and Figure~\ref{fig:ex:r1:unsync:exec} shows some execution steps, as done in the original project above. 

\begin{figure}[h]
\centering
\includegraphics[width=0.8\columnwidth]{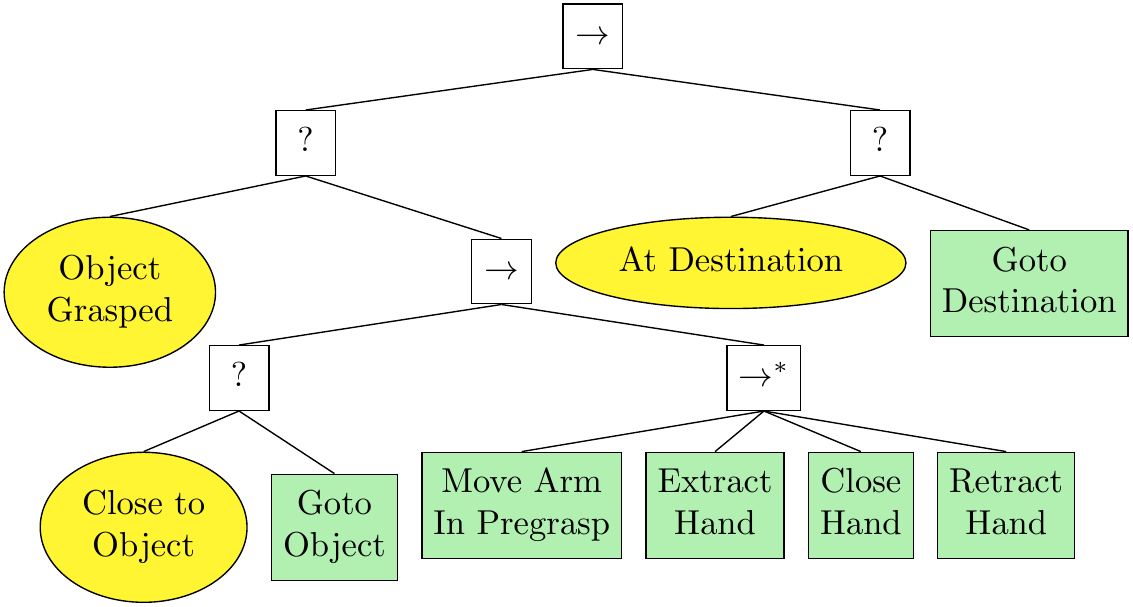}
\caption{Original BT encoding the behavior of Experiment~\ref{ex:r1}. The node labeled with $\rightarrow^*$ represents a sequence with memory.}
\label{fig:ex:r1:unsync:bt}
\end{figure}

However, the robot can execute the pregrasp motion as well as the extraction of the arm while it approaches the user and the retraction of the arm while it navigates towards the destination, whereas the grasping action needs the robot to be still. Hence we can execute the pre-grasp and post-grasp action while the robot moves, speeding up the execution. The BT in Figure~\ref{fig:ex:r1:sync:bt} models such behavior, taking advantage of the resource synchronization, where the action \say{Goto} and \say{Close Hand} allocate the resource \emph{Mobile Base} as long as they are running.
Figure~\ref{fig:ex:r1:sync:exec} shows some execution steps. The concurrent execution of some actions allows a faster overall behavior as in \cite{colledanchise2016advantages, brunner2019autonomous}.

\begin{figure}[h!]
\centering
\includegraphics[width=0.9\columnwidth]{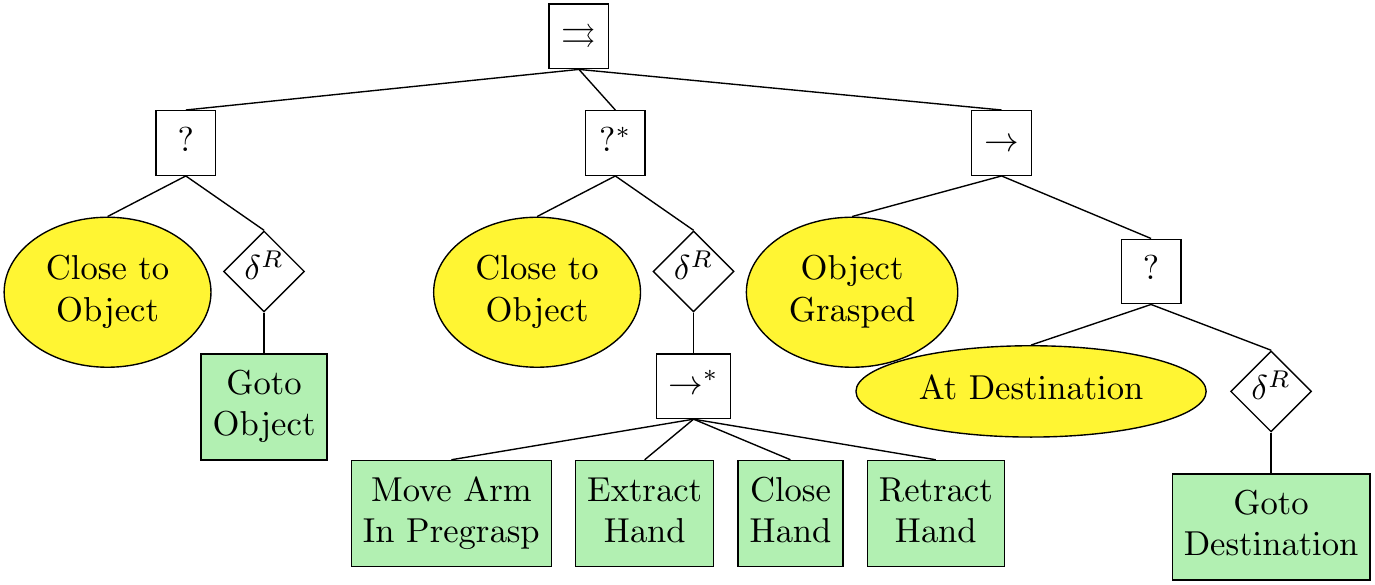}
\caption{BT encoding the behavior of Experiment~\ref{ex:r1} using the resource synchronization mechanism. This BT is significantly simpler than the one in Figure~\ref{fig:ex:r1:unsync:bt}.}
\label{fig:ex:r1:sync:bt}
\end{figure}
\end{experiment}

\begin{figure}[t!]
\begin{subfigure}[t]{0.49\columnwidth}
\includegraphics[width=\columnwidth]{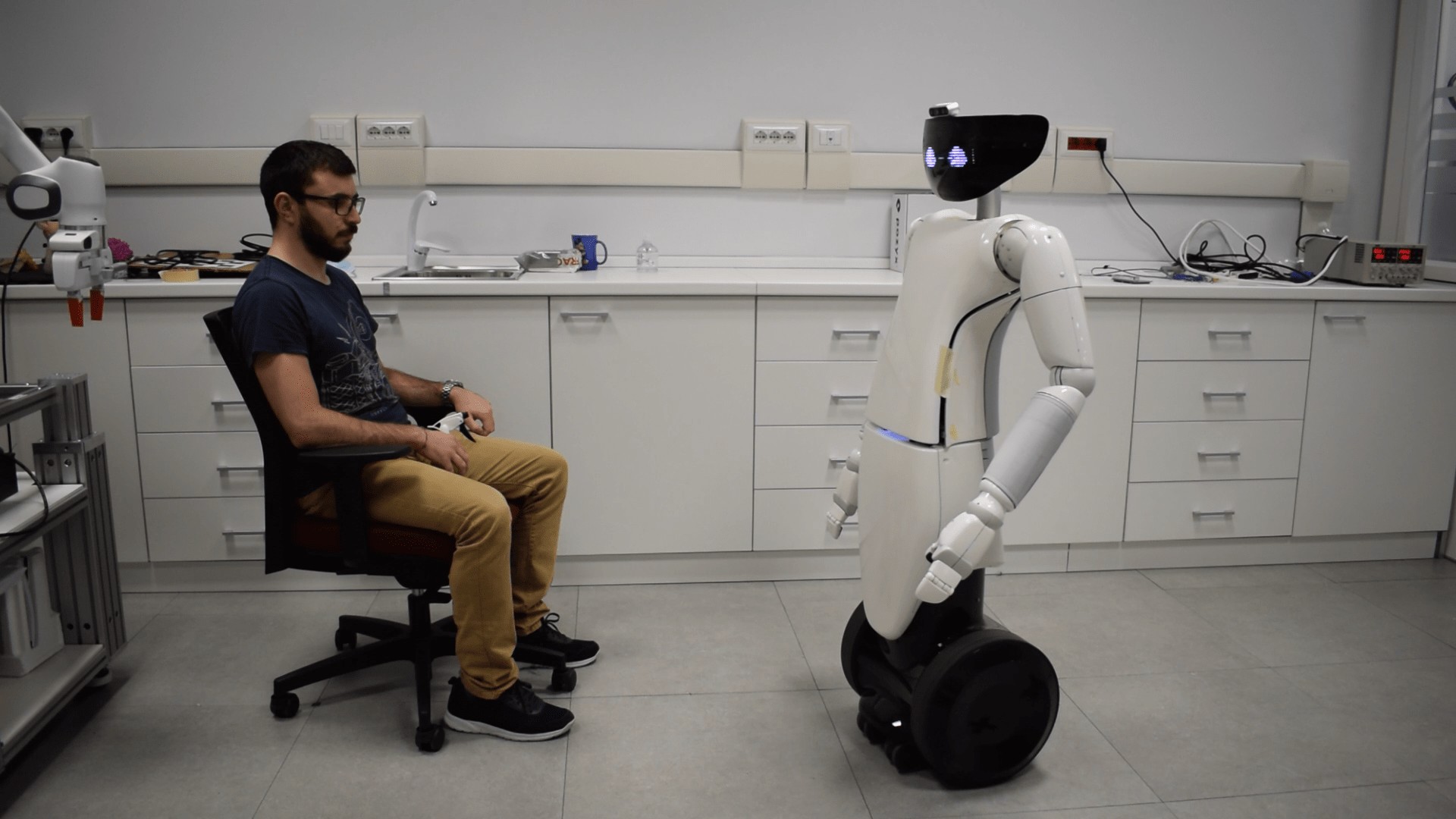}
\caption{The robot approaches the user. \\  Action Executed: \say{Goto Object}.}
\end{subfigure}
\begin{subfigure}[t]{0.49\columnwidth}
\includegraphics[width=\columnwidth]{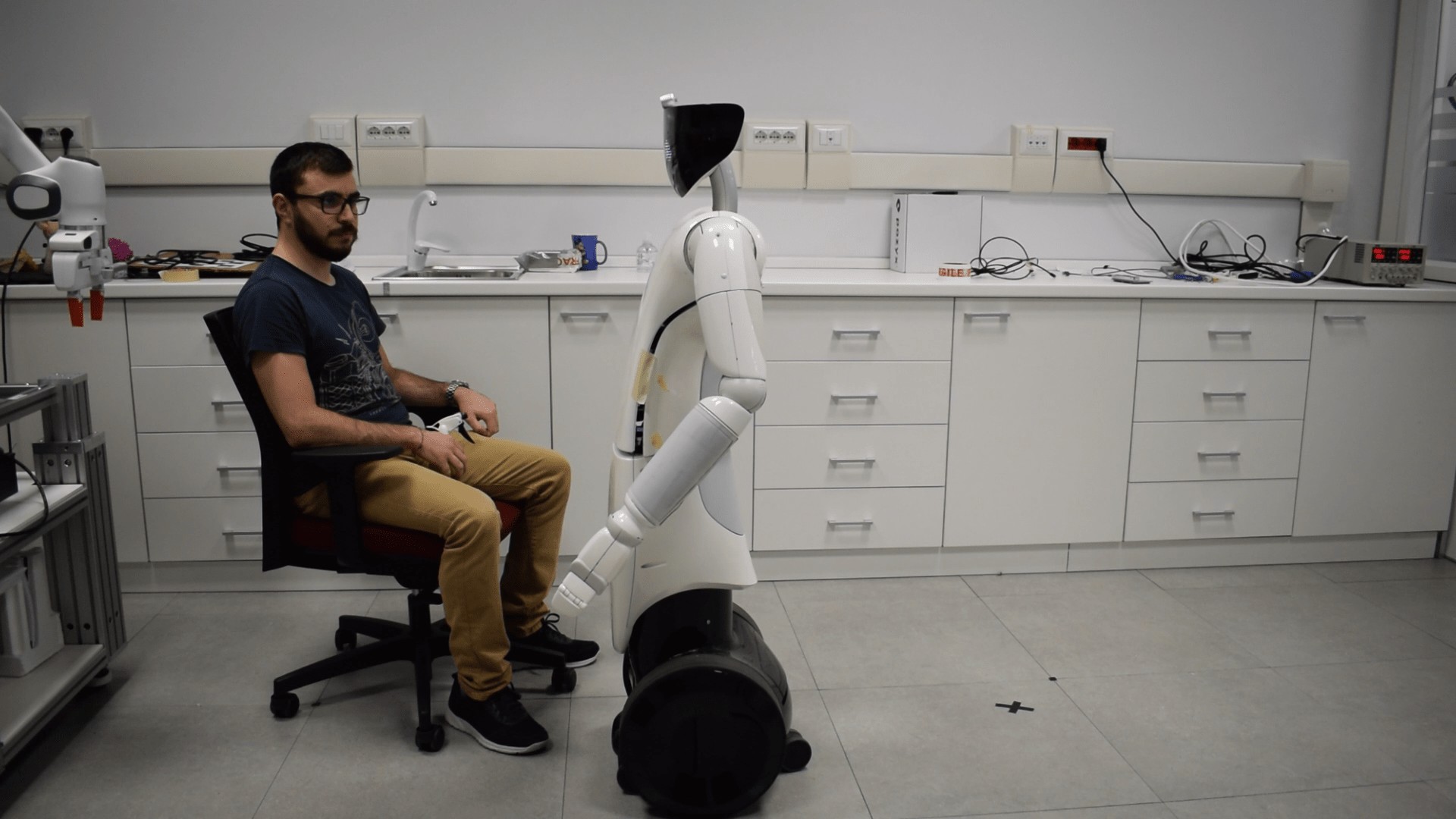}
\caption{The robot reaches the user. \\  Action Executed: \say{Goto Object}.}
\end{subfigure}

\vspace*{0.5em}
\begin{subfigure}[t]{0.49\columnwidth}
\includegraphics[width=\columnwidth]{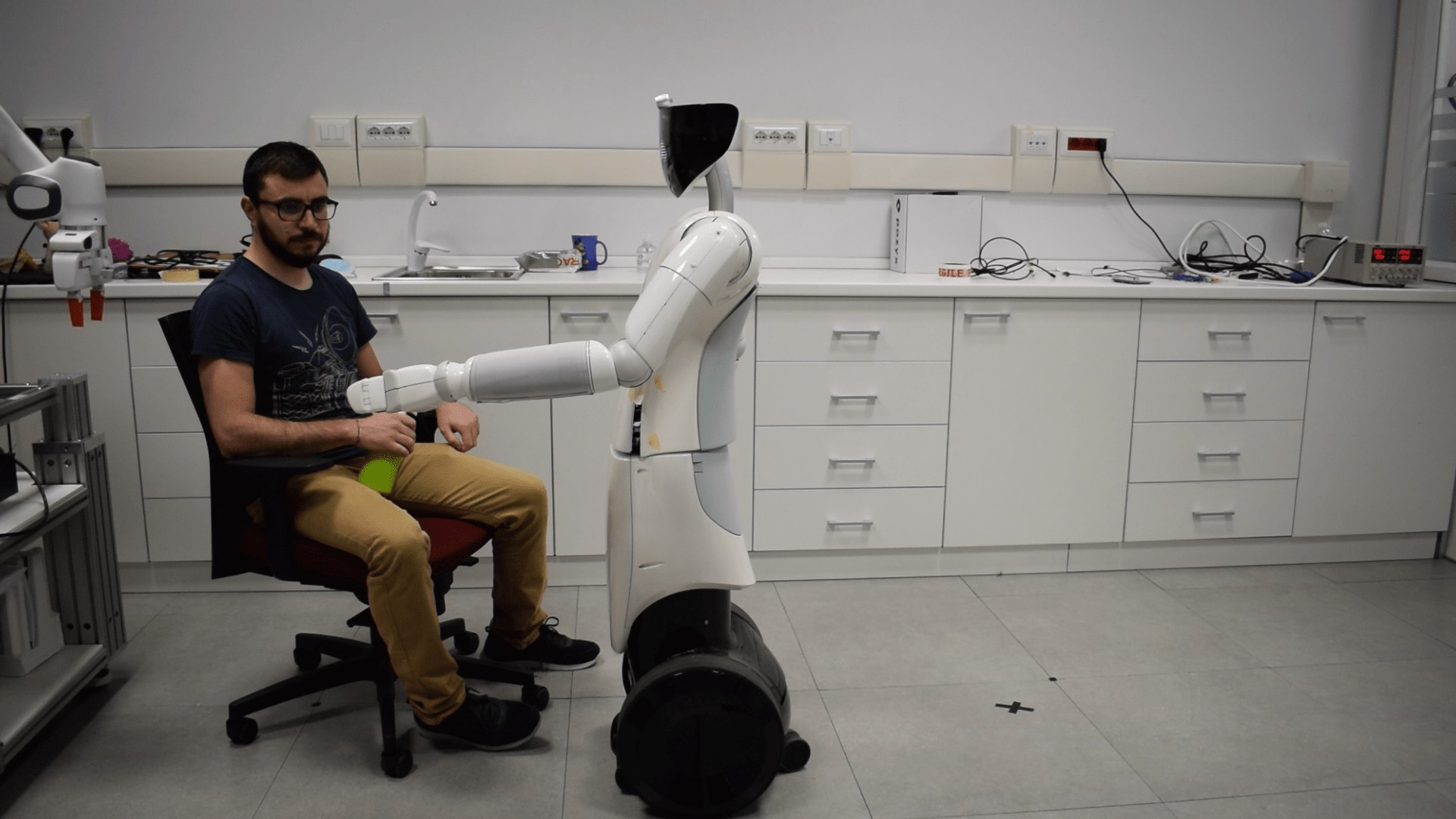}
\caption{The robot moves the arm in pre-grasp position. Action Executed: \say{Move Arm}.}
\end{subfigure}
\begin{subfigure}[t]{0.49\columnwidth}
\includegraphics[width=\columnwidth]{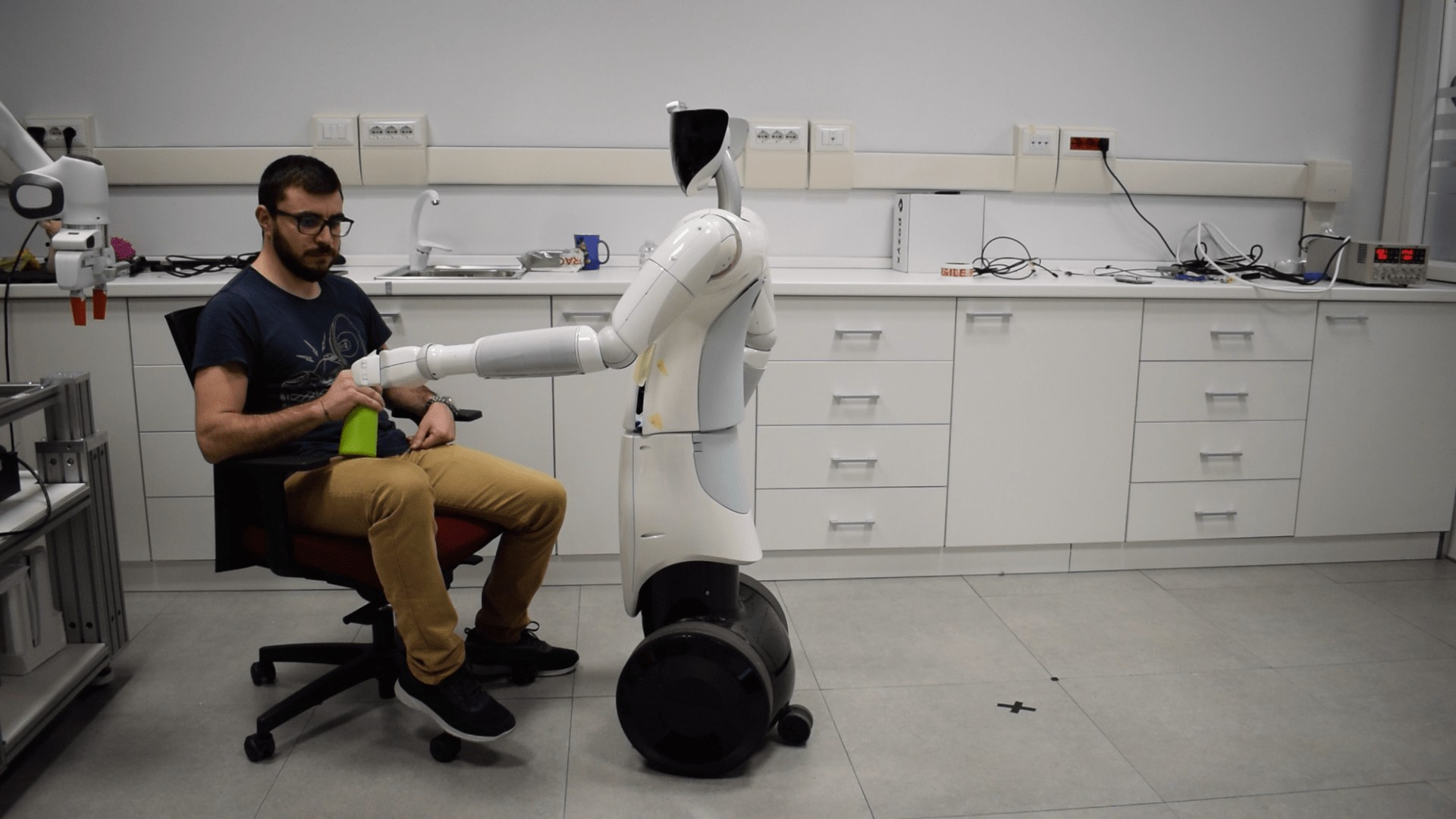}
\caption{The robot extracts the hand.\\  Actions Executed: \say{Extract Hand}.}
\end{subfigure}

\vspace*{0.5em}

\begin{subfigure}[t]{0.49\columnwidth}
\includegraphics[width=\columnwidth]{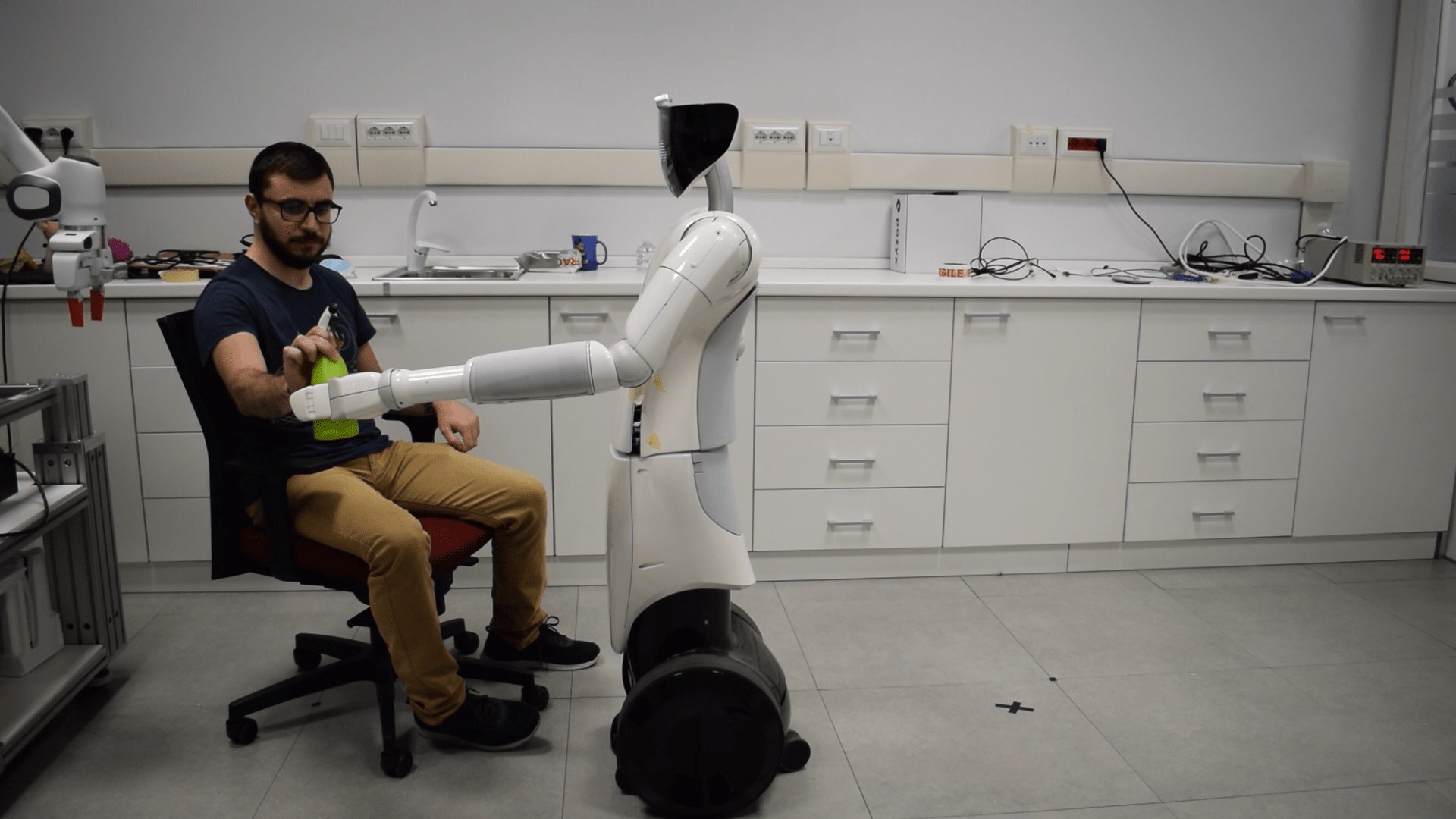}
\caption{The robot closes the hand. \\  Actions Executed: \say{Close Hand}.}
\end{subfigure}
\begin{subfigure}[t]{0.49\columnwidth}
\includegraphics[width=\columnwidth]{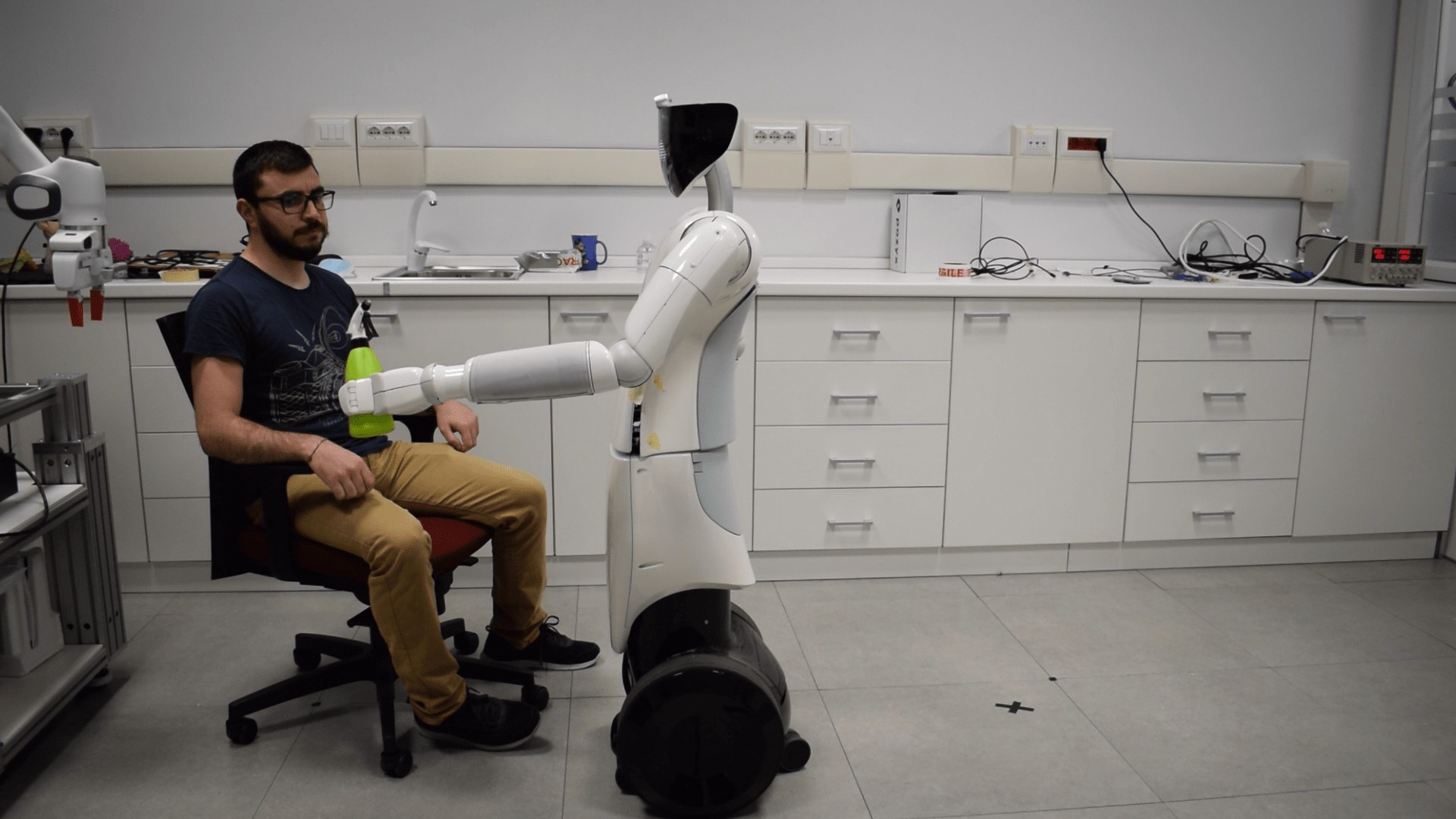}
\caption{The robot retracts the hand. \\  Actions Executed: \say{Retract Hand}.}
\end{subfigure}

\vspace*{0.5em}

\begin{subfigure}[t]{0.49\columnwidth}
\includegraphics[width=\columnwidth]{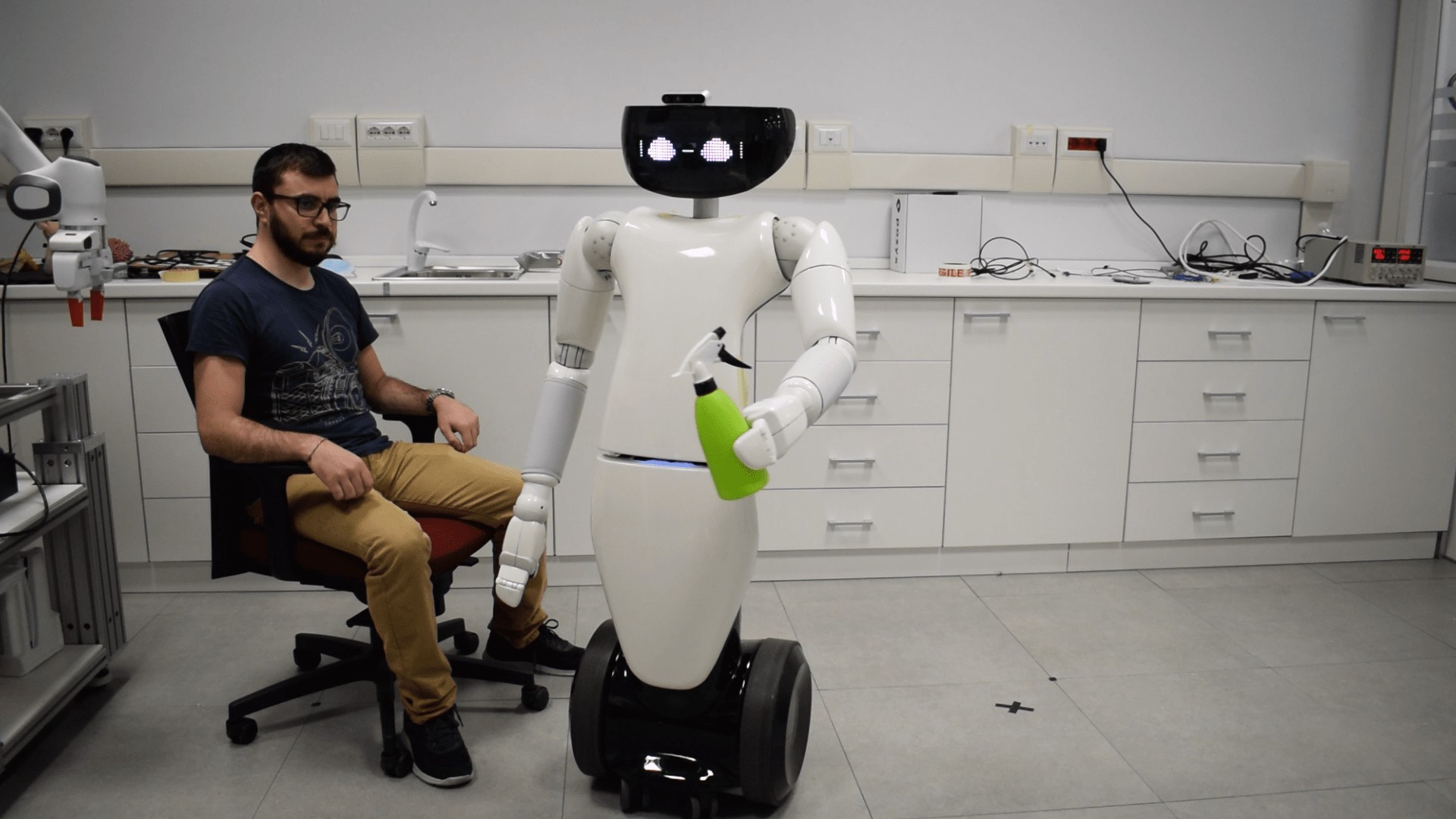}
\caption{The robot moves towards the destination. Actions Executed: \say{Goto Destination}.}
\end{subfigure}
\begin{subfigure}[t]{0.49\columnwidth}
\includegraphics[width=\columnwidth]{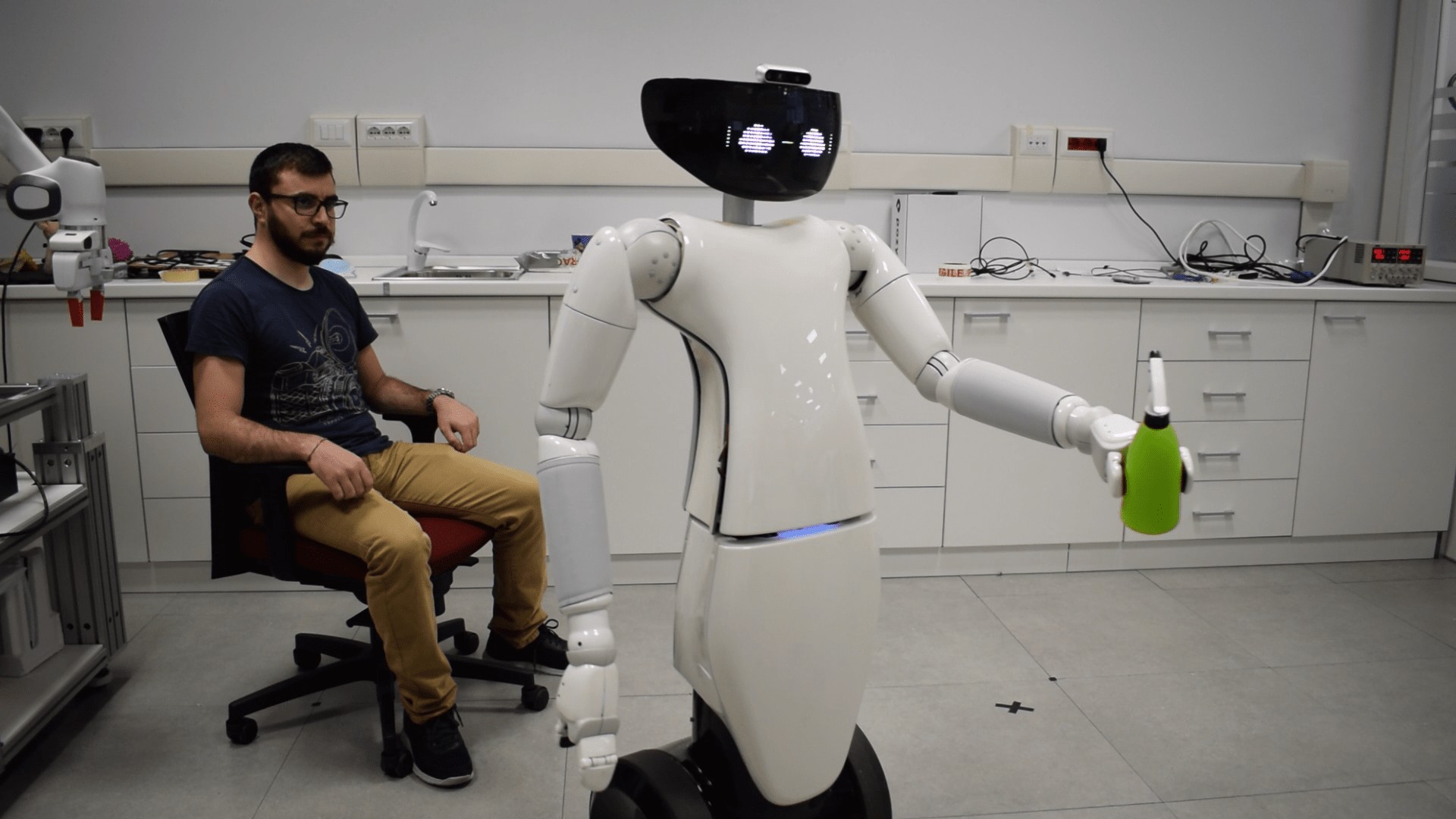}
\caption{The robot moves towards the destination. Actions Executed: \say{Goto Destination}.}
\end{subfigure}
\caption{Execution screenshots of Experiment~\ref{ex:r1} running the BT in Figure~\ref{fig:ex:r1:unsync:bt}. }
\label{fig:ex:r1:unsync:exec}
\end{figure} 

\begin{figure}[h!]	
\begin{subfigure}[t]{0.49\columnwidth}
\includegraphics[width=\columnwidth]{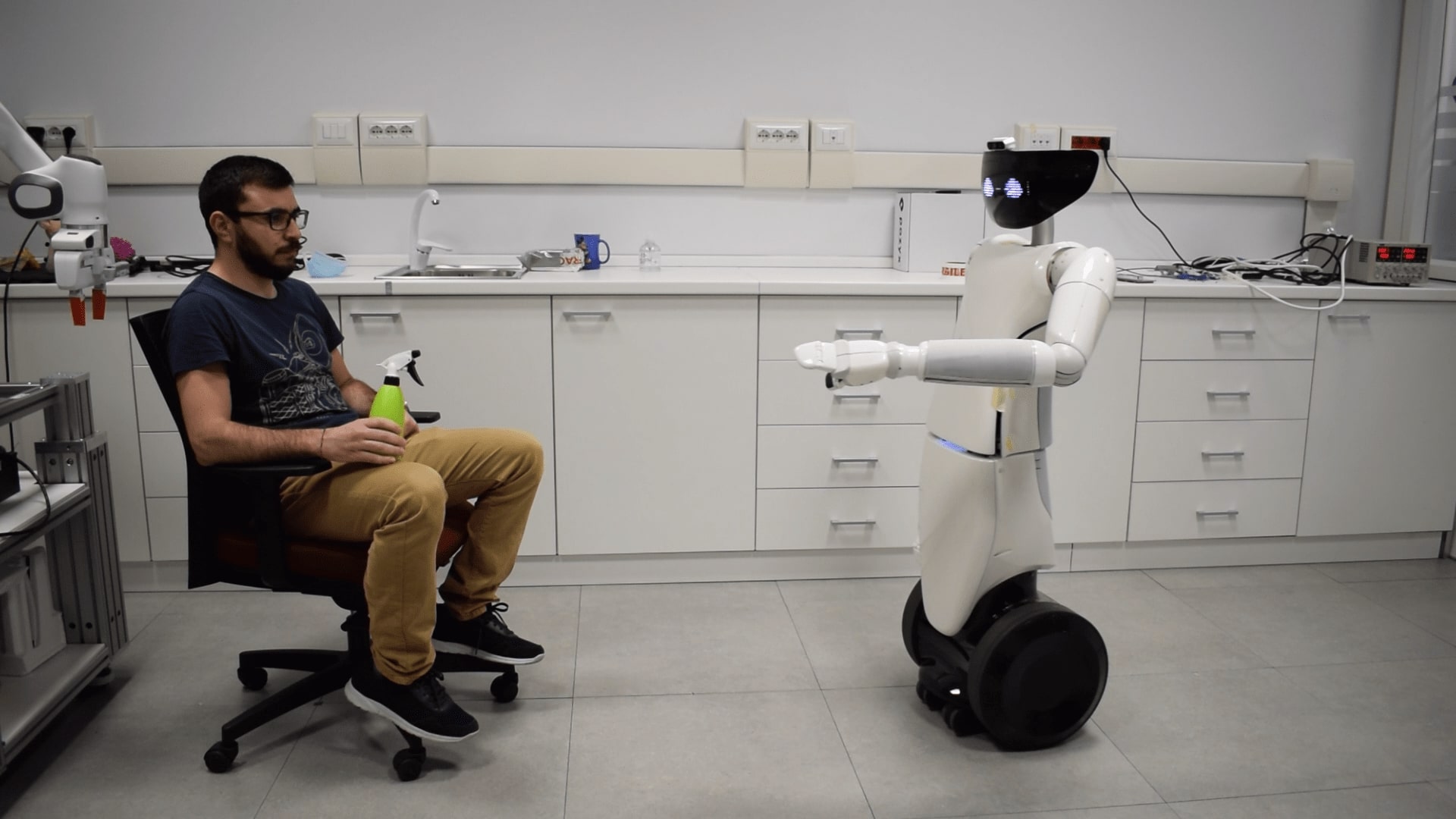}
\caption{The robot moves towards the user while positioning the arm and hand.  Actions Executed: \say{Goto Object}, \say{Move Arm In Pregrasp}, and \say{Extract Hand}.}
\end{subfigure}
\begin{subfigure}[t]{0.49\columnwidth}
\includegraphics[width=\columnwidth]{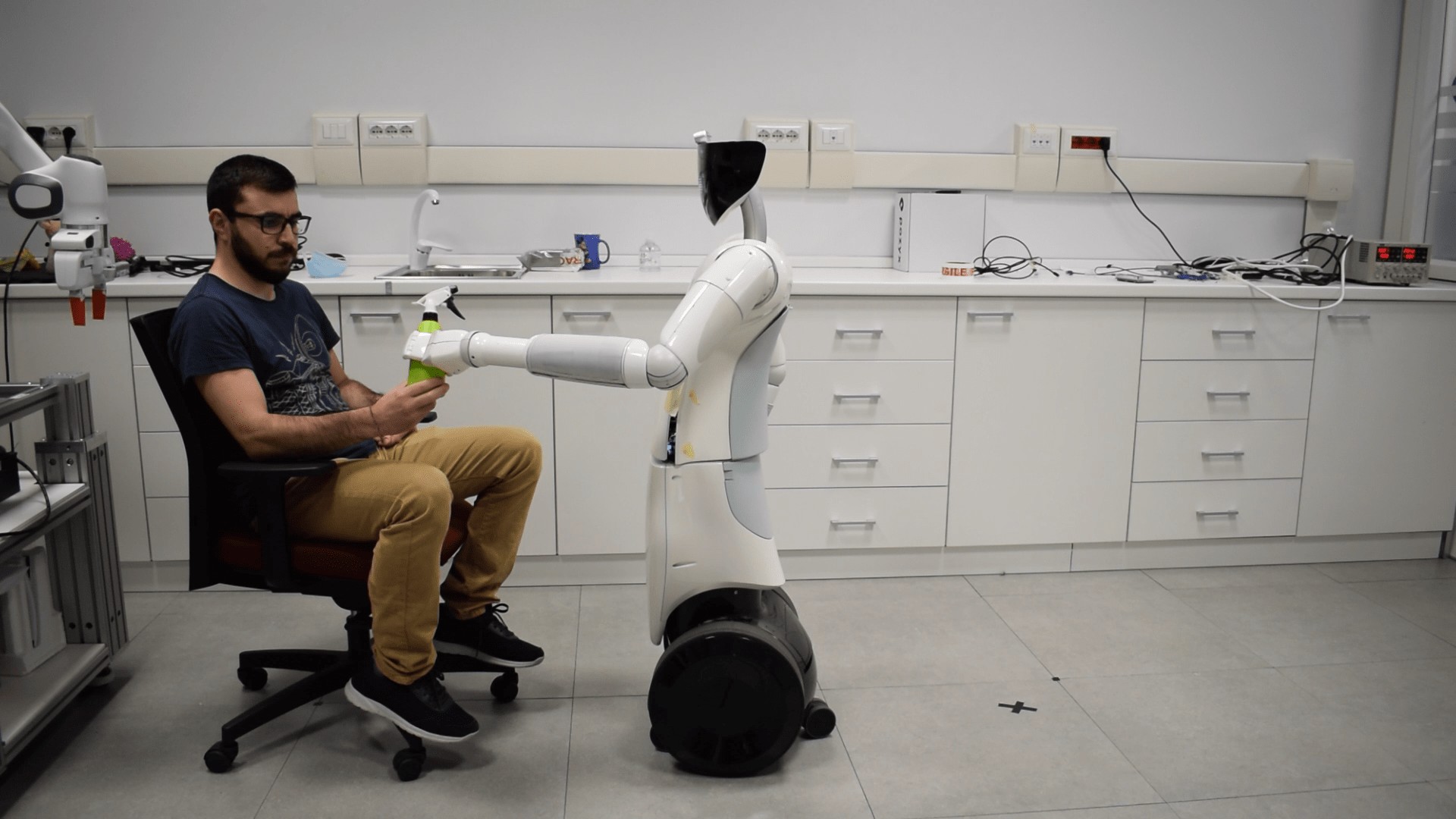}
\caption{The robot reaches the user and then closes the hand. \\  Action Executed: \say{Close Hand}.}
\end{subfigure}

\vspace*{0.5em}

\begin{subfigure}[t]{0.49\columnwidth}
\includegraphics[width=\columnwidth]{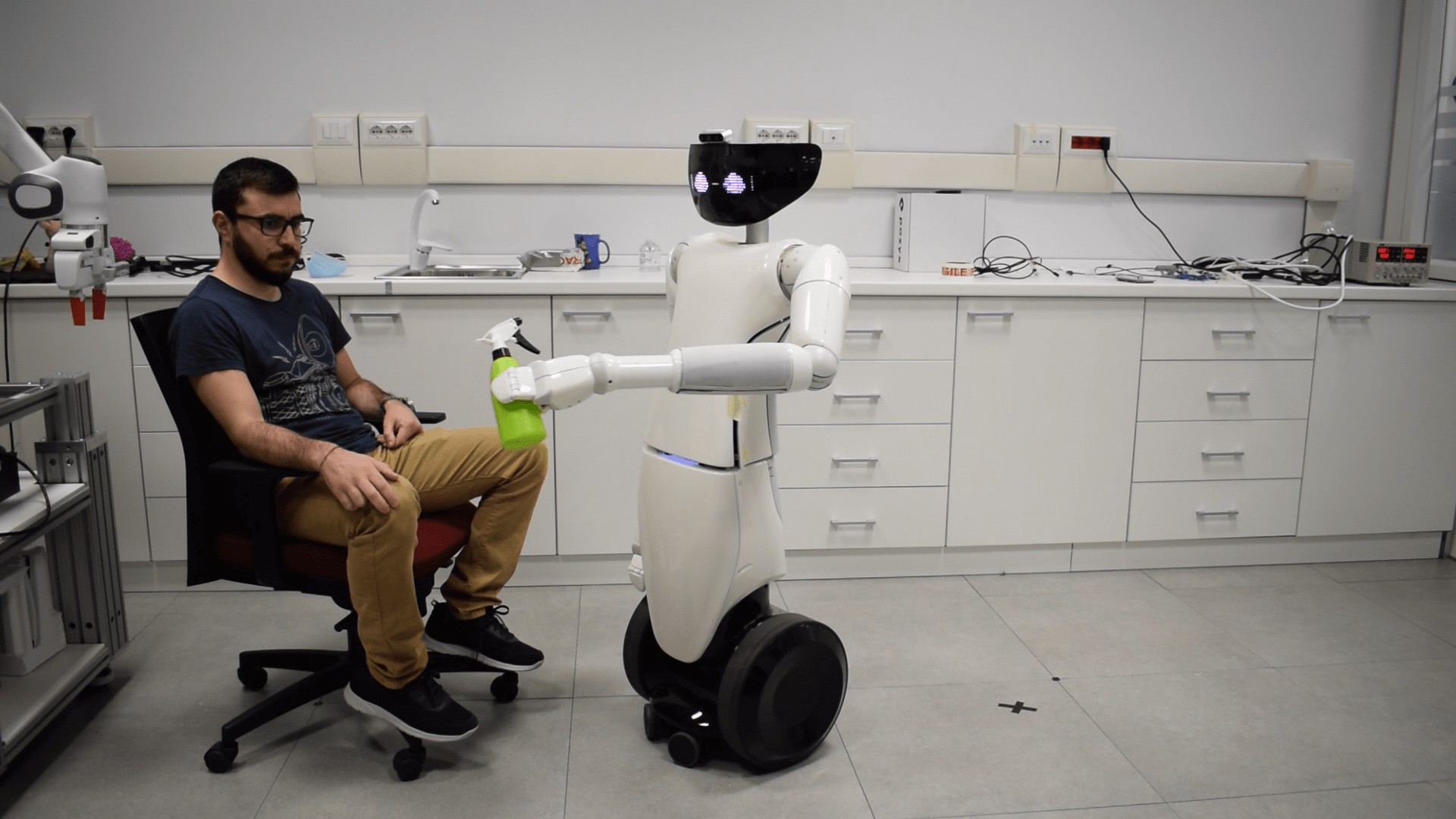}
\caption{The robot moves towards the destination while retracting the hand. \\ Actions Executed: \say{Goto Destination} and \say{Retract Hand}.}
\end{subfigure}
\begin{subfigure}[t]{0.49\columnwidth}
\includegraphics[width=\columnwidth]{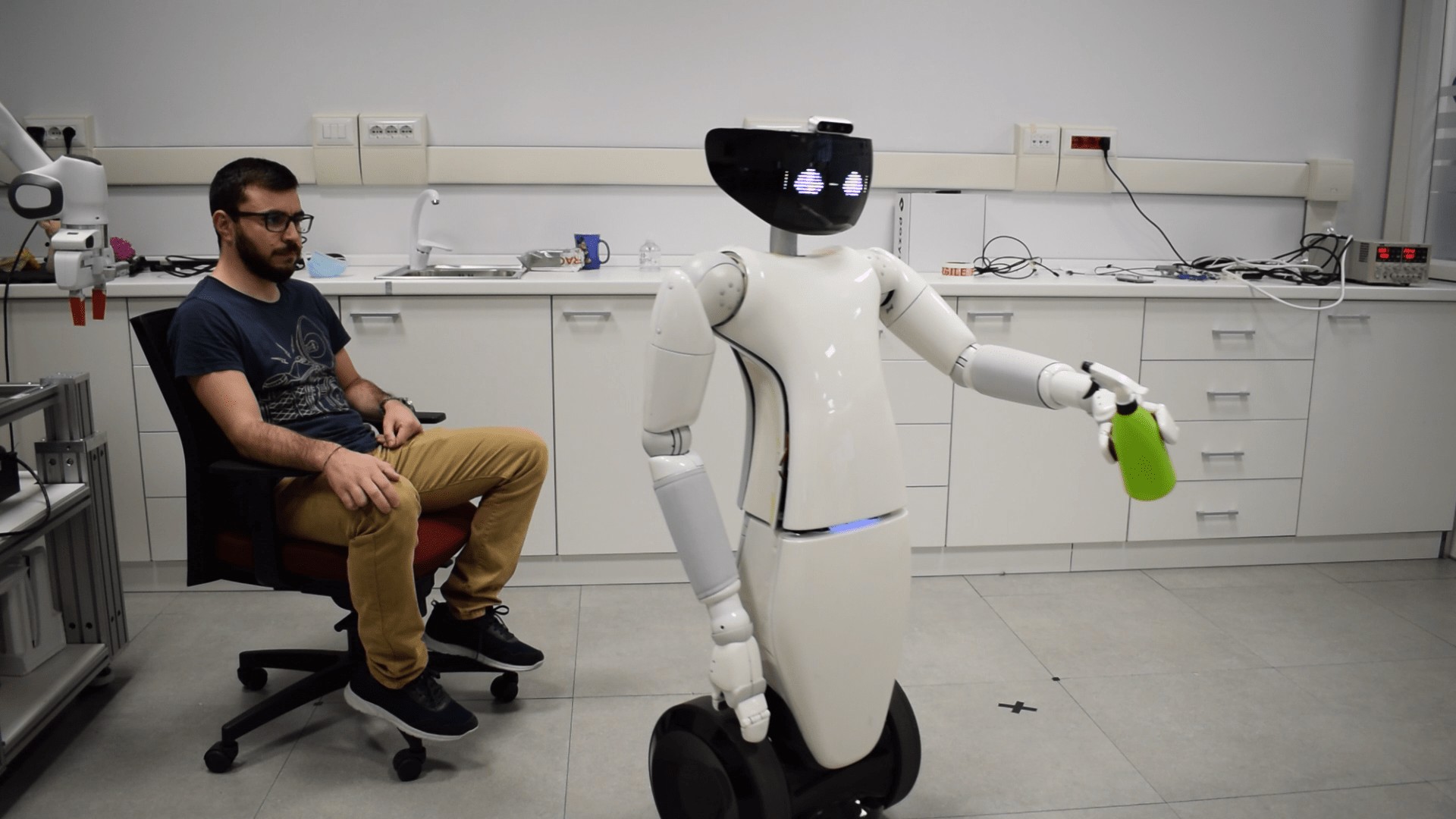}
\caption{The hand gets retracted. \\ Action Executed: \say{Goto Destination}.}
\end{subfigure}
\caption{Execution screenshots of Experiment~\ref{ex:r1} running the BT in Figure~\ref{fig:ex:r1:sync:bt}.}
\label{fig:ex:r1:sync:exec}
\end{figure}

 \paragraph*{Perpetual Actions}
 We now present an experiment where we show the applicability of our approach with perpetual actions. Experiment~\ref{ex:panda} below presents an implementation of Example~\ref{ex:perpetual}, inspired by the literature~\cite{rovida2018motion}.

\begin{experiment}[Panda Robot]
\label{ex:panda}
 An industrial manipulator has to insert a piston into a hollow cylinder.  The piston's rod and the piston's head are attached via a revolute joint. To correctly insert the rod, the robot must keep it aligned during the insertion into the cylinder. During the execution, the end-effector gets misaligned, requiring the robot to realign the rod. Figure~\ref{fig:ex:perpetual:bt:sync} depicts the BT that encodes this task. The action progress equals the ones of Example~\ref{ex:perpetual}. Figure~\ref{fig:ex:panda:exec} shows the execution steps of this experiment with and without synchronization. We see how the synchronized execution fails since the robot inserts the piston too fast for the alignment sub-behavior to have an effect.
\vspace*{-0.5em}

\begin{figure}[h!]
%

\begin{subfigure}[t]{0.49\columnwidth}
\centering
\includegraphics[width=\columnwidth,angle=-90,trim={7cm 2cm 15cm 2cm},clip]{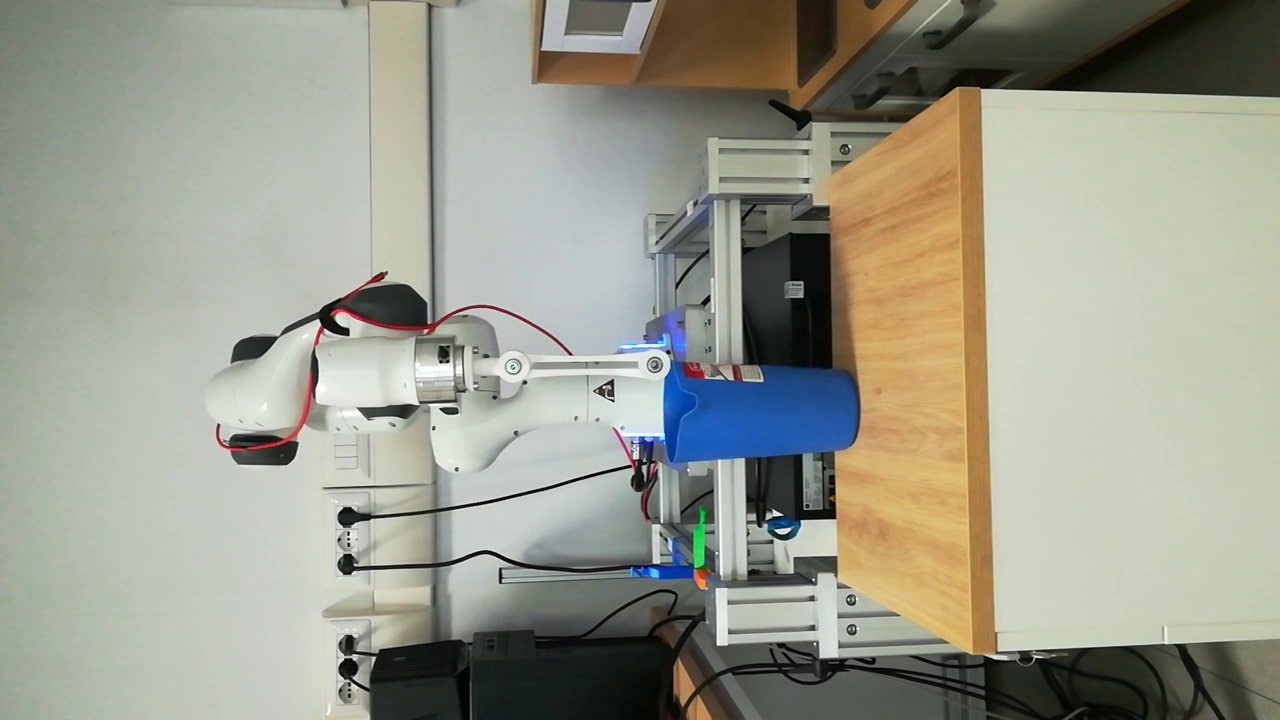}
\caption{[Unsync] The rod gets misaligned. The robot keeps inserting the rod.}
\end{subfigure}
\begin{subfigure}[t]{0.49\columnwidth}
\centering
\includegraphics[width=\columnwidth,angle=-90,trim={7cm 2cm 15cm 2cm},clip]{panda-unsync-1.jpg}
\caption{[Sync] The rod gets misaligned. The robot stops inserting the rod.}
\end{subfigure}

\vspace*{0.5em}

\begin{subfigure}[t]{0.49\columnwidth}
\centering
\includegraphics[width=\columnwidth,angle=-90,trim={7cm 2cm 15cm 2cm},clip]{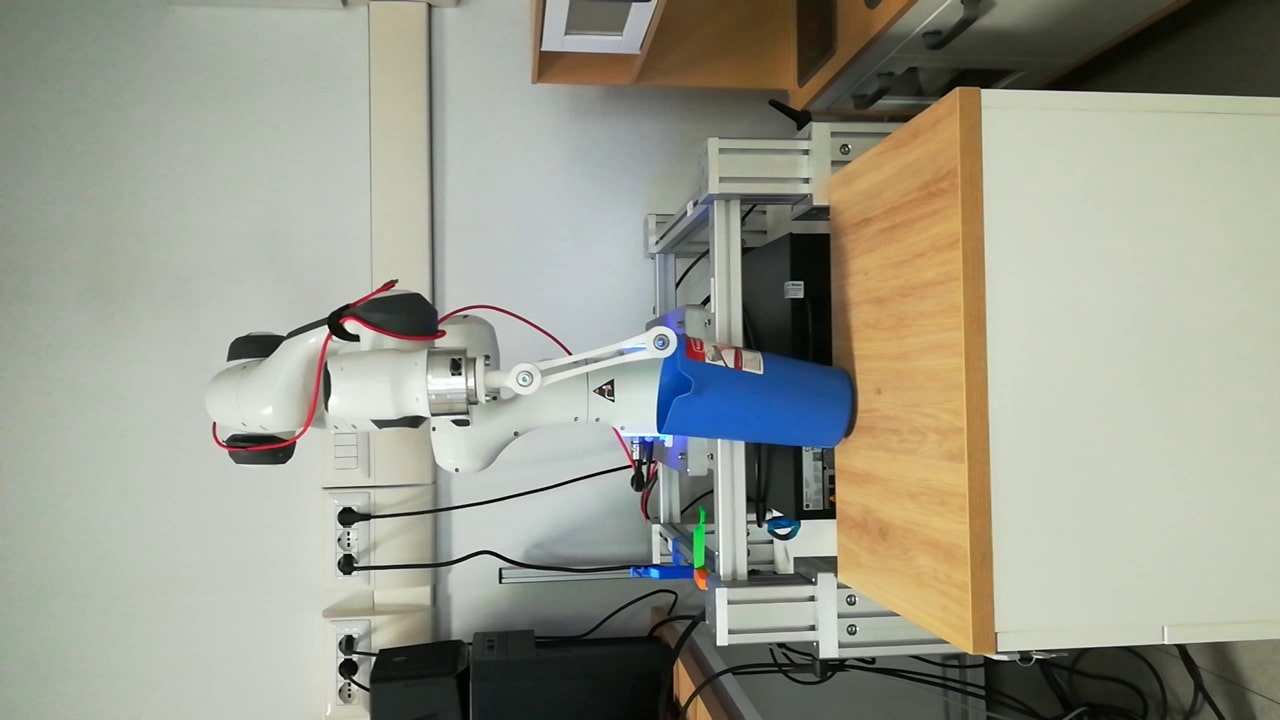}
\caption{[Unsync] The robot aligns the rod while this moves downwards. The rod hits the cylinder.}
\end{subfigure}
\begin{subfigure}[t]{0.49\columnwidth}
\centering
\includegraphics[width=\columnwidth,angle=-90,trim={7cm 2cm 15cm 2cm},clip]{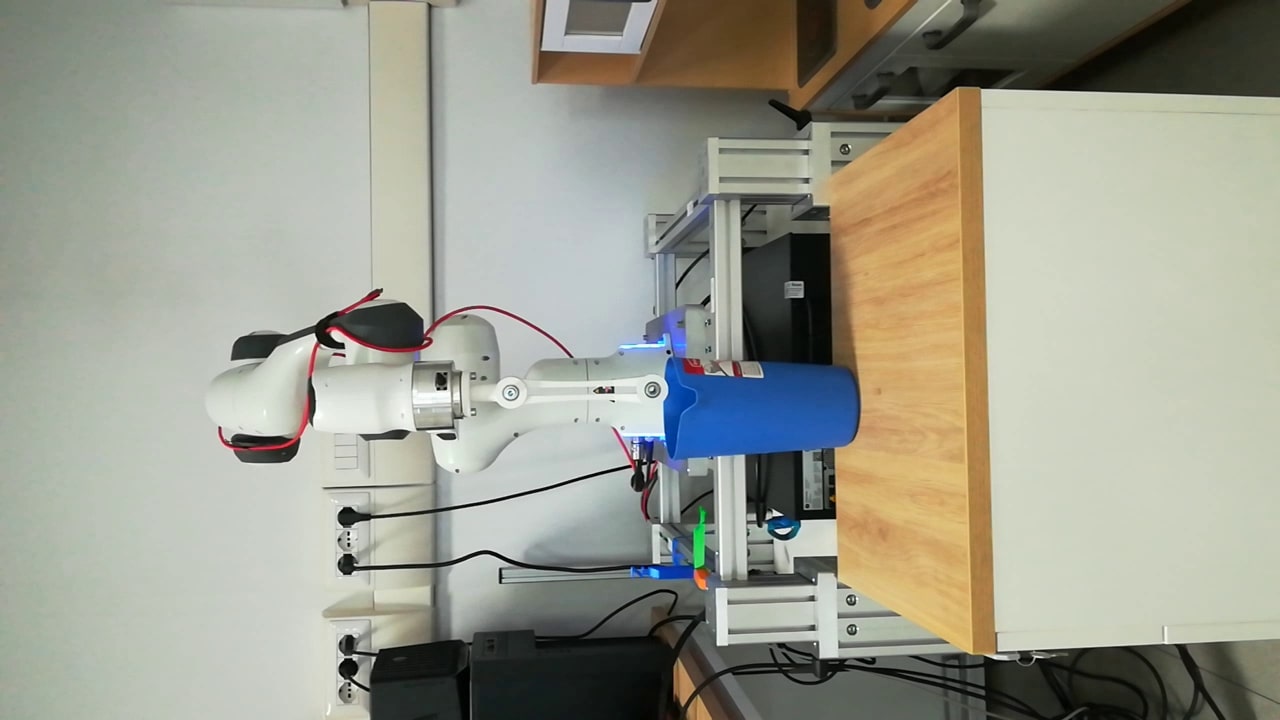}
\caption{[Sync] The robot aligns the rod.}
\end{subfigure}
\vspace*{0.5em}

\begin{subfigure}[t]{0.49\columnwidth}
\centering
\includegraphics[width=\columnwidth,angle=-90,trim={7cm 2cm 15cm 2cm},clip]{panda-unsync-2.jpg}
\caption{[Unsync] A safety fault stops the execution.}
\end{subfigure}
\begin{subfigure}[t]{0.49\columnwidth}
\centering
\includegraphics[width=\columnwidth,angle=-90,trim={7cm 2cm 15cm 2cm},clip]{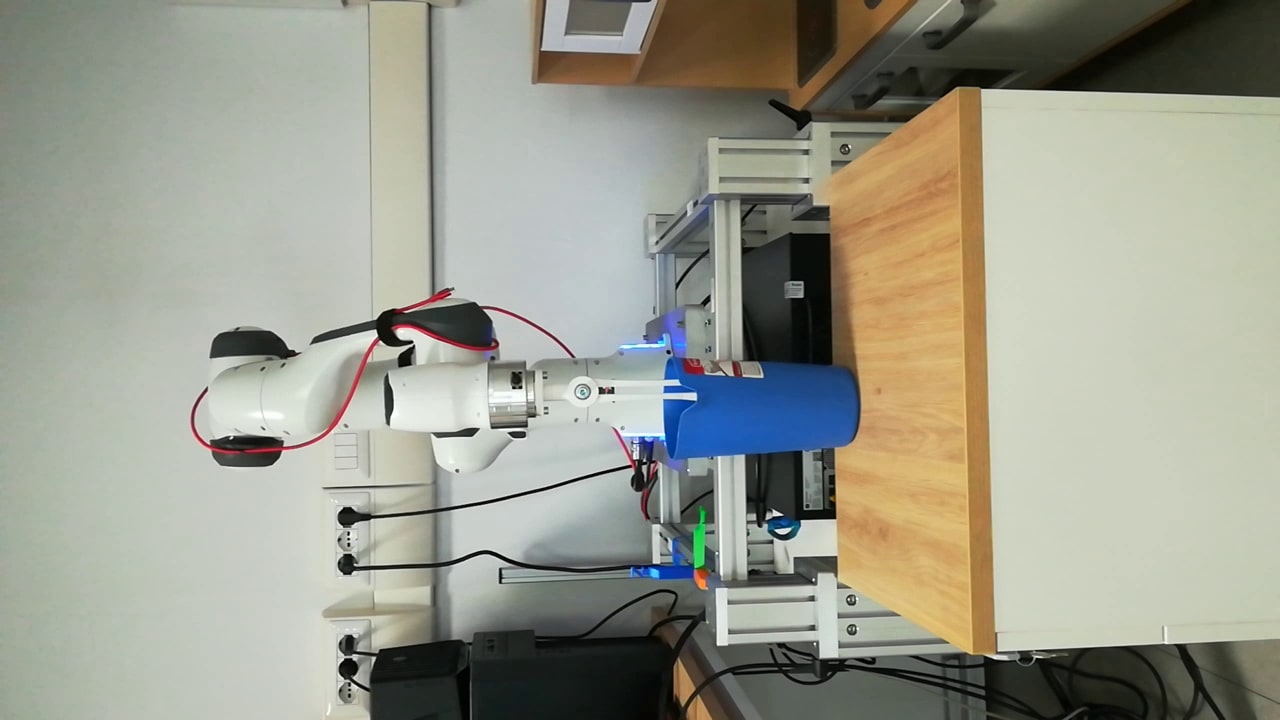}
\caption{[Sync] The insertion task resumes}
\end{subfigure}
\caption{Execution screenshots of Experiment~\ref{ex:panda} with (right) and without (left) synchronization. }
\label{fig:ex:panda:exec}
\end{figure} 

\end{experiment}

\newpage
\section{Software Library}
\label{sec:library}

\begin{figure}[h!]
\centering
\includegraphics[width=\columnwidth]{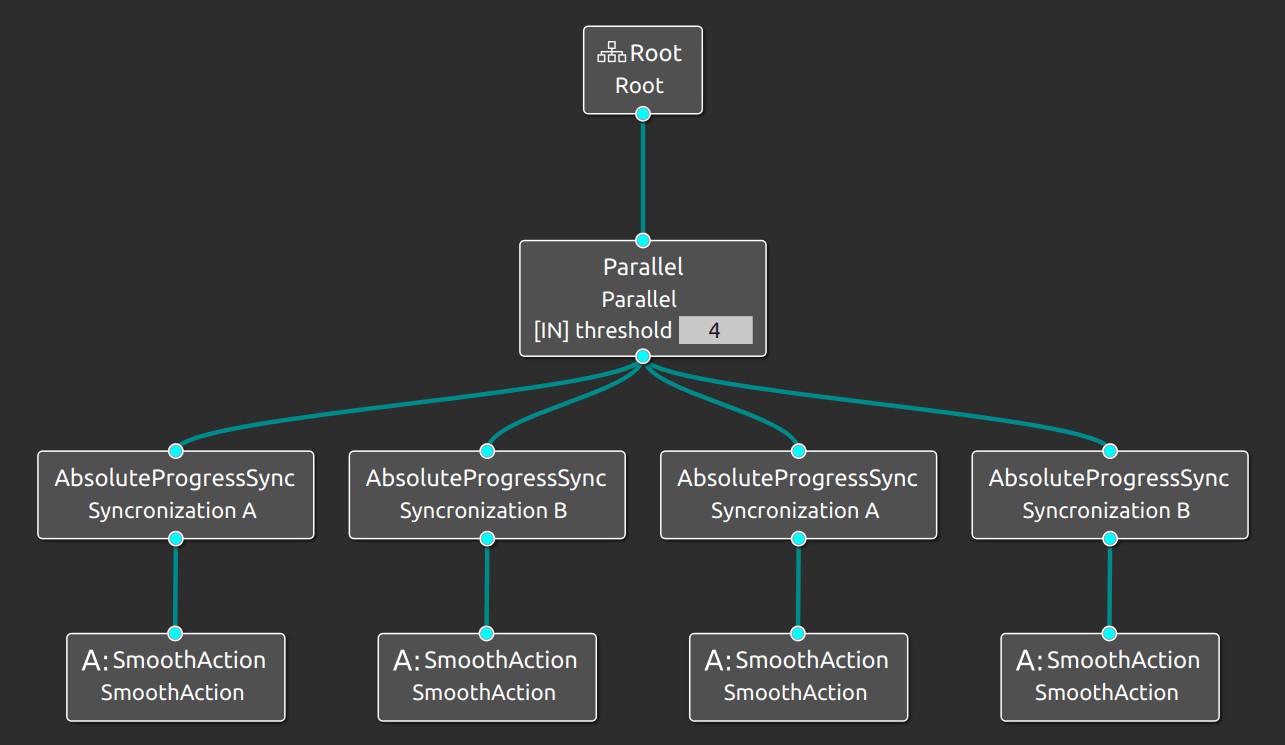}
\caption{Concurrent BT of example using the Groot GUI.}
\label{fig:sw}
\end{figure}

This section presents the third contribution of the paper. We made publicly available an implementation of the nodes presented in this paper. The decorators work with the BehaviorTree.CPP engine~\cite{BTCpp} and the Groot GUI~\cite{Groot}. The user can define the values of barriers $|B|$ (for absolute progress synchronization), the threshold $\Delta$ (for relative progress synchronization), or the priority increment function $g$ of Definition~\ref{def:priorityincrement}. The BT can also have independent synchronizations, as shown in the BT of Figure~\ref{fig:sw}. We made the details available in the library's repository.\footnote{\url{https://github.com/miccol/TRO2021-code}}

\subsection{Implement concurrent BTs with BehaviorTree.CPP}

Listing~\ref{btcpp} below shows the code to implement Example~\ref{ex:absolute} above with the BehaviorTree.CPP engine. The user has to instantiate the root and the actions nodes (Lines 1-4), then it defines the barriers (Lines 6-7), it instantiates the decorators (Lines 9-10), and finally, it constructs the BT (Lines 12-16).

\begin{lstlisting}[language=C++, caption={Implementation code for the BT in Example~\ref{ex:absolute} }, label=btcpp]
BT::ParallelNode parallel("root",2);

SyncSmoothAction action1("Arm Movement",0,0.015);
SyncSmoothAction action2("Base Movement",0,0.01);

AbsoluteBarrier barrier({0.1,0.2,0.3,0.4,0.5,0.6,
                         0.7,0.8,0.9,1.0});

DecoratorProgressSync dec1("dec1", &barrier);
DecoratorProgressSync dec2("dec2", &barrier);

dec1.addChild(&action1);
dec2.addChild(&action2);

parallel.addChild(&dec1);
parallel.addChild(&dec2);
\end{lstlisting}

\subsection{Implement concurrent BTs with Groot}
We provide a palette of nodes that allow the user to instantiate them using the  Groot GUI in a drag-and-drop fashion. The instructions on how to load the palette are available in the library's documentation. Details on how to instantiate and run a generic BT are available in the Groot library's documentation.

\newpage

\section{Theoretical Analysis}
\label{sec:analysis}
This section presents the fourth contribution of the paper. We first give a set of formal definitions, and then we provide a mathematical analysis of the BT synchronization.
As the decorators defined in this paper may disable the execution of some sub-trees, we need to identify the circumstances that preserve the entire BT's properties.

\subsection{State-space Formulation of Behavior Trees}
\label{sec:background.ss}
The state-space formulation of BTs~\cite{BTBook} allows us to study them from a mathematical standpoint. A recursive function call represents the tick. We will use this formulation in the proofs below.	

\begin{definition}[Behavior Tree \cite{BTBook}]
\label{bg.def:BT}
A BT is a three-tuple 
\begin{equation}
 \bt_i\triangleq\{f_i,r_i, \Delta t\}, 
\end{equation}
where $i\in \mathbb{N}$ is the index of the tree, $f_i: \mathbb{R}^n \rightarrow  \mathbb{R}^n$ is the right hand side of a difference equation, $\Delta t$ is a time step and 
$r_i$ is the return status that can be equal to either \emph{Running}, \emph{Success}, or \emph{Failure}. Finally,  let $x_k\triangleq x(t_k)$ be the system state at time $t_k$, then the execution of a BT $\bt_i$ is described by:
\begin{eqnarray}
 x_{k+1}&=&f_i( x_{k}),  \label{bts:eq:executionOfBT}\\
 t_{k+1}&=&t_{k}+\Delta t.
\end{eqnarray}
 \end{definition}

\begin{definition}[Sequence compositions of BTs~\cite{BTBook}]
\label{bts:def.seq}
 Two or more BTs  can be composed into a more complex BT using a Sequence operator,
 $$\bt_0=\mbox{Sequence}(\bt_1,\bt_2).$$ 
 Then $r_0,f_0$ are defined as follows
\begin{eqnarray}
   \mbox{If }x_k\in S_1&& \\
   r_0(x_k) &=&  r_2(x_k) \\
   f_0(x_k) &=&  f_2(x_k) \label{eq:seq1}\\ 
   \mbox{ else }&& \nonumber \\
   r_0(x_k) &=&  r_1(x_k) \\
   f_0(x_k) &=&  f_1(x_k). \label{eq:seq2}
 \end{eqnarray}
\end{definition}
$\bt_1$ and $\bt_2$ are called children of $\bt_0$. 

\begin{remark}
When executing the new BT, $\bt_0$ first keeps executing its first child $\bt_1$ as long as it returns Running or Failure.  
\end{remark}
For notational convenience, we write:
\begin{equation}
 \mbox{Sequence}(\bt_1, \mbox{Sequence}(\bt_2,\bt_3))= \mbox{Sequence}(\bt_1,\bt_2, \bt_3)
\end{equation}
and similarly for arbitrarily long compositions.

\begin{definition}[Fallback compositions of BTs~\cite{BTBook}]
\label{bts:def.fal}
 Two or more BTs  can be composed into a more complex BT using a Fallback operator,
 $$\bt_0=\mbox{Fallback}(\bt_1,\bt_2).$$ 
 Then $r_0,f_0$ are defined as follows
\begin{eqnarray}
   \mbox{If }x_k\in \mathcal{F}_1&& \\
   r_0(x_k) &=&  r_2(x_k) \\
   f_0(x_k) &=&  f_2(x_k) \\ 
   \mbox{ else }&&\nonumber \\
   r_0(x_k) &=&  r_1(x_k) \\
   f_0(x_k) &=&  f_1(x_k).
 \end{eqnarray}
\end{definition}

For notational convenience, we write:
\begin{equation}
 \mbox{Fallback}(\bt_1, \mbox{Fallback}(\bt_2,\bt_3))= \mbox{Fallback}(\bt_1,\bt_2, \bt_3)
\end{equation}
and similarly for arbitrarily long compositions.

 \begin{definition}[Parallel compositions of BTs~\cite{BTBook}]
 \label{bts:def:parallel}
 Two or more BTs  can be composed into a more complex BT using a Parallel operator,
 $$\bt_0=\mbox{Parallel}(\bt_1,\bt_2, M).$$ 
Where $f_0(x) \triangleq (f_{1}(x),f_{2}(x))$ and $r_0$ is defined as follows
\begin{eqnarray}
   \mbox{If } M=1&&\nonumber \\
   r_0(x) &=&  \mathcal{S}  \mbox{ If } r_1(x)=\mathcal{S} \vee r_2(x)=\mathcal{S}\\
   r_0(x) &=&  \mathcal{F}  \mbox{ If } r_1(x)=\mathcal{F} \wedge r_2(x)=\mathcal{F}\\
   r_0(x) &=&  \mathcal{R}  \mbox{ else } \\
   \mbox{If } M=2&&\nonumber \\
   r_0(x) &=&  \mathcal{S}  \mbox{ If } r_1(x)=\mathcal{S} \wedge r_2(x)=\mathcal{S}\\
   r_0(x) &=&  \mathcal{F}  \mbox{ If } r_1(x)=\mathcal{F} \vee r_2(x)=\mathcal{F}\\
   r_0(x) &=&  \mathcal{R}  \mbox{ else } 
 \end{eqnarray}
\end{definition}
For notational convenience, we write:
\begin{equation}
 \mbox{Parallel}(\bt_1, \mbox{Parallel}(\bt_2,\bt_3,2), 2)= \mbox{Parallel}(\bt_1,\bt_2, \bt_3 ,3)
\end{equation}
as well as:
\begin{equation}
 \mbox{Parallel}(\bt_1, \mbox{Parallel}(\bt_2,\bt_3,1), 1)= \mbox{Parallel}(\bt_1,\bt_2, \bt_3 ,1)
\end{equation}
and similarly for arbitrarily long compositions.


\begin{definition}[Finite Time Successful~\cite{BTBook}]
\label{properties:def:FTS}
 A BT is Finite Time Successful (FTS) with region of attraction $R'$, if for all starting points $x(0)\in R'\subset R$, there is a time $\tau$, and a time $\tau'(x(0))$ such that $\tau'(x)\leq \tau$ for all starting points, and
 $x(t)\in R'  $ for
 all $t\in [0,\tau')$ 
 and $x(t)\in S$ for
  $t = \tau'$
\end{definition}
As noted in the following lemma, exponential stability implies FTS, given the right choices of the sets $S,F,R$.
\begin{lemma}[Exponential stability and FTS~\cite{BTBook}]
 A BT for which $x_s$ is a globally exponentially stable equilibrium of the execution,
 and $S \supset \{x: ||x-x_s||\leq \epsilon\}$, $\epsilon>0$, $F=\emptyset$,  $R=\mathbb{R}^n \setminus S$, is FTS.
\end{lemma}


\emph{Safety} is the ability to avoid a particular portion of the state-space, which we denote as the \emph{Obstacle Region}.  To make statements about the safety of composite BTs, we need the following definition. Details on safe BTs can be found in the literature~\cite{BTBook}.

\begin{definition}[Safeguarding~\cite{BTBook}]
\label{def:Safeguarding}
 A BT is safeguarding, with respect to the step length $d$, the obstacle region $O \subset \mathbb{R}^n$, and the initialization region $I \subset R$, if it is safe, and FTS with region of attraction $R' \supset I$ and a success region $S$, such that $I$ surrounds $S$ in the following sense:
\begin{equation}
  \{x\in X \subset \mathbb{R}^n: \inf_{s\in S} || x-s  || \leq d \} \subset I,
\end{equation}
where $X$ is the reachable part of the state space $\mathbb{R}^n$. 
  \end{definition}
This implies that the system, under the control of another BT with maximal statespace steplength $d$, cannot leave $S$ without entering $I$, and thus avoiding $O$~\cite{BTBook}.

\begin{definition}[Safe~\cite{BTBook}]
\label{properties:def:Safe}
 A BT is safe, with respect to the obstacle region $O \subset \mathbb{R}^n$, and the initialization region $I \subset R$,
 if for all starting points $x(0)\in I$, we have that $x(t) \not \in O$, for all $t \geq 0$.
 \end{definition}

%

\subsection{CBT's Definition}
We now formulate additional definitions. We use these definitions to provide a state-space formulation for CBTs (Definition~\ref{bts.def:CBT} below) and to prove system properties.

\begin{definition}[Progress Function]
\label{def:progress}
The function $p: \mathbb{R}^n \to [0,1]$ is the progress function. It indicates the progress of the BT's execution at each state.
\end{definition}

\begin{definition}[Resources]
\label{ps.def.L}
$R$ is a collection of symbols that represents the resources available in the system.
\end{definition}


\begin{definition}[Allocated Resource]
\label{ps.def.allocatedresources}
Let $\mathcal{N}$ be the set of all the nodes of a BT, the function $\alpha : \mathbb{R}^n \times R \to \mathcal{N}$ is the resource allocation function. It indicates the BT using a resource.
\end{definition}

\begin{definition}[Resource Function]
\label{ps.def.resources}
The function $Q : \mathbb{R}^n \to 2 ^ R$ is the resource function. It indicates the set of resources needed for a BT's execution at each state.
\end{definition}

\begin{definition}[Node priority]
\label{def:priority}
The function $\rho : \mathbb{R}^n \to \mathbb{R}$ is the priority function. It indicates the node's priority to access a resource.
\end{definition}

\begin{definition}[Priority Increment Function]
\label{def:priorityincrement}
The function $g : \mathbb{R^n} \to \mathbb{R}$ is the priority increment function. It indicates how the priority changes while a node is waiting for a resource.
\end{definition}

We can now define a CBT as BT with information regarding its progress and the resources needed as follows:

\begin{definition}[Concurrent BTs]
\label{bts.def:CBT}
A CBT is a tuple 
\begin{equation}
 \bt_i\triangleq\{f_i,r_i, \Delta t, p_i, q_i\}, 
\end{equation}
where $i$, $f_i$, $\Delta t$, 
$r_i$ are defined as in Definition~\ref{bg.def:BT}, $p_i$ is a progress function, and  $q$ is a resource function.
\end{definition}

A CBT has the functions $p_i$ and $q_i$ in addition to the others of Definition~\ref{bg.def:BT}. These functions are user-defined for Actions and Condition. For the classical operators, the functions are defined below.

\newpage
\begin{definition}[Sequence compositions of CBTs]
\label{bts:def.smoothseq}
 Two CBTs  can be composed into a more complex CBT using a Sequence operator,
 $$\bt_0=\mbox{Sequence}(\bt_1,\bt_2).$$ 
 The functions $r_0,f_0$ match those introduced in Definition~\ref{bts:def.seq}, while the functions $p_0,q_0$ are defined as follows
\begin{eqnarray}
   \mbox{If }x_k\in S_1&& \\
   p_0(x_k) &=&  \frac{p_1(x_k) + p_2(x_k)}{2} \\
   q_0(x_k) &=& q_2(x_k)\\
   \mbox{ else }&& \nonumber \\
   p_0(x_k) &=& \frac{p_1(x_k)}{2}\\
   q_0(x_k) &=& q_1(x_k).
 \end{eqnarray}
\end{definition}

\begin{definition}[Fallback compositions of CBTs]
\label{bts:def.smoothfal}
 Two CBTs  can be composed into a more complex CBT using a Fallback operator,
 $$\bt_0=\mbox{Fallback}(\bt_1,\bt_2).$$ 
 The functions $r_0,f_0$ are defined as in Definition~\ref{bts:def.fal}, while the functions $p_0,q_0$ are defined as follows
\begin{eqnarray}
   \mbox{If }x_k\in {F}_1&& \\
   p_0(x_k) &=& p_2(x_k)\\
   q_0(x_k) &=& q_2(x_k) \\ 
   \mbox{ else }&&\nonumber \\
   p_0(x_k) &=& p_1(x_k)\\
   q_0(x_k) &=& q_1(x_k).
 \end{eqnarray}
\end{definition}

\begin{definition}[Parallel compositions of CBTs]
 \label{bts:def:smoothpar}
 Two CBTs  can be composed into a more complex CBT using a Parallel operator,
 $$\bt_0=\mbox{Parallel}(\bt_1,\bt_2).$$ 
 The functions $r_0,f_0$ are defined as in Definition~\ref{bts:def:parallel}, while the functions $p_0$ and $q_0$ are defined as follows
\begin{eqnarray}
   p_0(x_k) &=& \mbox{min}(p_1(x_k), p_2(x_k)) \\ 
   q_0(x_k) &=& q_1(x_k) \cup q_2(x_k)
 \end{eqnarray}
\end{definition}
\begin{remark}
Conditions nodes do not perform any action. Hence their progress function can be defined as $p(x_k) = 0$ and their resource function as $q(x_k) = \emptyset$ $\forall x_k \in \mathbb{R}^n$.
\end{remark}

\begin{definition}[Absolute Barrier]
\label{def:absbarrier}
An absolute barrier is defined as:
\begin{equation}
\begin{split}
b(x_k) \triangleq min \{b_i \in B : \forall T_j \in T, p_j(x_k) \geq b_{i-1}  \land  \\ \land \exists \bt_k : p_k(x_k) \geq b_{i} \}
\end{split}
\end{equation}
with $B$ a finite set of progress values.
\end{definition}
\begin{definition}[Relative Barrier]
\label{def:relbarrier}
A relative barrier is defined as:
\begin{equation}
b(t_k) \triangleq min_{\bt_i \in T} \{p_i \} + \Delta
\end{equation}
with $\Delta \in  \left[0, 1\right]$
\end{definition}

\begin{definition}[Functional Formulation of a Progress Decorator Node]
 A CBT $\bt_1$ can be composed into a more complex BT using an Absolute Progress Decorator operator,
  $$\bt_0=\mbox{AbsoluteProgress}(\bt_1, b(x_k)).$$ 
 
Then $r_0,f_0, p_0, Q_0$ are defined as follows
\begin{align}
&p_0(x_k)=p_1(x_k)\\
&Q_0(x_k) = Q_1(x_k)\\
   \mbox{If } &p_1(x_k) < b(x_k) \nonumber \\
   &f_0(x_k) = f_1(x_k)  \label{eq:form:progress:f}
 \\ 
     & r_0(x_k) = r_1(x_k) \label{eq:form:progress:r} \\
     \mbox{else }&  \nonumber \\
             &f_0(x_k) = x_k \\
               & r_0(x_k) = \mathcal{R} 
\end{align} 

 With $b(x_k)$ as in Definition~\ref{def:absbarrier} for an absolute synchronization or as in Definition~\ref{def:relbarrier} for a relative synchronization.
\end{definition}

\begin{definition}[Functional Formulation of a Resource Decorator Node]
 A Smooth BT $\bt_1$ can be composed into a more complex BT using a Resource Decorator operator,
 $$\bt_0=\mbox{ResourceDecorator}(\bt_1, g).$$ 
 
With $g$ as in Definition~\ref{def:priorityincrement}. then $r_0,f_0, p_0, Q_0$ are defined as follows
\begin{align} 
&p_0(x_k)=p_1(x_k)\\
   \mbox{If } &(\forall  q \in Q_1(x_k): \alpha(x_{k}, q) = \bt_1 \land \rho_1(x_k) \geq \rho_{max})  \nonumber \\ &\lor  \alpha(x_k, q) = \emptyset \nonumber \\
   &r_0(x_k) = r_1(x_k) \label{eq:form:resource:r}\\
   &f_0(x_k) = f_1(x_k) \label{eq:form:resource:f} \\
   &Q_0(x_k) = Q_1(x_k) \\
   &\alpha(x_{k}, q) = \begin{cases}
   \bt_1 & \mbox{if } q \in Q_1(x_k) \\
   \emptyset & \mbox{if }  q \not\in Q_1(x_k) :\\  & \hspace{0.5em} \alpha(x_{k-1}, q) = \bt_1\\
   \alpha(x_{k-1}, q) &  \mbox{otherwise}\\
           & \rho(x_k) =  \rho(x_{k-1})
   \end{cases}\\
      \mbox{els}&\mbox{e } \nonumber \\
         &r_0(x_k) = \mathcal{R} \\
        &f_0(x_k) = x_k \\
        &Q_0(x_k) = \emptyset\\
        & \rho_1(x_k) = \rho_1(x_{k-1}) + g_1(x_k) \label{eq:form:resource:g}
\end{align} 
 With $\alpha$ from Definition~\ref{ps.def.allocatedresources}.

\end{definition}
\begin{definition}[Active node]
A BT node $\bt_i$ is said to be active in a given BT if $\bt_i$ is either the root node or whenever $r(x_k) = R$,  $\bt_i$ will eventually receive a tick.
\end{definition}
An \emph{active} node is a node that eventually will receive a tick when its returns status is running.
\newpage
\subsection{Lemmas}
The synchronization mechanism proposed in this paper may jeopardize the FTS property (described in Definition~\ref{properties:def:FTS}) of a BT. In particular, a FTS BT may no longer receive ticks from decorators proposed in this paper as this will wait for another action indefinitely. This relates to the problem of 
\emph{starvation}, where a process waits for a critical resource and other processes, with a higher priority, prevent access to such resource~\cite{tanenbaum2015modern}.

\begin{lemma}[ProgressSync FTS BTs]
\label{th.APSFTS}
Let $\bt_1$ and $\bt_2$ be two FTS, with region of attraction $R_1$ and $R_2$ respectively, and active sub-BTs in the BT $\bt$. The sub-BTs $\tilde{\bt_1} = DecoratorSync(\bt_1, b(x_k))$ and $\tilde{\bt_2} = DecoratorSync(\bt_2, b(x_k))$ in the BT $\tilde \bt$ obtained by replacing in $\bt$ $\bt_1$ and $\bt_2$ with $\tilde{\bt_1}$ and $\tilde{\bt_2}$ respectively, are FTS.
\begin{proof}
Since the  $\bt_1$ and $\bt_2$ are active they will receive ticks as long as their return status is running, from Equations~ \eqref{bts.def:CBT} and~\eqref{eq:form:progress:r}, $\tilde{\bt_1}$ and $\tilde{\bt_2}$ have the same return statuses of $\bt_1$ and $\bt_2$ respectively, they both will receive ticks as long as they return status is running.  Since $\bt_1$ and $\bt_2$ are FTS with region of attraction $R$ they will eventually reach a state $x_{\bar k} \in S_i$, which implies $r_i(x_{\bar k}) = S$ hence eventually $p_i(x_{\bar k}) =  1$. In such case the the decorator propagate every tick that it receive. 
\end{proof}
\end{lemma}

\begin{corollary}[of Lemma~\ref{th.APSFTS}]
Let $\bt_1$, $\bt_2$, $\cdots$, and $\bt_N$ be $N$ FTS with region of attraction $R_i$ and active sub-BTs. Each sub-bt $\tilde{\bt_i} = DecoratorSync(\bt_i, B)$  is FTS if $r_i(x_k) = S \implies p_i(x_k) = 1$ hold.
\begin{proof}
The proof is similar to the one of Lemma~\ref{th.APSFTS}. 
\end{proof}
\end{corollary}
%
%

\begin{lemma}
Let $\bt_1$ be a safeguarding BT with respect to the step length $d$, the obstacle region $O \subset \mathbb{R}^n$, and the initialization region $I \subset R$. Then $\bt_0 = ProgressSync(\bt, b(x_k))$ is also safeguarding with respect to the step length $d$, the obstacle region $O \subset \mathbb{R}^n$, and the initialization region $I \subset R$, for any value of $b(x_k)$.
\begin{proof}
From Definition~\ref{def:Safeguarding}, $\bt$ holds the following:  $\{x: \inf_{s\in S_1} || x-s  || \leq d \} \subset I$ hence $|f_1(x_k) - x_k| \leq d$ holds. From Definition~\ref{def:progress}, $f_0(x_k)$ is either $f_1(x_k)$ or $x_{k}$, hence  hence $|f_0(x_k) - x_k| \leq d$ holds.
\end{proof}
\end{lemma}
\begin{lemma}
Let $\bt_0 = ResourceDecorator(\bt_1, g)$, if \\$g(x_k)>0$ $\forall x_k \in \mathbb{R}$ then the execution of $\bt_0$ is starvation-free regardless the resources allocated.
\begin{proof}
According to Equation~\eqref{eq:form:resource:g}, since $g(x_k) > 0$, whenever the $\bt_0$ does not propagate ticks to the BT $\bt_1$ 
it gradually increases the priority of $\bt_1$. 
\end{proof} 
Setting $g(x_k) > 0$ implements aging, a technique to avoid starvation~\cite{tanenbaum2015modern}. We could also shape the function $g$ such that it implements different scheduling policies. However, that falls beyond the scope of the paper.
\end{lemma}

\section{Conclusions}
\label{sec:conclusions}
This paper proposed two new BTs control flow nodes for resource and progress synchronization with different synchronization policies, absolute and relative. We proposed measures to assess the synchronization between different sub-BTs and the predictability of robot execution. Moreover, we observed how design choices for synchronization might affect the performance. The experimental validation supports such observations. 

We showed our approach's applicability in a simulation system that allowed us to run the experiments several times in different settings to collect statistically significant data. We also showed the applicability of our approach in real robot scenarios taken from the literature. We provided the source code of our experimental validation and the code for the control flow nodes aforementioned. Finally, we studied the proposed node from a theoretical standpoint, which allowed us to identify the assumptions under which the synchronization does not jeopardize some BT properties.

\section*{Acknowledgment}
This work was carried out in the context of the SCOPE project, which has received funding from the European Union's Horizon 2020 research and innovation programme under grant agreement No 732410, in the form of financial support to third parties of the RobMoSys project. We also thank, in alphabetical order, Fabrizio Bottarel, Marco Monforte, Luca Nobile, Nicola Piga, and Elena Rampone for the support in the experimental validation. 
\balance
\bibliographystyle{IEEEtran}
\bibliography{behaviorTreeRefs}	
\clearpage

\nobalance

\end{document}